\documentclass{article}
\usepackage{jfrExamplee}
\usepackage{graphicx}
\usepackage{apalike}
\usepackage{setspace}
\usepackage{authblk}
\usepackage{array}
\usepackage{floatrow}
\newfloatcommand{capbtabbox}{table}[][\FBwidth]
\usepackage{subcaption}

\usepackage{multirow}
\usepackage{blindtext}
\usepackage{graphics} 
\usepackage{placeins}
\usepackage{comment}
\usepackage{amsmath}
\usepackage{amssymb}
\usepackage[ruled,linesnumbered]{algorithm2e}
\usepackage[dvipsnames]{xcolor}

\usepackage[backend=biber,      
    style=ieee,
    bibstyle=ieee,              
    mincitenames=1,
    maxcitenames=2,
    sortcites=true,
    giveninits=true,            
    backref=false
]{biblatex}
\newlength{\imagewidth}
\newcommand{\subgraphics}[2]{
    \settowidth{\imagewidth}{\includegraphics[height=#2]{#1}}%
    \begin{subfigure}{\imagewidth}%
        \includegraphics[height=#2]{#1}%
    \end{subfigure}%
}

\title{Fast and Modular Autonomy Software for Autonomous Racing Vehicles}
\author[1]{Andrew Saba}

\author[6]{Aderotimi Adetunji}
\author[1]{Adam Johnson}
\author[3]{Aadi Kothari}
\author[1]{Matthew Sivaprakasam}
\author[1]{Joshua Spisak}

\author[2]{Prem Bharatia}
\author[1]{Arjun Chauhan}
\author[2]{Brendan Duff Jr.}
\author[2]{Noah Gasparro}
\author[5]{Charles King}
\author[4]{Ryan Larkin}
\author[4]{Brian Mao}
\author[2]{Micah Nye}
\author[3]{Anjali Parashar}

\author[5]{Joseph Attias}
\author[2]{Aurimas Balciunas}
\author[5]{Austin Brown}
\author[3]{Chris Chang}
\author[2]{Ming Gao}
\author[3]{Cindy Heredia}
\author[5]{Andrew Keats}
\author[3]{Jose Lavariega}
\author[2]{William Muckelroy III}
\author[4]{Andre Slavescu}
\author[3]{Nickolas Stathas}
\author[1]{Nayana Suvarna}
\author[4]{Chuan Tian Zhang}

\author[1]{Sebastian Scherer}
\author[1]{Deva Ramanan}

\affil[1]{Carnegie Mellon University}
\affil[2]{University of Pittsburgh}
\affil[3]{Massachusetts Institute of Technology}
\affil[4]{University of Waterloo}
\affil[5]{Rochester Institute of Technology}
\affil[6]{Cornell University}

\begin{document}

\maketitle

\begin{abstract}
 Autonomous motorsports aim to replicate the human racecar driver with software and sensors. As in  traditional motorsports, Autonomous Racing Vehicles (ARVs) are pushed to their handling limits in multi-agent scenarios at extremely high ($\geq 150mph$) speeds. This Operational Design Domain (ODD) presents unique challenges across the autonomy stack. The Indy Autonomous Challenge (IAC) is an international competition aiming to advance autonomous vehicle development through ARV competitions. While far from challenging what a human racecar driver can do, the IAC is pushing the state of the art by facilitating full-sized ARV competitions. This paper details the  MIT-Pitt-RW Team's approach to autonomous racing in the IAC. In this work, we present our modular and fast approach to agent detection, motion planning and controls to create an autonomy stack. We also provide analysis of the performance of the software stack in single and multi-agent scenarios for rapid deployment in a fast-paced competition environment. We also cover what did and did not work when deployed on a physical system (\textit{the Dallara AV-21 platform}) and potential improvements to address these shortcomings. Finally, we convey lessons learned and discuss limitations and future directions for improvement.

\end{abstract}

\section{Introduction}

Historically, motorsports have been a venue for advancing automotive technology in the name of competition and brand recognition. Teams develop increasingly sophisticated technologies to shave off seconds from lap times. Over time, technology and lessons learned from car racing have been commercialized and adopted in standard passenger vehicles. With the advent of Autonomous Vehicle (AV) technology, motorsports are poised to play a similar role in its development. Autonomous Racing Vehicle (ARV) leagues, such as the Indy Autonomous Challenge (IAC) and Roborace, are challenging software, not drivers, to operate a vehicle at the performance limit.

Apart from motorsports, AVs are starting to be adopted for public use \cite{caAuthorizesCruise}. These conventional AVs are either specially retrofitted passenger vehicles, typically deployed in urban and suburban environments, or tractor-trailers, customized for and deployed in long-haul, highway, and interstate environments. In all of these domains, safety under all conditions and circumstances is vital; however, no matter the size and scope of any test program, edge cases, by the very nature of their rarity and difficulty, will continue to challenge safety verification and necessitate further development.

\begin{figure}[t!]
\centering
\includegraphics[width=.9\linewidth]{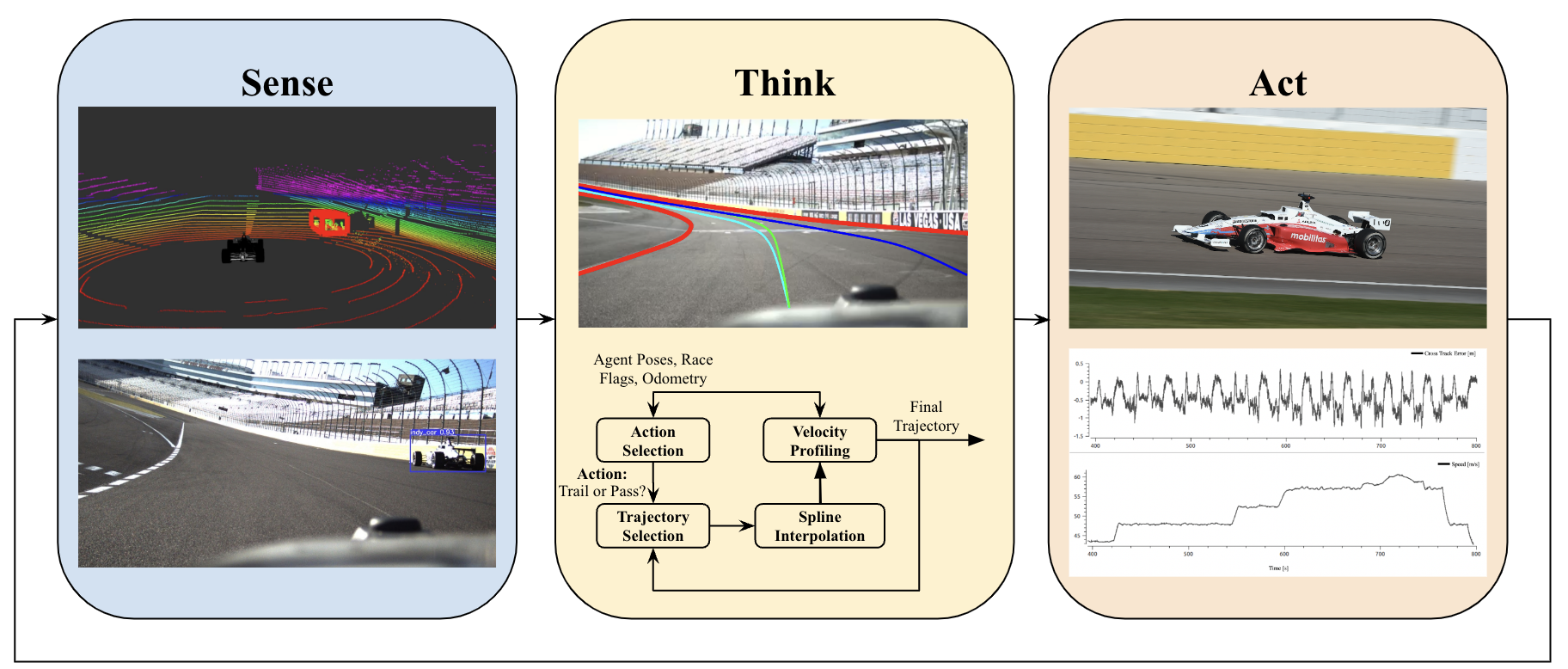}
   \caption{Autonomy for AVs and ARVs is typically distilled into three broad tasks: \textbf{Sense}, \textbf{Think}, and \textbf{Act}. \textbf{Sense} is measuring the state of the environment. \textbf{Think} is deciding the best course of action to take. \textbf{Act} is executing on that course of action.}
\label{fig:sense_think_act}
\end{figure}

The three broad tasks commonly associated with software for AVs and ARVs are \textbf{Sense}, \textbf{Think}, and \textbf{Act}. Figure \ref{fig:sense_think_act} shows the relationship between the three tasks.

\begin{figure}[t!]
\centering
\includegraphics[width=\linewidth]{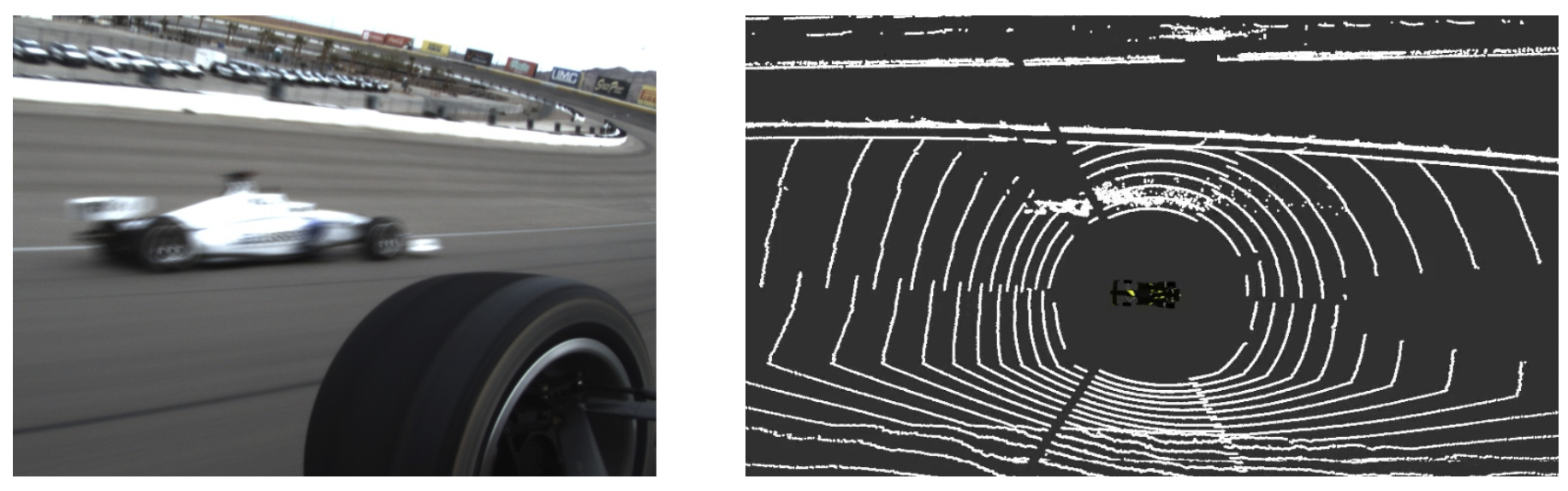}
   \caption{Example of perception challenges faced by Autonomous Racing Vehicles (ARVs). \textbf{(Left)} Speed deltas between agents can be enough to cause motion blur. \textbf{(Right)} As the other agent passes, their acceleration is kicking up significant dirt, dust, and debris, which challenges LiDAR detection. Overall, sensor noise impairs robust detection and localization of opponent ARVs.}
\label{fig:motion_blur}
\end{figure}

\textbf{Sense} uses sensors to measure the state of the environment. Sensors in ARVs detect and track the opponent and measure the specific gravity and angular rate of the ego vehicle. In the case of a conventional AV, sensors detect and track pedestrians and other vehicles on the road. However, unlike an ARV, AVs typically localize themselves onto High Definition (HD) Maps, which act as a strong prior to understand the environment, rules, and semantics of the road. For an ARV, prior information about the track bounds, banking, and theoretical maximum dynamic limits can be pre-computed for locations along the track. Figure \ref{fig:motion_blur} shows some of the challenges sensors encounter, such as occlusions, degraded sensor quality, and high aliasing due to high acceleration and noise.  ARV sensors must have fast processing to handle conditions akin to highways, such as unexpected events, previously unseen agents, or high speeds. While the environment of an ARV has fewer actors at any given time, those actors are capable of extremely high (over $20m/s^2$) accelerations and speeds. This demands low latency and long-range perception to allow other modules ample time to react. Lastly, high speeds and accelerations introduce noise and unique physical challenges to sensors that often require robust software solutions.

\textbf{Think} processes sensor information into a prediction of how the environment will evolve, ultimately deciding the best course of action to take next. For example, in the case of an ARV, it may be whether or not an opponent is attempting an overtake and whether or not it should defend against it. However, for an AV, the question may be whether or not it is safe to take a left turn. In both scenarios, significant uncertainty exists in how the world will evolve into the future, complicating decision-making. However, in an ARV, an additional layer of uncertainty is considered because agents are capable of very high accelerations (over $20m/s^2$) and are operating in direct competition with one another. Higher speeds often mean that over a given distance traveled, there are fewer measurements of the other agent's motion, meaning that a prediction must be made in less time with less information. Additionally, predictions can become obsolete rapidly, as the other agent may rapidly change its motion and trajectory. Finally, intelligently predicting what an agent will do over the next $N$ seconds is complex in urban or highway environments, where rules and regulations guide what slower-moving actors can and will do. In racing, there is an added layer of complexity due to every actor attempting to win and far fewer rules dictating appropriate behavior. Deciding the best actions to take must consider multiple potential futures and which one will result in the greatest chance of winning.

\textbf{Act} executes the actions that were decided upon. For both ARVs and AVs, this involves taking the sequence of actions decided upon (typically a target space-time trajectory) and determining the optimal set of control commands to follow it safely. A traditional AV must often navigate complicated and crowded environments, react quickly to commands, and handle changing road and environmental conditions, such as snow and ice. Safely navigating while slipping on snow and ice is still an open area of research, as it requires advanced vehicle dynamics modeling and control techniques \cite{tri}. While they are not driven in snow, racecars are pushed to their handling limits, to the point that simple kinematic and dynamic models begin to fall apart, much like in slippery conditions. The highly non-linear tire dynamics, aerodynamic interactions, effects of track temperature and surface, and more begin to break down the assumptions made in these models. Additionally, these effects are not constant, changing as the race progresses. For example, a worn-out tire on a racecar is like a typical passenger vehicle trying to drive through snow: both are prone to slipping at any moment. Any controller navigating an ARV at the physical limits of handling must understand and account for these dynamics to balance high performance and safety. 

\subsection{Related work}
\begin{figure}[h!]
\centering
\includegraphics[width=0.9\linewidth]{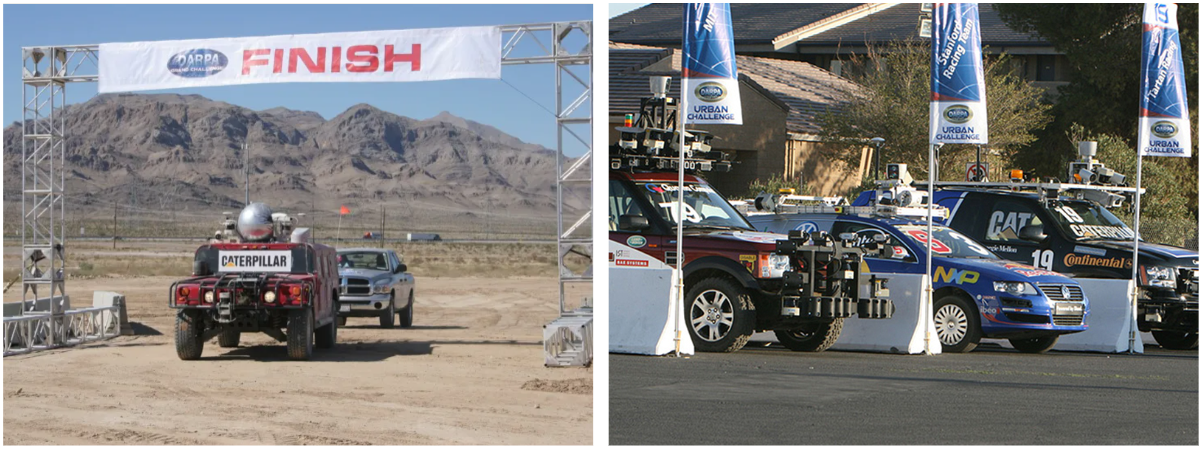}
   \caption{DARPA Grand and Urban Challenges, two prize competitions for American autonomous vehicles. The Challenges were created to spur the development of autonomous vehicle technologies capable of completing a substantial off-road course, and later an urban environment course, within a limited time. Images from \cite{grand_challenge}\cite{urban_challenge}.}
\label{fig:urban_challenge}
\end{figure}

The call for advancing autonomous vehicle technology has been present since the early twenty-first century when the Defense Advanced Research Projects Agency (DARPA) launched the 2004 and 2005 DARPA Grand Challenges \cite{DARPA2004}, \cite{DARPA2005}. These challenges, shown in Figure \ref{fig:urban_challenge}, demonstrated some of the capabilities of AVs. The teams in the Grand Challenge autonomously navigated across southern Nevada on a 132-mile course of rugged desert terrain. Succeeding the Grand Challenges was the 2007 DARPA Urban Challenge \cite{DARPA2007}, which introduced a time-based competition focused on city driving. This competition maintained the competitive nature of completing a course, but focused on navigating an urban environment. Each team needed to stop at stop signs, yield for oncoming traffic, complete U-turns, and obey all other traffic laws. These challenges were the first full-scale autonomous racing competitions and laid the groundwork for future AV research and development.


Since then, there have been several autonomous racing competitions, such as Formula Student \cite{AMZ_Driverless}, \cite{ETH_Driverless_Planning}, Roborace \cite{roborace1}, \cite{roborace2}, and now the Indy Autonomous Challenge. Moreover, companies such as Argo AI, Motional, Waymo, and many more have been publicizing and realizing AV development around the globe. These companies have tasked themselves with challenges to motion plan in dynamic, unpredictable environments and perceive in inclement conditions. The challenges ARVs face differ, focusing on detecting vehicles, planning motion, and actuation while driving at speeds over $150mph$. Unlike AVs that operate in an open world, ARVs do not need to worry about cyclists on the road \cite{cyclist} or pedestrians crossing a street; however, the issue of making quick and accurate detections and actions remains a nontrivial problem still under research.

{\bf Perception (sense):} Although there is substantial work demonstrating perception in conventional AVs, less work has been published discussing the deployment of perception algorithms for autonomous racing. In Formula Student Driverless, the ARV drives on the track alone and is solely tasked with detecting white and blue cones \cite{AMZ_Driverless}, \cite{ETH_Driverless_Planning}. Currently, there are few works that present full perception stacks for detecting other agents for autonomous racing. One such work is \cite{tum_whole_stack}, which presents a full perception stack that utilizes camera, radar, and LiDAR sensors.
Improving perception efficiency is an extensively researched topic in AV development. The point-cloud clustering-based detection system in \cite{eu_clustering} is fast and efficient at detecting other actors.  Works such as \cite{Li2020StreamingP} and \cite{vision_streaming} focus on Streaming Perception, which emphasizes combining latency and accuracy when developing benchmarks for computer vision algorithms. Multi-modal perception is also a very well-studied area of research. The work in \cite{waymo_fusion} fuses LiDAR and camera features with a learned cross-modal attention alignment. In \cite{BEV_FUSION}, a combined LiDAR-camera birds-eye-view (BEV) projection is generated efficiently and can be used for downstream tasks such as object detection. In both works, multiple sensor modalities are fused early in the detection pipeline. 

{\bf Planning (think):} Motion planning involves determining the best sequence of actions to be taken and generating a trajectory to execute those actions. For oval racing, it is possible to distill the problem in a series of action primitives, including maintaining the current trajectory behind an opponent, e.g., "trailing" or passing. One way to approach the passing problem is to treat it as a sequence of lane changes, where the ego vehicle merges between several lanes of travel on the track. Lane merging is a well-studied area, with previous work utilizing polynomials \cite{LiuYonggang2022DLTP}, splines \cite{StahlTim2019MGTP}, \cite{FunkeJoseph2016SCLC}, or B$\acute{e}$zier curves to parameterize paths \cite{ZhengLing2020Bctp}. In \cite{polimove_planning}, $5^{th}$ order polynomials are generated within a Darboux frame using a convex combination of the origin and target paths. Heading and curvature continuity is guaranteed for any lane change maneuver without numerical differentiation. In addition, compliance with track boundaries is also guaranteed a \textit{priori}.

{\bf Controls (act):} While AVs cannot yet legally drive faster than 80 to $130mph$, depending on locale and road conditions \cite{NHSTA}, the operating domain for ARVs is typically 100 to $200+mph$. This sizeable difference dictates the differences in vehicle architectures, modeling, and controller techniques. Additionally, at higher speeds, assumptions made in vehicle dynamics models may not apply, necessitating better vehicle modeling. Some works have explored addressing this problem by combining a model predictive controller (MPC) with a deep-learning-based model \cite{deep_learning_control}, \cite{MPC_book}. Other works, such as \cite{tum-friction}, look to estimate tire friction parameters online for use in an MPC controller. Finally, due to model limitations at higher speeds, the controller must be robust and capable of reasoning about potential bounds on actual dynamics. Works such as \cite{wischnewski2022tube} address this with a Tube-MPC controller that can reason about uncertainty in the dynamics. Additional work, such as \cite{TII_Control}, layout a whole navigation stack for use in an ARV. Finally, for our approach, we looked to robust and fast controllers, such as linear-quadratic regulators (LQR) and iterative LQR control \cite{iLQR}, \cite{iLQR2}.

\subsection{Overview \& Key Takeaways}

Our approach follows two main themes: modularity and speed. We have developed every portion of our software stack (hereinafter referred to as ``the stack") to be stand-alone, allowing for replacing modules as requirements change. Since very little prior work existed with racing at the speeds the competition demands, it is challenging to develop a one-size-fits-all approach. There is a high level of uncertainty because of the many unknowns regarding how sensors or the vehicle will behave at higher speeds. Additionally, with a fast-paced competition and prototype hardware, requirements change day-to-day, necessitating frequent modifications to core functionality.

Secondly, we define our approach by its speed. When racing at high speeds, algorithms must finish execution quickly and deterministically, i.e., sudden high execution times can be disastrous if they lead to instability. For example, our motion controller uses a dynamics motion model to generate an optimal feedback policy that is cheap and fast to compute, allowing for a high rate of execution with little deviation, which is vital for navigating at very high speeds and accelerations. However, our approach does not sacrifice quality to achieve its speed and efficiency; instead, the key challenge has been to choose algorithms intelligently and design efficient architectures around them.

In this work, we present a detailed description of our approach and system design for a full ARV software stack for the Indy Autonomous Challenge (IAC). We will also elaborate on successes, failures, and lessons learned during extensive field testing on oval race tracks over two competition seasons. Finally, we will provide insights in our design process across the whole stack and results from this approach. Overall, our stack demonstrates the following capabilities:
\begin{itemize}
    \item Stable trajectory tracking at speeds over $150mph$ while maintaining reasonable lateral deviations from the desired trajectory
    \item Reliably detecting and tracking an opponent ARV at over $100m$ away, even at high speeds (i.e. $\geq 125mph$)
    \item Safely passing and trailing an opponent ARV vehicle at high speeds (i.e. $\geq 125mph$)
\end{itemize}

\section{The Competition}\label{section:competition}

\begin{figure}[h!]
\centering
\includegraphics[width=.5\linewidth]{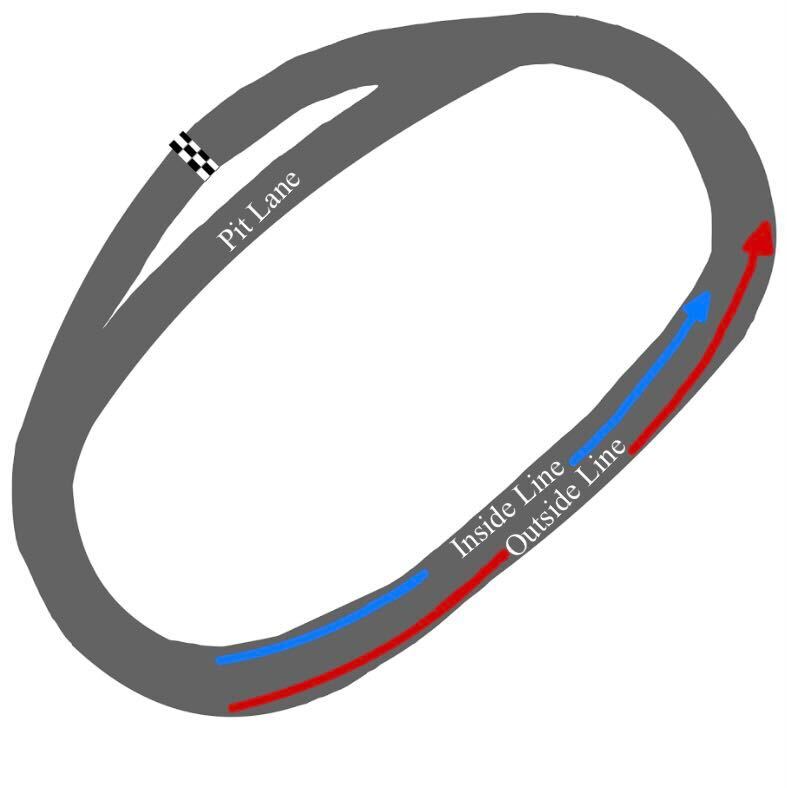}
   \caption{Racing lanes for the Autonomous Challenge @CES at the Las Vegas Motor Speedway (January 2022/January 2023)}
\label{fig:competition_rules}
\end{figure}

The MIT-Pitt-RW autonomous racing team is a team of students forming one of the nine teams that have successfully qualified to participate in the Indy Autonomous Challenge (IAC). The IAC is a global competition in which university teams compete to develop software for a standardized ARV platform named the Dallara AV-21. To date, there have been two seasons with two physical installments each, as seen in Table \ref{tab:iac-events}. The first installment was a single-agent, fastest-lap competition, and the following installments have been two-agent passing competitions. These installments have all been on oval super-speedways which has dictated the strategies for developing the stack. Before the in-person installments, there were multiple simulation practice events and a simulation competition, where the teams verified their software in single and multi-agent scenarios. All instalments of the competition have been supervised and executed by "race control", managed by the IAC. During competition, the race flags and team roles are remotely controlled by race control, allowing for minimal human intervention during the race. 

\begin{table}[h!]
    \centering
    \begin{tabular}{ |p{.4\textwidth}|p{.4\textwidth}|  }
        \hline
            \multicolumn{2}{|c|}{\textbf{Indy Autonomous Challenge Events}} \\
        \hline
            \multicolumn{2}{|c|}{\textbf{Season One (2021-22)}} \\
        \hline
            \multicolumn{1}{|c|}{\textbf{Track}} & \multicolumn{1}{c|}{\textbf{Event Format}} \\
        \hline
            Indianapolis Motor Speedway (IMS) & Single-Agent with static avoidance \\
            Las Vegas Motor Speedway (LVMS) & Passing Competition \\
        \hline
            \multicolumn{2}{|c|}{\textbf{Season Two (2022-23)}} \\
        \hline
            Texas Motor Speedway (TMS) & Passing Competition, relaxed racing lines \\
            Las Vegas Motor Speedway (LVMS) & Passing Competition, relaxed racing lines \\
        \hline
    \end{tabular}
    \caption{The Indy Autonomous Challenge (IAC) has held four events thus far, at three different tracks, with two formats. Season Two saw a continuation of the passing competition introduced at the AC@CES2022 installment in Las Vegas. For Season Two, the defender is given more freedom on the racing line they may take, which makes passing more challenging.}
    \label{tab:iac-events}
\end{table}

The multi-agent passing competition \cite{iac_rules} assigns one competitor the "attacker" role and the other the "defender" role. During each lap, the defender is remotely assigned a speed at which the attacker must pass the defender within two laps. If the pass is successful, the roles are exchanged and the defender's speed is incrementally increased ($125mph$, $135mph$, etc.). A pass is complete once the attacker gains its position in front of the defender with a longitudinal gap of at least $30m$. If an attacker fails to pass at a certain speed, the roles are exchanged. The winner of the round is determined once one of the attackers cannot complete a pass. If both teams cannot complete the pass, the round ends in a draw. Figure \ref{fig:competition_rules} shows a breakdown of the track and the possible paths to take into consideration.

There are two factors the attacker must consider while deciding to make a pass: safety and dynamic limitations. The attacker must maintain safe lateral and longitudinal separation from the defender at all times. Additionally, when considering a pass, the attacker must ensure its trajectory will keep the car within the dynamic limitations of the vehicle and not result in loss of control. Combining the two, the attacker must ensure that the accelerations and decelerations are timed appropriately to stay within the dynamic limits of the vehicle. As the passing competition progresses, it becomes more difficult for the attacker to exceed the defender's speed, particularly in corners of the track. 

In addition to these considerations for the attacker, the defender can make passing more difficult for the attacker by adjusting their position within their lane, as long as they maintain rule compliance. For example, if the defender moves outwards, the attacker has to travel more distance to complete the pass. A winning strategy for an attacker is to maintain the minimum allowed distance to the defender and to initiate the pass whenever the attacker can maneuver it safely. Completing a pass is also further complicated if the attacker starts the pass too late; they may get trapped too far out on the outer lane into the corners, thereby increasing the distance they need to cover. Prediction and motion forecasting of the opponent agent is imperative to make intelligent strategic decisions.

In this work, we present our approach to the Indy Autonomous Challenge (IAC) for both the 2021-22 and 2022-23 Seasons ("Season One" and "Season Two", respectively). Table \ref{tab:iac-events} shows a timeline of the four IAC events. The results shown here are from the the Las Vegas event in Season One and the Texas and Las Vegas events in Season Two.  The overall approach was the same for all four events, but more mature and better tested by the Season Two, evidenced by the more than doubling of our highest achieved speed from $69mph$ to over $150mph$. This multi-season evaluation provides a unique perspective into a continual and evolving engineering and testing process, with numerous lessons learned along the way.

\subsection{AV-21 platform}

\begin{figure}[h!]
\centering
\includegraphics[width=.7\linewidth]{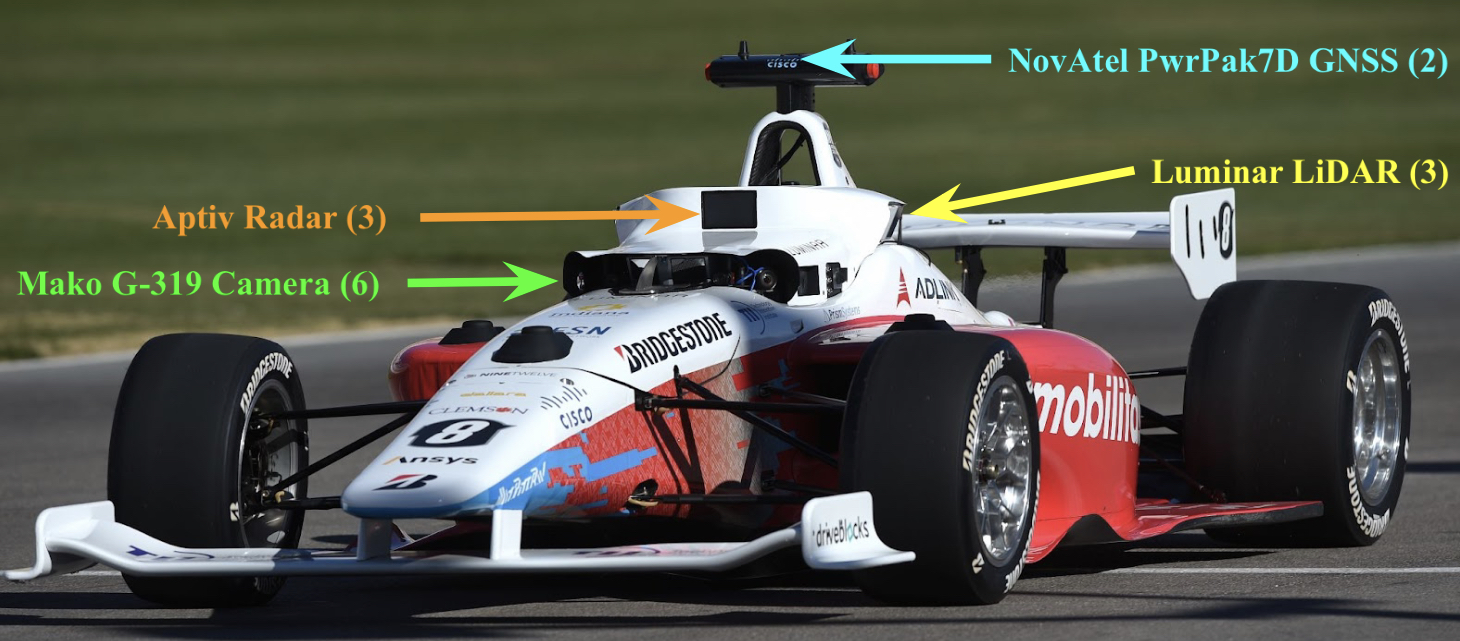}
   \caption{Sensors on the AV-21. Six cameras and three LiDARs provide redundant $360^{\circ}$ coverage and over $200m$ of sensing range.}
\label{fig:sensors}
\end{figure}

The Dallara AV-21 is the official vehicle of the Indy Autonomous Challenge (IAC). Every competitor must use the same hardware, including vehicle setup, autonomy sensors, and compute. The vehicle is a modified version of the Indy Lights IL-15 chassis, retrofitted with a package of automated vehicle sensors, drive by wire, and compute. The engine is a 4 Piston Racing-built Honda K20C. Sensors onboard the AV-21 include 3 Luminar Hydra LiDARs, 3 Aptiv Medium Range Radars, 2 NovAtel PwrPak7D-E1 GNSS, and 6 Mako G-319 Cameras. In total, these sensors provide redundant $360^{\circ}$ coverage and over $200m$ of sensing range. Figure \ref{fig:sensors} shows the AV-21 platform and sensor locations.

\begin{figure}[h!]
\centering
\includegraphics[width=.9\linewidth]{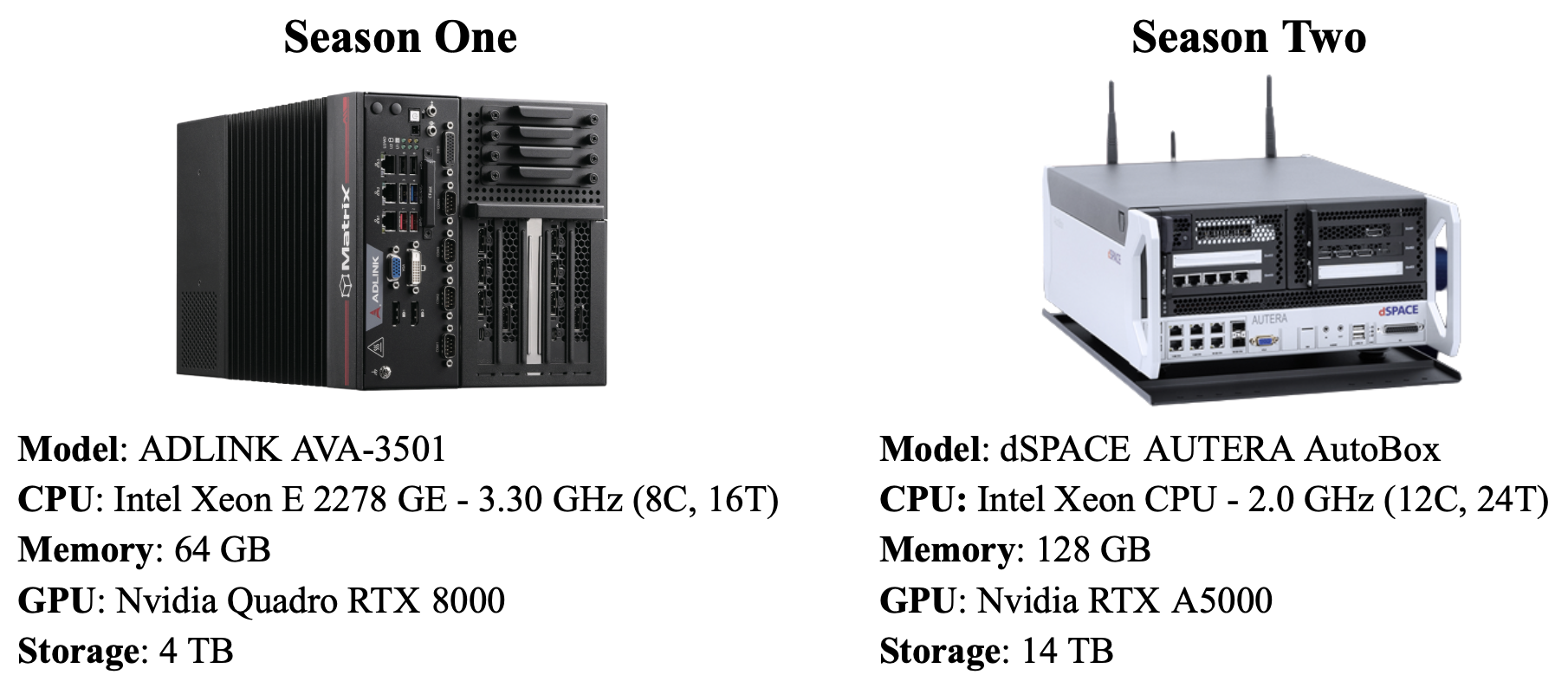}
   \caption{Main compute platforms for Seasons One and Two of the IAC. Each platform has numerous advantages and disadvantages when compared to one another. While the AutoBox provides automotive-level reliability and integration, its slower per core clock speeds resulted in worse single threaded performance. Since many critical core algorithms are fundamentally single threaded (i.e. controller calculations, state estimation updates, etc.), careful considerations were made into what ultimately ran on vehicle to ensure enough capacity for the entire stack. Specifications and images taken from respective product websites\cite{adlinktechTitle}\cite{dspaceAUTERA}.}
\label{fig:s1_s2_compute}
\end{figure}

Between Seasons One and Two of the IAC, the AV-21 underwent a hardware refresh that included the addition of a VN-310 Vectornav GNSS system and an update to the main compute platform. In Season One, the main compute was an ADLINK AVA-3501 with an 8 core, 16 thread Intel Xeon CPU and an NVIDIA Quadro RTX 8000 GPU. In Season Two, a dSPACE AUTERA AutoBox with a 12 core, 24 thread Intel Xeon CPU and an RTX A5000 NVIDIA GPU served as the main compute platform. The AutoBox provided many advantages over the ADLINK, including automotive-grade ruggedness, higher available networking bandwidth, and CAN channels built into the computer. However, these automotive features came at the cost of slower single threaded performance, which necessitated critical  engineering design decisions to accommodate all critical software pieces onto a single computer. A full breakdown of the compute platform differences can be seen in Figure \ref{fig:s1_s2_compute}. 

\section{Approach}

\begin{figure}[h!]
\centering
\includegraphics[width=.85\linewidth]{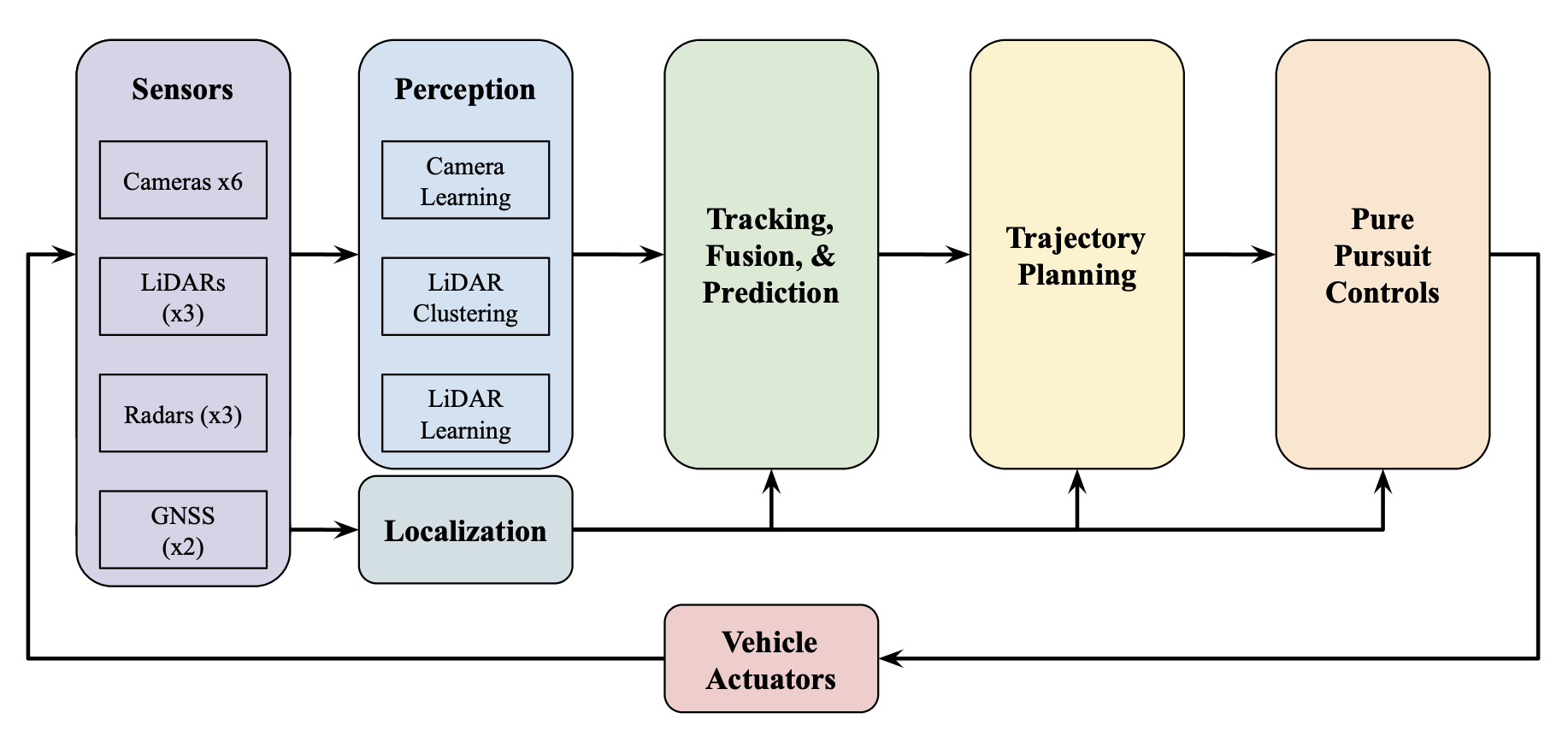}
   \caption{Overview of the complete stack. The perception module in particular is designed to be modular itself, allowing multiple algorithms developed over time to be substituted for one another. This modularity, present throughout the whole stack, proved invaluable, as it provided the maximum flexibility in deciding what runs on the vehicle. The trade-off, however, is duplication and a potential sacrifice on peak performance.}
\label{fig:stack_overview}
\end{figure}

\subsection{Stack Overview}
Our software architecture follows a typical, standard autonomy software design, with localization, perception, tracking, prediction and motion planning, and controls. The Robot Operating System (ROS), specifically ROS 2 Galactic, is used for communication between each process, or node. Various libraries and frameworks are utilized from ROS for visualization, math utilities, communication, and more. Figure \ref{fig:stack_overview} shows the data flow of the whole stack. All modules run asynchronously with one another, usually on a preset frequency, except for perception and portions of localization, which are driven by sensor data arrival.

\subsection{Perception}

\begin{figure}[h!]
\centering
\includegraphics[width=.8\linewidth]{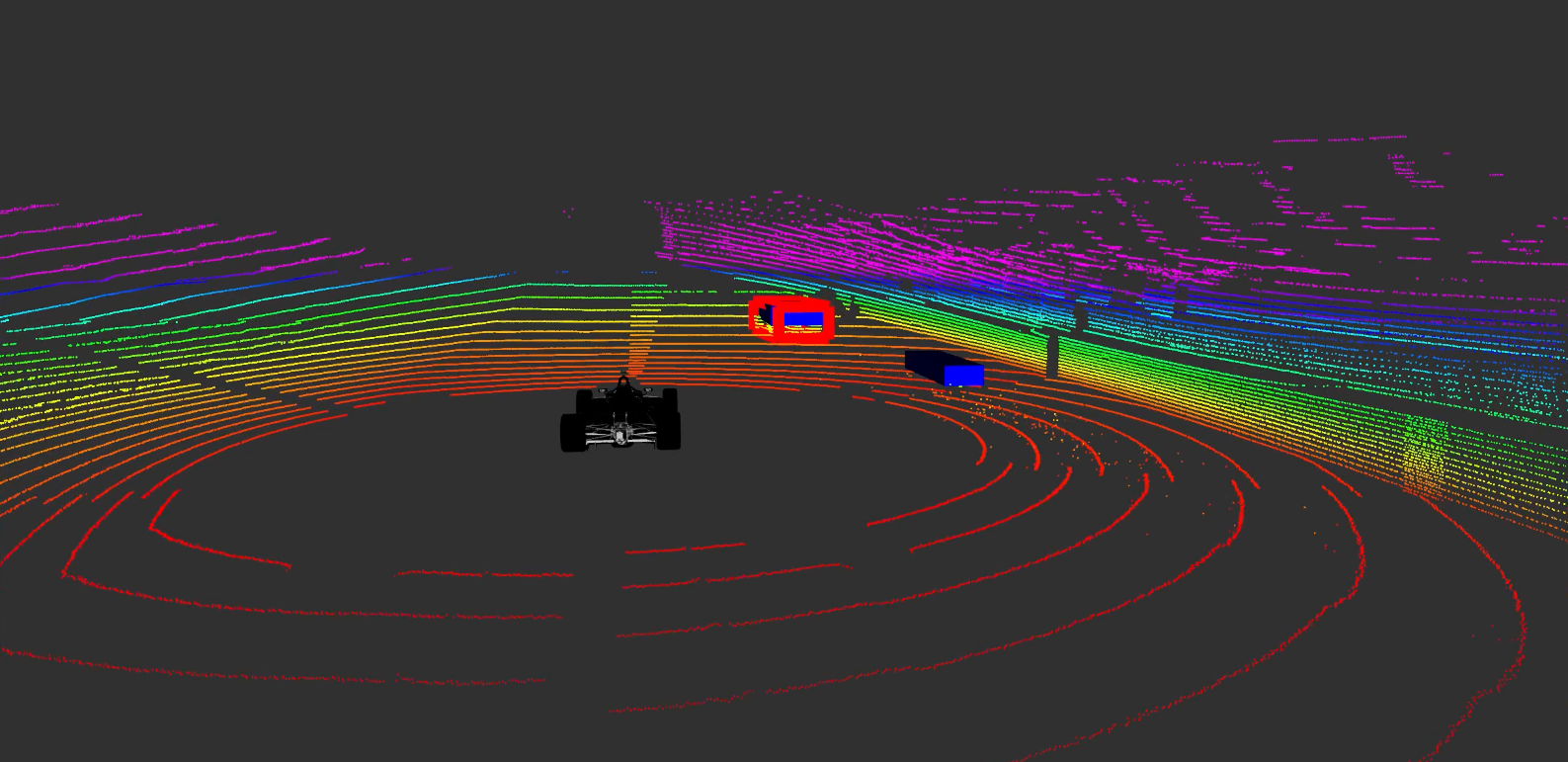}
   \caption{Red bounding box is from a data-driven detector (PointPillars) and blue is from clustering. Clustering is susceptible to detecting dust and noise as other agents. Clustering is unsupervised, but, as a result, is unable to differentiate between other agents and dust and debris. This motivated our final data-driven approach shown in Figure~\ref{fig:stack_perception}.}
\label{fig:cluster_bad}
\end{figure}

\subsubsection{Challenges and Requirements}

The AV-21 is capable of very high accelerations (greater than 20$m/s^2$) and speeds (greater than 180$mph$), meaning LiDAR or camera frame-to-frame movement can be significant. Additionally, there are no existing data sets for detecting AV-21s and little data showing the performance of sensors, such as LiDARs and cameras, at our target speeds. As a result, initially, data-driven approaches were not feasible, and the performance at higher speeds could not be immediately evaluated. 

{\bf Initial efforts:} Due to the lack of training data, our initial LiDAR perception approach for Season One utilized an unsupervised clustering algorithm. A Cloth Simulation Filter (CSF) \cite{csf} was used to remove ground points and a density-based clustering technique, DBSCAN \cite{dbscan}, was used to identify obstacles of specific dimensions within the bounds of the track. While seemingly viable, this approach proved to have many failure cases, such as identifying dust above the track as an obstacle, shown in Figure \ref{fig:cluster_bad}. By observing these failures and the potential to break down at higher vehicle speeds, the need for a robust, efficient, learning-based perception approach was evident.

\subsubsection{Overview of Approach}

\begin{figure}[h!]
\centering
\includegraphics[width=\linewidth]{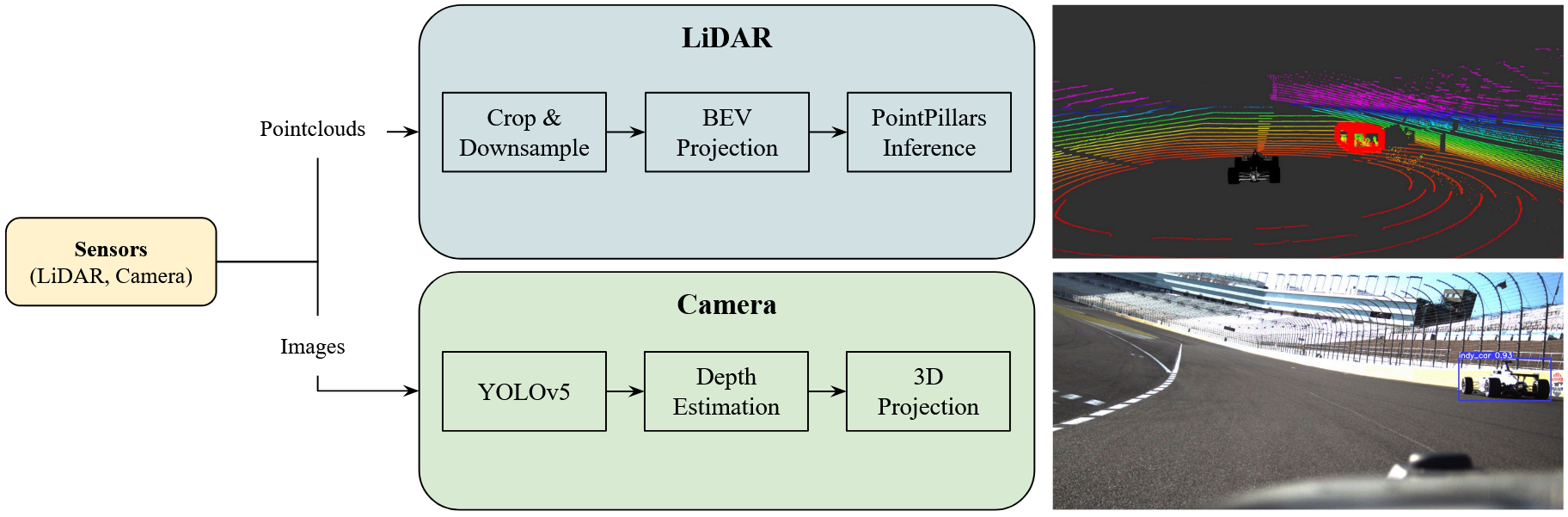}
   \caption{Perception Stack Overview. Because the sensors are processed independently, detection results are not dependent on the other sensor operating. This allows for a modular stack that is robust to a sensor modality failing. For example, if the cameras were not functioning correctly, the camera pipeline can easily be disabled without the rest of the stack being affected.}
\label{fig:stack_perception}
\end{figure}

\begin{figure}[h!]
\centering
\includegraphics[width=.7\linewidth]{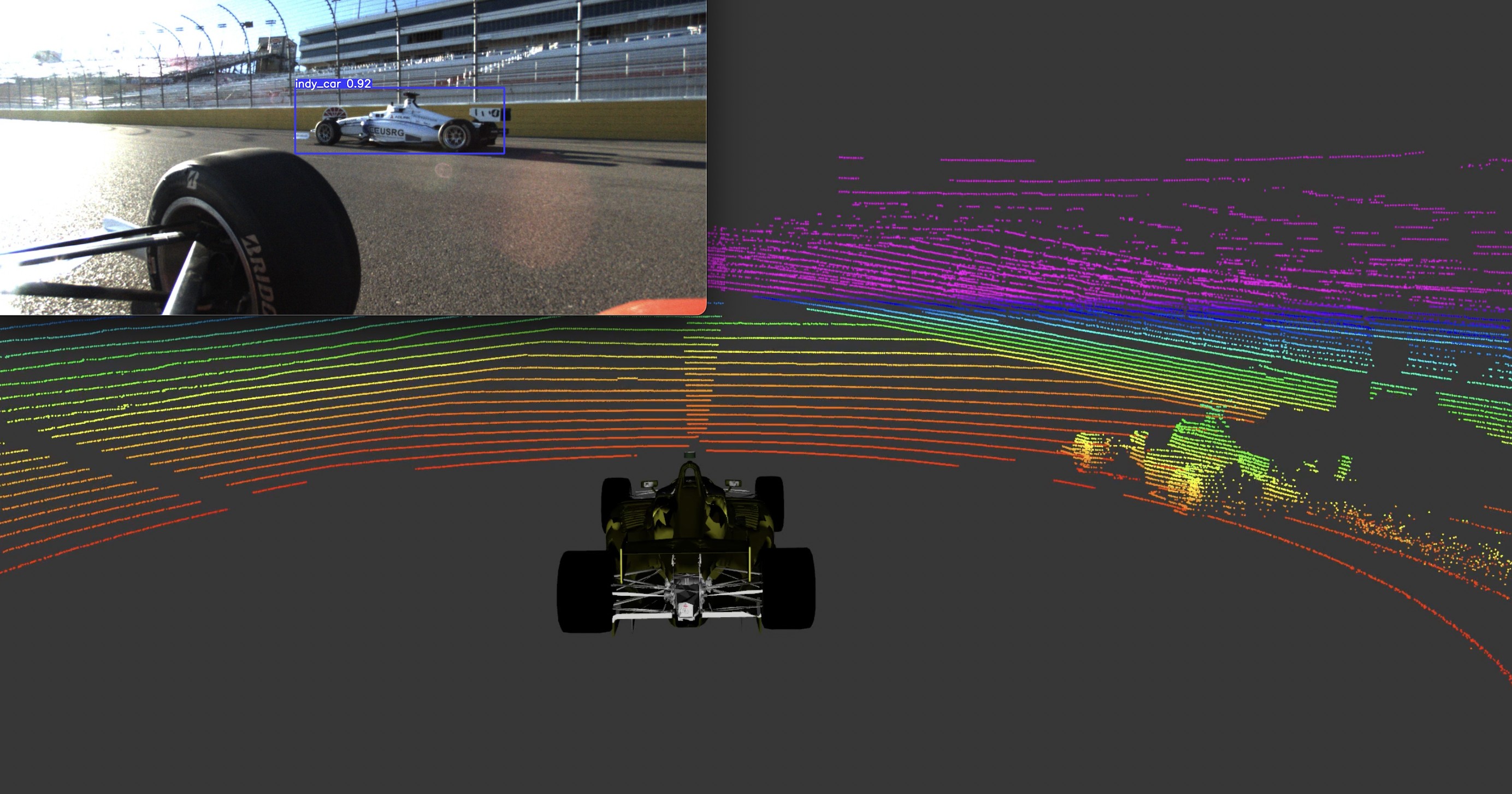}
   \caption{Perception Stack Challenges. Lens flare, vibrations and motion blur, glare, and other challenging light and environmental conditions make camera detection challenging. Additionally, dust challenges simpler, unsupervised LiDAR detectors, such as clustering, but requires robust data engineering to ensure a robust deep neural network model. Note: Point cloud is colorized on the z-axis.}
\label{fig:perception_challenge}
\end{figure}

{\bf Final perception stack:} Figure \ref{fig:stack_perception} shows our final perception stack's decoupled and multi-modal approach to accurately detecting and localizing all other agents on the track, which notably, no longer makes use of clustering. On the AV-21 Platform (Figure \ref{fig:sensors}), there is an assortment of LiDARs, cameras, and radars, each with advantages and disadvantages. For example, cameras alone do not give an accurate depth estimation but can operate at a much higher frame rate (up to $75Hz$) and resolution ($2064 \times 960$) than LiDARs. A camera-based detection pipeline can provide a higher frequency update on our belief of the world and has the potential to see other agents from further away. Figure \ref{fig:perception_challenge} shows some additional challenges faced with perception. To best exploit the sensors' strengths and for robustness and redundancy, our perception stack uses each sensor independently and in parallel and feeds all localized detections to our tracking stack. 

\subsubsection{Camera}

\begin{figure}[t!]
\centering
\includegraphics[width=.8\linewidth]{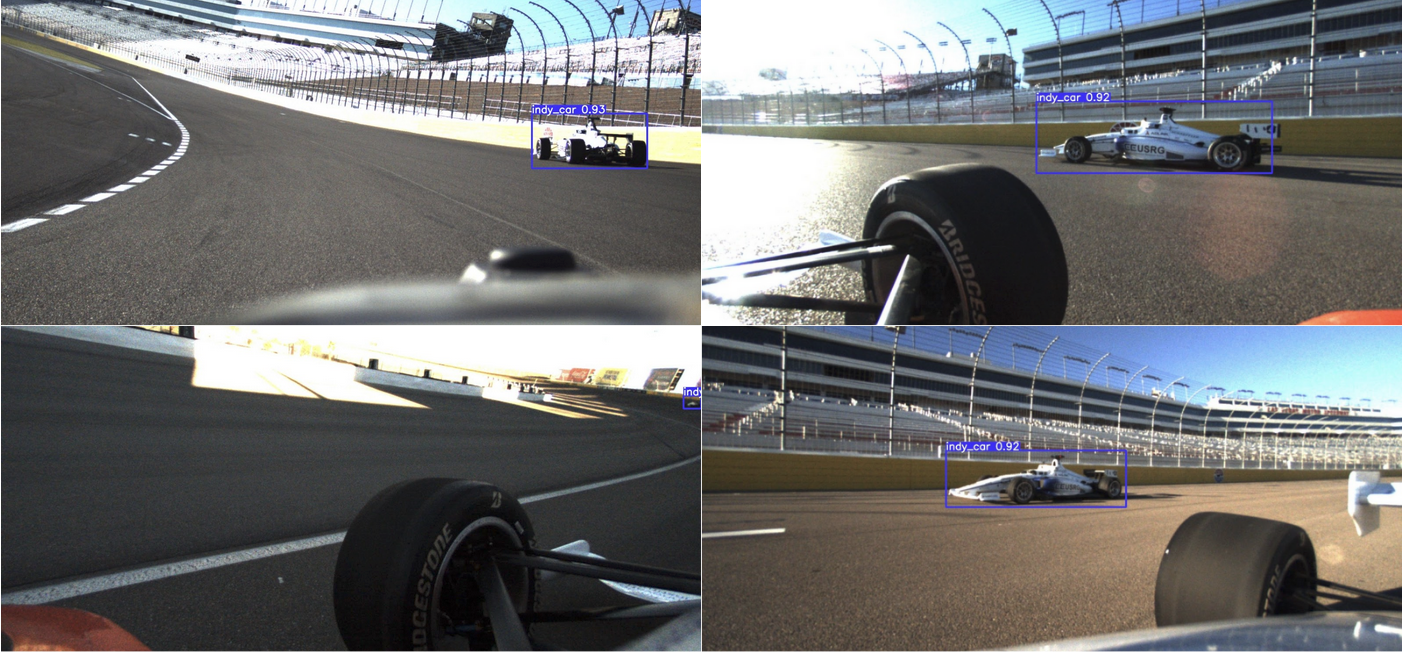}
   \caption{Camera Detection Results}
\label{fig:camera_detection_results}
\end{figure}

For full coverage and maximum range, the two front-facing cameras utilize a narrow lens to improve far-field resolution. The four remaining cameras use a wider field of view (FOV) to provide $360^\circ$ coverage around the vehicle. Due to the lower effort required to label 2D bounding boxes, YOLO v5 \cite{yolo_v5} was chosen for our initial approach. The model was trained on a custom, hand-labeled data set of other AV-21 vehicles, with images taken from onboard our vehicle. Because the model outputs 2D bounding boxes, other assumptions and processing is required to provide a 3D pose of the other agents. By exploiting the fact that the size and shape of the vehicles are known, we can estimate a depth from the 2D bounding boxes from the model by using a standard pinhole optics model~\cite{forsyth2011computer}:
\begin{equation} \label{eq1}
    Depth = \frac{(Height_{known} * f)}{Height_{pixels}}
\end{equation}
where $f$ is the calibrated focal length of the camera, $Height_{known}$ is the known height of the vehicle in meters, and $Height_{pixels}$ is the detected height of the detected vehicle in pixels. This monocular algorithm yields accurate results for mid/far-field detections; however, the error increases proportionally with the real-world distance between the camera and the other agent. While far-field detections ($>100m$) tend to be less accurate, the additional sensor modalities, including LiDAR, cannot see as far as the camera with nearly the exact resolution and fidelity, so some measurement is better than none. As the other agent gets into the LiDAR operating range, we refine the estimates using these detections, and our confidence in the agent's position increases. The unique long-range capability of the camera perception pipeline can provide motion planning more time to respond to agents in our path. Figure \ref{fig:camera_detection_results} shows the result of the camera detection pipeline. 

\subsubsection{PointPillars}

\begin{figure}[t!]
    \centering
    \includegraphics[width=.8\textwidth]{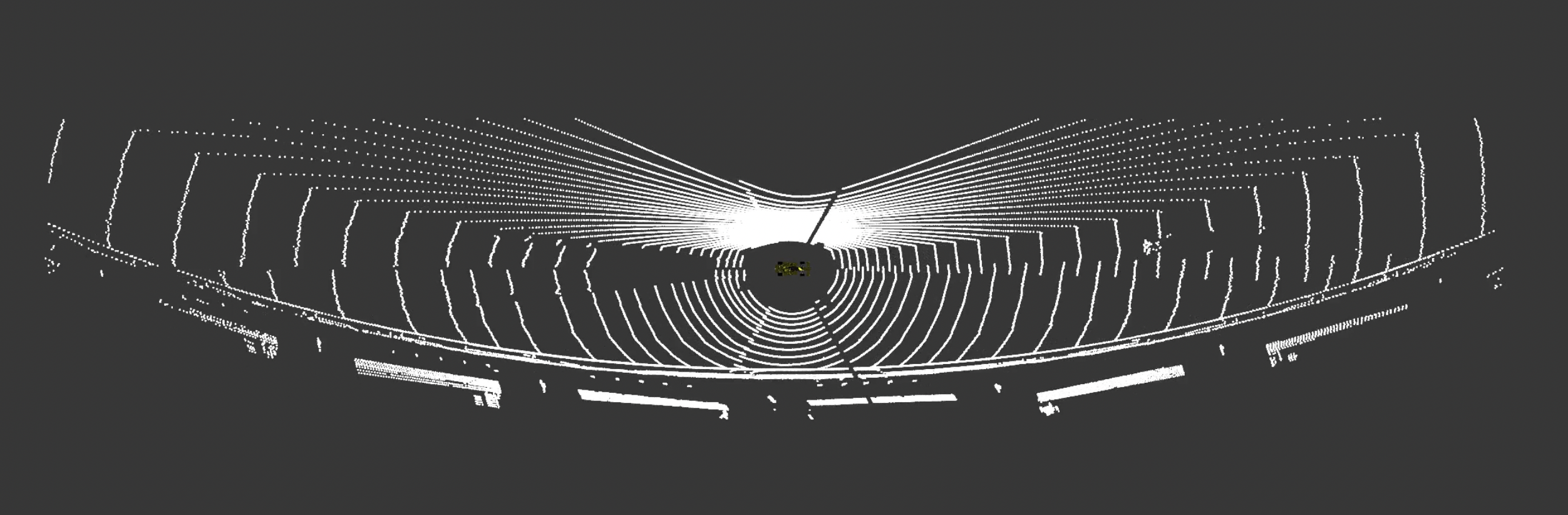}
    \caption{Pointcloud from all three Luminar LiDARs. The LiDARs each can see over 100 meters. The clouds are combined and cropped to provide coverage of $200m \times 40m$. }
    \label{fig:lidar_coverage}
\end{figure}

The AV-21 platform has three Luminar Hydra LiDARs\cite{luminarhydra} positioned in a triangular fashion. Each LiDAR has a field of view (FOV) of $120^\circ$, together allowing for $360^\circ$ coverage around the vehicle. Each LiDAR is capable of excellent coverage at over $100m$, thereby providing an over $200m$ radius circle of coverage around the track. Since the track is only so wide, this cloud is cropped further to being $200m \times 40m$. An example cloud can be seen in Figure \ref{fig:lidar_coverage}.

Numerous Deep Learning methods of object detection using LiDARs have shown promising results, such as VoxelNet \cite{voxelnet}, PointRCNN \cite{pointrcnn}, SECOND \cite{second}, and others. Low-latency inference and accurate detections are of the utmost importance for our use case of high-speed autonomous racing. For this reason, PointPillars \cite{pointpillars} serves as our primary detection method, capable of reliably detecting vehicles at ranges up to $100m$ away. The birds-eye-view projection and 2D convolutions used within PointPillars allow for the removal of computationally expensive and time-consuming sparse 3D convolutions performed by other LiDAR networks.

\begin{figure}[t!]
    \centering
    \includegraphics[width=\textwidth]{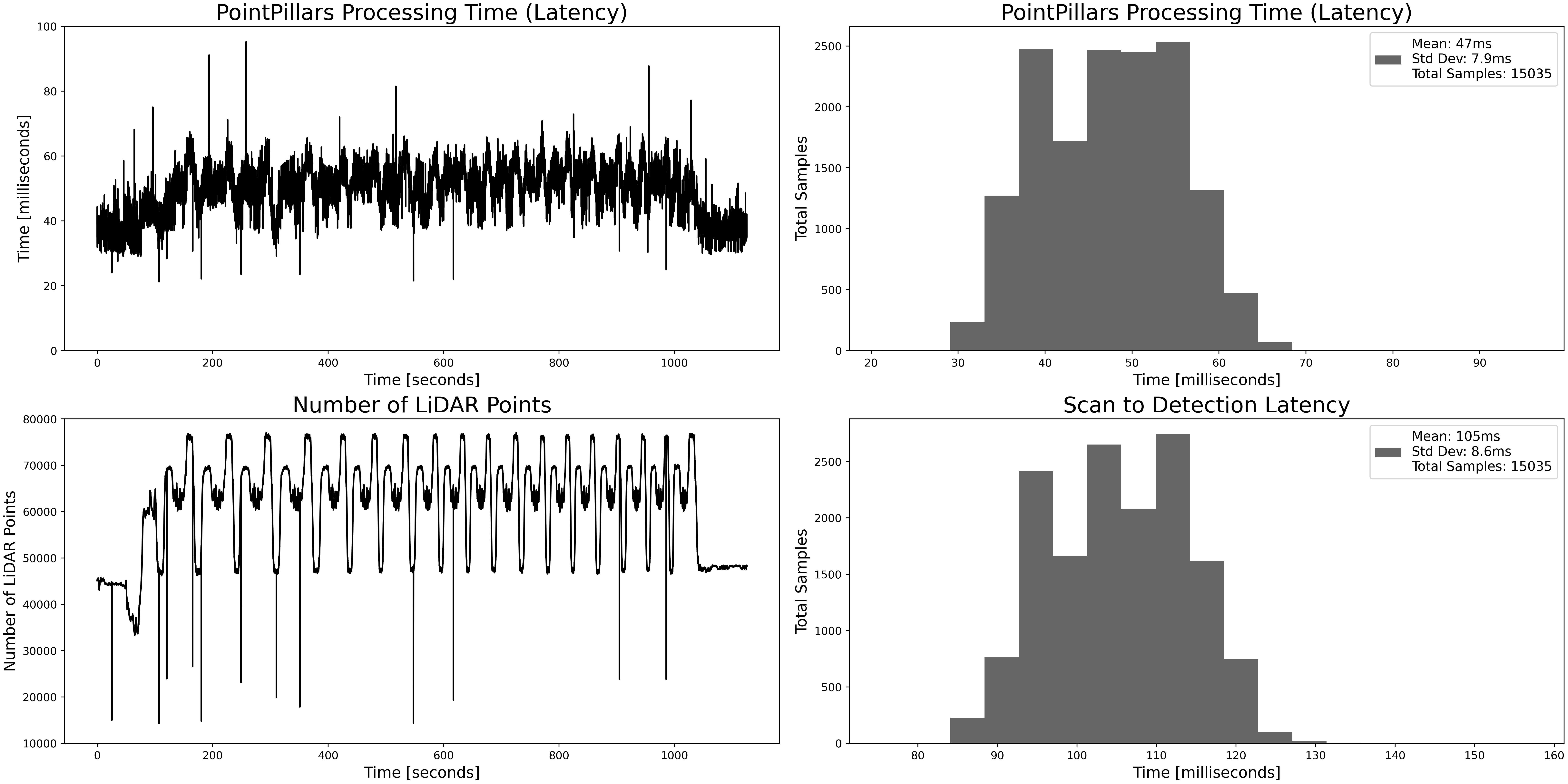}
    \caption{PointPillars Processing Time (latency) from the semi-finals match in Season Two in Las Vegas. \textbf{(Top Left)} Total processing of a single LiDAR scan, over the course of the run. \textbf{(Bottom Left)} Number of LiDAR points in each scan being processed. \textbf{(Top Right)} Histogram of processing time over the whole run. \textbf{(Bottom Right)}  Histogram of total time from the timestamp of the oldest point in the point cloud to when a detection result is sent downstream. The right histograms show that, on average, PointPillars processing time is about $47ms$ and the time from when the first LiDAR point is scanned until a detection is produced is on average $105ms$.}
    \label{fig:lidar_latency_good}
\end{figure}

Figure \ref{fig:lidar_latency_good} shows the outcomes of PointPillars. To reduce processing time, our PointPillars implementation is single-sweep, meaning we do not accumulate scans over time before running inference. Additionally, to further simplify the pipeline, inference is done directly on the raw scans, after down-sampling and applying a crop. We explicitly chose to not compensate for distortion caused by the ego vehicle's motion. Based on the data observed and practical considerations within the larger stack, motion compensation was deemed not worth the additionally complexity and processing time required. The scanning rate ($\sim50ms$ from top to bottom) is faster than other LiDARs, which results in less distortion. Additionally, the LiDAR data is only used for detection and the relative speed between agents is low enough that the error due to motion distortion can be ignored. The potential gains do not outweigh the additional latency. Finally, work had been done to implement distortion correction, but was removed due to integration and performance issues, which will be discussed further in Section \ref{section:evaluation}.

Our implementation of PointPillars uses many of the same underlying optimizations seen in \cite{BEV_FUSION}, such as heavy utilization of Torch Sparse \cite{tang2022torchsparse}, a high-performance neural network library, specializing in point cloud processing. This allows for fast and efficient inference time, even with a Python-based implementation.

\subsubsection{Data Collection, Labeling, \& Training}

\begin{table}[]
    \centering
         \begin{tabular}{|c|c|c|}
            \hline
            \textbf{Source} &  \textbf{Number of Labels} & \textbf{Percent}  \\
            \hline
                Simulation     & 6744 & $92.2\%$ \\
                Real Vehicle   & 570  & $7.8\%$ \\
            \hline
                \textbf{Total} & \textbf{7314} & \textbf{$100\%$} \\
            \hline
        \end{tabular}
        \caption{Breakdown of label sources for training the initial PointPillars model. Later iterations incorporated many more labels from the real vehicle. Having a better model produces better auto-labels, thereby speeding up the data collection and training process.}
        \label{tab:lidar_labels}
\end{table}

\begin{figure}
\centering
\includegraphics[width=.9\linewidth]{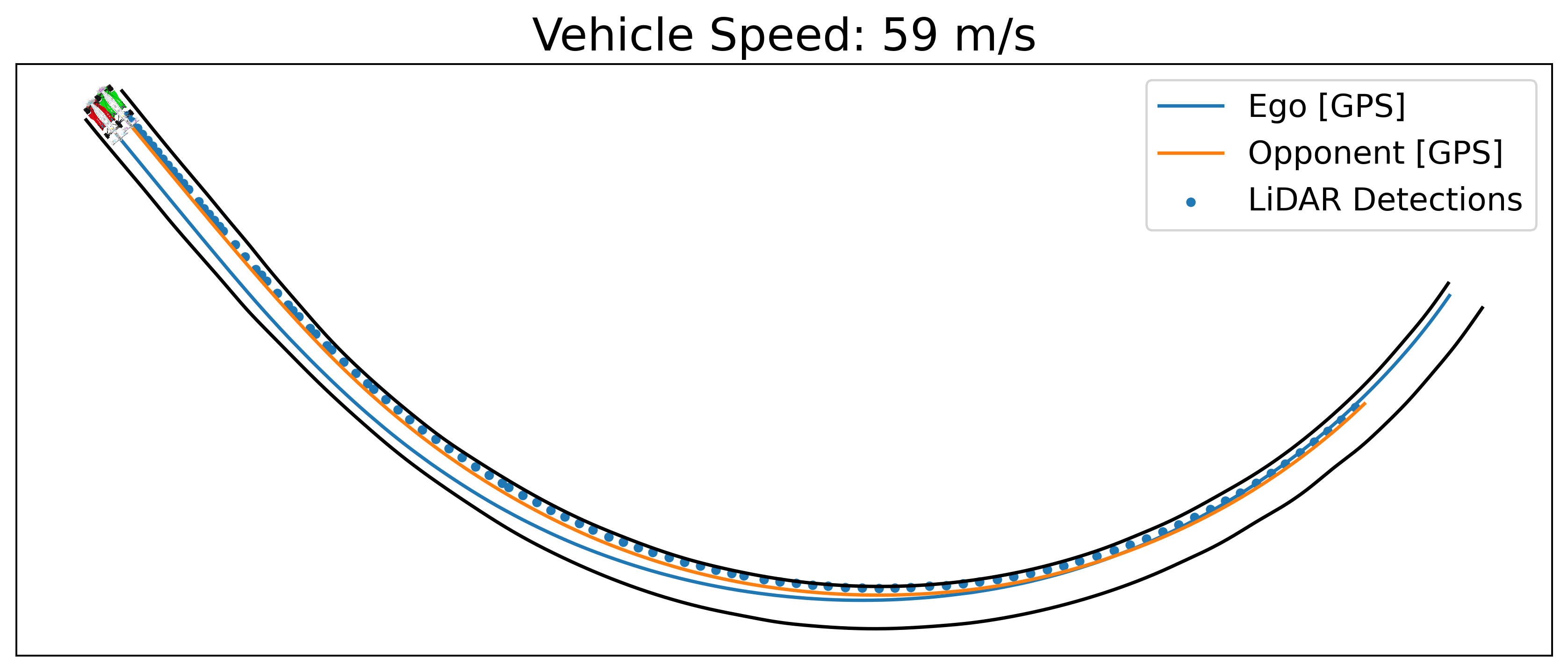}
    \caption{Comparison of PointPillars detections against the measured trajectory of the opponent ARV in Season Two at Las Vegas. The ego and opponent ARV positions are obtained using GPS with RTK corrections applied, which provides centimeter-level accuracy. Also shown are the track bounds (black) and the ego vehicle trajectory. Finally, this snapshot of the run was when the ego vehicle was traveling at over $59 m/s$, completing a pass of the opponent vehicle.}
\label{fig:lidar_results}
\end{figure}

No dataset exists that contains AV-21s racing head-to-head. Adequately training PointPillars required developing a large and robust dataset. Initially, data was collected in simulation, which helped developed an initial model. The first model dataset is broken down in Table \ref{tab:lidar_labels}. The simulation environment did not match the vehicle setup perfectly. In particular, while the range and coverage were similar, the point cloud was less dense than in real life. Interestingly, we found that the initial model trained off of this data transferred to detecting AV-21s on real data, especially at longer ranges, where the cloud is less dense. Figure \ref{fig:lidar_results} shows a comparison of PointPillars detections against the measured trajectory of an opponent ARV.

With an initial model, it was now possible to do ``auto-labeling", where the model is used to generate new labels that are then hand-verified by a human annotator. Because the model often provides a detection that is close to ground truth, the workload on the human annotator is reduced. Additionally, by using the existing model to label more data, labels can be focused on the areas where the model performed most poorly.

\subsubsection{Discussion: Strengths, Limitations, and Future Work }\label{cam_discussion_crash}

Given the prototypical nature of the AV-21 platform, the sensor plate must be disassembled every time the autonomy components need servicing. As a result, the extrinsic calibration between the cameras and the LiDARs changes frequently. This is less of an issue with the LiDARs, as they are all firmly fastened to the same aluminum plate. Additionally, the extreme operating conditions of the AV-21 platform (i.e. high speeds and accelerations) also necessitate re-calibrating the sensors regularly, even if the sensor plate has not been removed. Small extrinsic calibration errors can lead to very large projection errors, especially for distant objects. This problem is not exclusive to ARVs and is an active area of research \cite{aurora_calibration}\cite{calibration_survey}. Future work will center on streamlining the calibration process and developing systems that are less brittle to small errors.

\begin{figure}[h!]
\centering
\includegraphics[width=.8\linewidth]{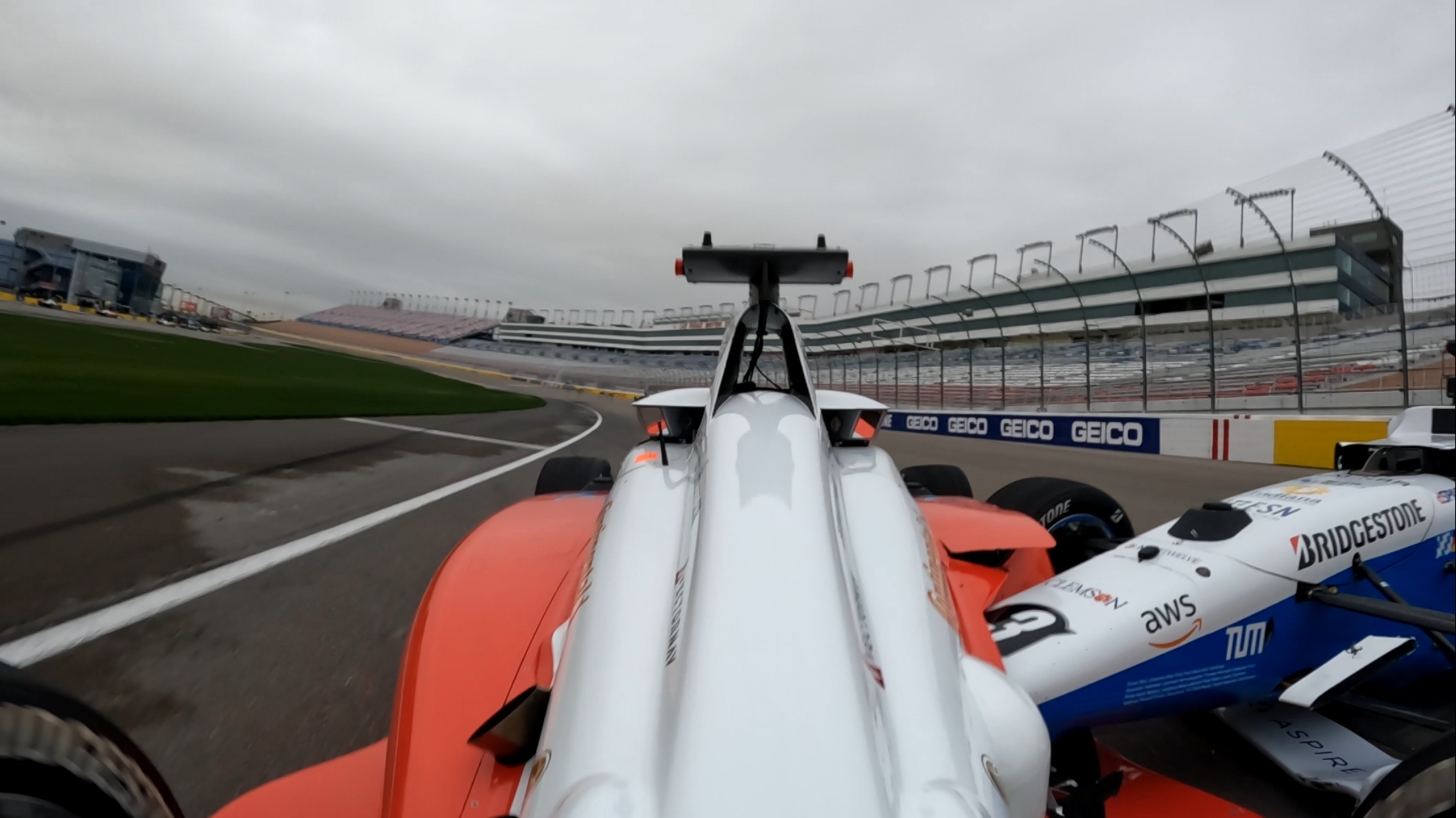}
   \caption{Onboard image from the first known major autonomous, multi-agent, head-to-head collision. The competition-style run had both vehicles acting as both Defender and Attacker. In this instant, the opponent AV-21 was attempting an overtake; however, when doing so, the vehicle commanded a very sharp acceleration and lost traction. This resulted in the vehicle spinning out of control and piercing the side of the AV-21. The resulting damage included a fried battery isolator, a destroyed alternator, torn timing belt, pierced side pod, several fuel injectors, a severed engine wiring loom, numerous severed fuel and oil lines, destroyed coolant tubes, and more. To repair the damage, the entire vehicle was disassembled, including removing the transmission and engine, then disassembling the cockpit and autonomy hardware suite. The vehicle was repaired and running again 48 hours later, in time for the competition the next day}.
\label{fig:crash_tum}
\end{figure}

Finally, due to a severe crash less than 72 hours before the competition in Season Two at Las Vegas, the camera detection pipeline was disabled for the competition events. An image from the footage of the crash, recorded by an onboard GoPro camera, can be seen in Figure \ref{fig:crash_tum}. With the focus being repairing the AV-21 vehicle, no time was available to properly calibrate the sensors and the potential for projection errors outweighed the benefits. Because of the modular design of the perception stack, it was trivial to make such a drastic change. In a fast-paced competition environment, this modularity and flexibility proved paramount in allowing the vehicle to operate during the competition. While the redundancy and peak performance of the stack was compromised, as shown later in Section \ref{section:evaluation}, the vehicle was still able to autonomously compete in three rounds, winning the first two, and losing the third after running out of fuel after attempting an overtake at over $150 mph$.


\subsection{Tracking}

\subsubsection{Challenges and Requirements}

Tracking within an ARV software stack serves to provide downstream tasks with a single belief of the states of other agents within the world. Different perception modalities capture different portions of a given agent's state space. For example, the monocular camera perception provides a noisy estimate of an agent's position, but cannot accurately predict its orientation. Our LiDAR perception produces full pose estimates of other agents, but currently does not infer the agent's velocity. While using only one of these detection methods will yield a belief that is severely limited by the outlined weaknesses, the effective fusion of both can result in each modality compensating for the drawbacks of the other.

Our implementation allows for the fusion of multiple sensing modalities in a straightforward manner, and serves to provide downstream planning tasks with the state of all perceived agents. Our decoupled approach to perception requires our tracking stack to meet the following requirements:

\begin{enumerate}
    \item Incorporate all modalities from perception, including LiDAR and monocular camera detections
    \item Estimate positions, velocities, and orientations in the world of all tracked agents
    \item Provide a precise and accurate state estimate of the opponent agents
    \item Provide a consistent measure of the uncertainty of the agents' state estimates
    \item Be robust to false positives, missed detections, and drop-outs from one or more sensor modalities
\end{enumerate}

Finally, tracking must perform all of the above while ensuring as little additional latency as possible, handling measurements from perception asynchronously and out of order, and compensating for any delay between sensor measurement and tracking.

\subsubsection{Overview of Approach}

\begin{figure}[h!]
    \centering
    \includegraphics[width=\textwidth]{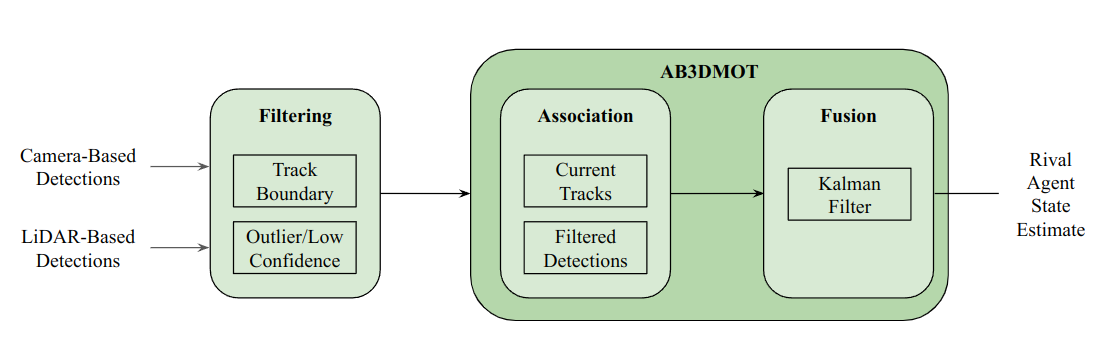}
    \caption{The Tracking pipeline architecture, from detection inputs to state estimate for a single opponent ARV. Our approach consists of three main components: Filtering, Association, and Fusion. Filtering removes outliers. Association determines whether or not a detection is of a previously seen agent. Fusion is incorporating new measurements of agents' states. }
    \label{fig:fusion_tracking_arch}
\end{figure}

Our approach consists of three main components: Filtering, Association, and Fusion. Filtering removes outliers. Association determines whether or not a detection is of a previously seen agent. Fusion is incorporating new measurements of agents' states. In order to minimize processing latency within the tracking stack, well-researched, efficient algorithms are leveraged for each module. Figure \ref{fig:fusion_tracking_arch} presents the Tracking pipeline architecture.

\subsubsubsection{\textbf{Filtering:}} Detections filtered by a confidence threshold, a hyper-parameter within our tracking stack, tuned empirically by analyzing the false positives and associated confidence produced by perception. Additionally, any detection that falls outside of the track bounds is ignored. The combination of these two filtering steps helps to ensure that only valid detections are processed and used to generate tracked agents.

\subsubsubsection{\textbf{Association:}} AB3DMOT \cite{AB3DMOT} provides the data association module utilized by tracking stack. By employing two computationally efficient algorithms, the Hungarian algorithm for data matching \cite{Hungarian} and the Kalman filter \cite{KF} for fusion and prediction, the authors demonstrate strong results on multiple open-source data sets while also providing high-frequency predictions. In practice, we observed that the Hungarian algorithm with Euclidean distance often resulted in poor data association, especially during temporary sensor drop-out. Therefore, our implementation uses simple greedy matching with the Mahalanobis distance~\cite{Mahalanobis}, which performed better in testing.

{\bf Track births and deaths:} To reduce the probability of false positives becoming valid tracks, a new potential track is instantiated (born) only after two detections (from successive sweeps) are associated with it. This hyper-parameter provides a means to balance between the quality and confidence of tracks and end to end latency in reacting to other agents. Finally, any tracks that have not had a detection associated with it within the last five seconds are also removed (killed) to prevent stale tracks from influencing future associations. 

\subsubsubsection{\textbf{Fusion:}} Once detections have been associated with an existing tracked agent, or have been repeatedly observed and classified as a new agent, we begin tracking the agent using fused multi-modal perception outputs. Again, we utilize a modified version of \cite{AB3DMOT} as the Kalman filter for performing sensor fusion. Since incoming detections from the camera perception pipeline have already been projected into a 3-dimensional position and transformed into a common frame, both LiDAR and camera measurements can be used to update the internal Kalman filter for a given tracked agent. In this way, sensor fusion becomes a simple task that can be asynchronous across the two modalities, and the states of tracked agents can be published at the receipt of each incoming detection.

\subsubsection{Discussion, Limitations, and Future Work}

 The tracking pipeline meets all requirements and is sufficiently accurately and performant to handle the IAC Passing Competition. Our modular design was especially important when the camera perception was disabled on race day, outlined in detail in Section \ref{cam_discussion_crash}. Given these successes, however, our system has not been robustly tested against multiple agents, specifically agents that are close together (i.e. $\leq5m$). In traditional motorsports, humans drive aggressively in close proximity to one another. While the competition format is far from this style of racing, future works will need to handle such operating domains in order to challenge professional drivers. Multiple agents in close proximity are more difficult to track, due to higher association ambiguity and occlusions.

Finally, a Kalman filter will be replaced by an Extended Kalman filter (EKF) to enable a non-linear motion model. With an EKF and a better motion model (i.e. constant curvature), predictions of agent tracks will be more accurate, which is especially important during periods of infrequent detections (i.e. the other agent is in the blind-spot produced by the rear wing of the vehicle).


\subsection{Planning}

\begin{figure}[h!]
\centering
\includegraphics[width=16cm,height=6cm]{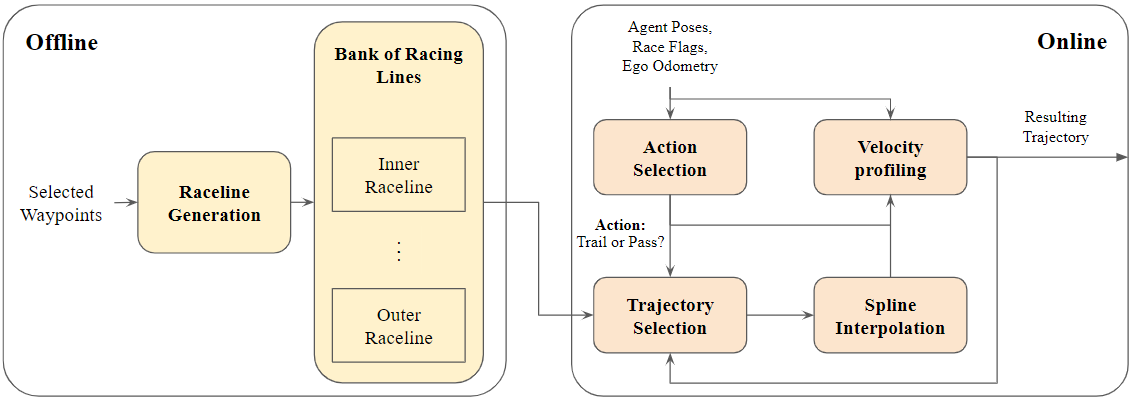}
   \caption{Planning is comprised of offline and online stages. Waypoints are chosen at critical points along the track within the offline portion. Next, a raceline generator fits a spline to these points to generate a trajectory. This is conducted for an arbitrary number of lines. A higher-level behavior module selects the best action to take in the online portion. This action incorporates the previous action, race state, and the locations of other agents. Given a passing or merge action, a trajectory is selected and interpolated from the current trajectory. Finally, this resulting trajectory is velocity profiled and passed onto controls.}
\label{fig:planning_flowchart}
\end{figure}

\subsubsection{Overview}

We developed a fast, modular motion planning stack capable of predicting agent behavior and safely trailing and passing other agents. Figure \ref{fig:planning_flowchart} shows a high-level overview of the motion planner. The path planner identifies the agent's position at points in the track where it is most optimal to pass and extrapolates its position in the following lap. We have based our path planning approach on a set of primitive behaviors from which higher-level strategies can select. These primitive behaviors include opponent trailing, raceline following, and lane switching. This modularity allows for a clean separation between high-level decision-making and trajectory selection and generation. 

We also take advantage of knowing the track geometry and develop a set of strong priors in the form of a trajectory bank. Offline, trajectories are generated for various lanes along the track, providing different levels of clearance from the inner and outer track boundaries. Online, the planner selects the best trajectory from the bank based on the selected action primitive.

\subsubsection{Offline Trajectory Generation}
The raceline generation is a two-step process that begins with defining a set of waypoints on the track. Afterward, splines are interpolated on these waypoints to ensure continuity. This process relies heavily on the manual selection of waypoints at apexes to properly leverage spline properties.

\begin{figure}[h!]
\centering
\includegraphics[width=.8\linewidth]{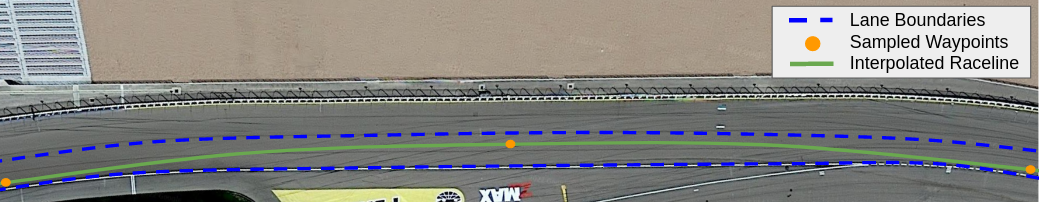}
  \caption{A visualization of the raceline generation for the inner lane.}
\label{fig:inner_raceline_generation}
\end{figure}

For the overtaking competition, waypoints were selected manually from two lanes. We consider a lane to be defined as a path with both an inner and outer boundary and derive a center-line equidistant from the two boundaries. We obtain these two lanes by dividing the width of the track in half along the full length. Within these lanes, the process of manual sampling begins. A series of $N$ waypoints, $(q_1, ... , q_N)$, are used to interpolate closed Spiro splines \cite{Levien09}. Figure \ref{fig:inner_raceline_generation} displays waypoint selection and the interpolated raceline for the inner lane.

While the tracking controller is compatible with any reasonably smooth raceline, the Spiro spline family is well-suited for several reasons. Firstly, Spiro splines maintain $G^4$-continuity, meaning that the second derivative of curvature is continuous. In addition, Spiro splines are an efficient approximation of the Minimum Variation Curve (MVC), which minimizes the integral of curvature rate \cite{Levien09}, corresponding to minimizing the steering effort of the vehicle. By selecting waypoints at apexes, the above properties make a raceline generated from a Spiro spline effective at minimizing downstream instability, reducing steering effort, and resulting in faster lap times and smoother driving than naively selecting waypoints alone. Future work includes incorporating the Spiro spline representation within a larger optimization, such as the work in \cite{tum_min_curv} and \cite{tum_min_time}, to eliminate the need for manual waypoint selection and ensure time-optimal trajectories.

\subsubsection{Online Action Selection}
The multi-agent passing competition sets two roles for the competitors: "defender" and "attacker". As defined in Section \ref{section:competition}, the attacker passes the defender that is maintaining a raceline within the inside lane of the track (see Figure \ref{fig:competition_rules}). The attacker is also responsible for maintaining a safe distance from the defender. We define a set of Action Primitives to encode these maneuvers:

\begin{itemize}
    \item \textbf{Maintain}: \\
        Maintain the current raceline at a given speed.
    \item \textbf{Trail}: \\
        Maintain a fixed distance behind the opponent vehicle on the current raceline.
    \item \textbf{Safe Merge}: \\
        Merge between arbitrary lanes safely (i.e., avoid collisions with other agents).
\end{itemize}

Selection of the primitives is dependent on the current role (defender or attacker) and the current Track Condition (i.e., Green, Yellow, Red, Waving Green), which are both defined by flags sent to the vehicle from Race Control.

\textbf{Attacker:}
\begin{itemize}
    \item Under a Green Flag, the Attacker must close the gap with the defender (\textbf{Trail})
    \item Under a Waving Green Flag, the Attacker may initiate the pass (\textbf{Safe Merge})
    \item Under a Waving Green Flag and after the Attacker has passed the opponent by at least $30m$, the Attacker must ``Close the Door" by merging back to the inside lane (\textbf{Safe Merge})
    \item Under a Green Flag, the Attacker has the freedom to take any lane, but may not begin passing until explicitly allowed (\textbf{Trail} or \textbf{Safe Merge})
\end{itemize}

\textbf{Defender:}
\begin{itemize}
    \item Under a Green Flag, the Defender must move to the inside lane and maintain the speed set for that round (\textbf{Maintain})
\end{itemize}

While this logic is simple for the passing competition, the use of these primitives can scale to more complex logic. For example, rewards and costs could be assigned to each of these primitives which are then utilized by a search-based planner or a reinforcement-learning algorithm to estimate expected rewards and costs by taking a set of actions and to determine the best solution given the scenario. By structuring the planner around a modular set of action primitives, we can explore multiple solutions to the behavioral decision-making problem very easily. Additionally, it is easy to add more primitives in the future when necessary.

\subsubsection{Online Raceline Merging}
\begin{figure}[h!]
\centering
\includegraphics[width=.8\linewidth]{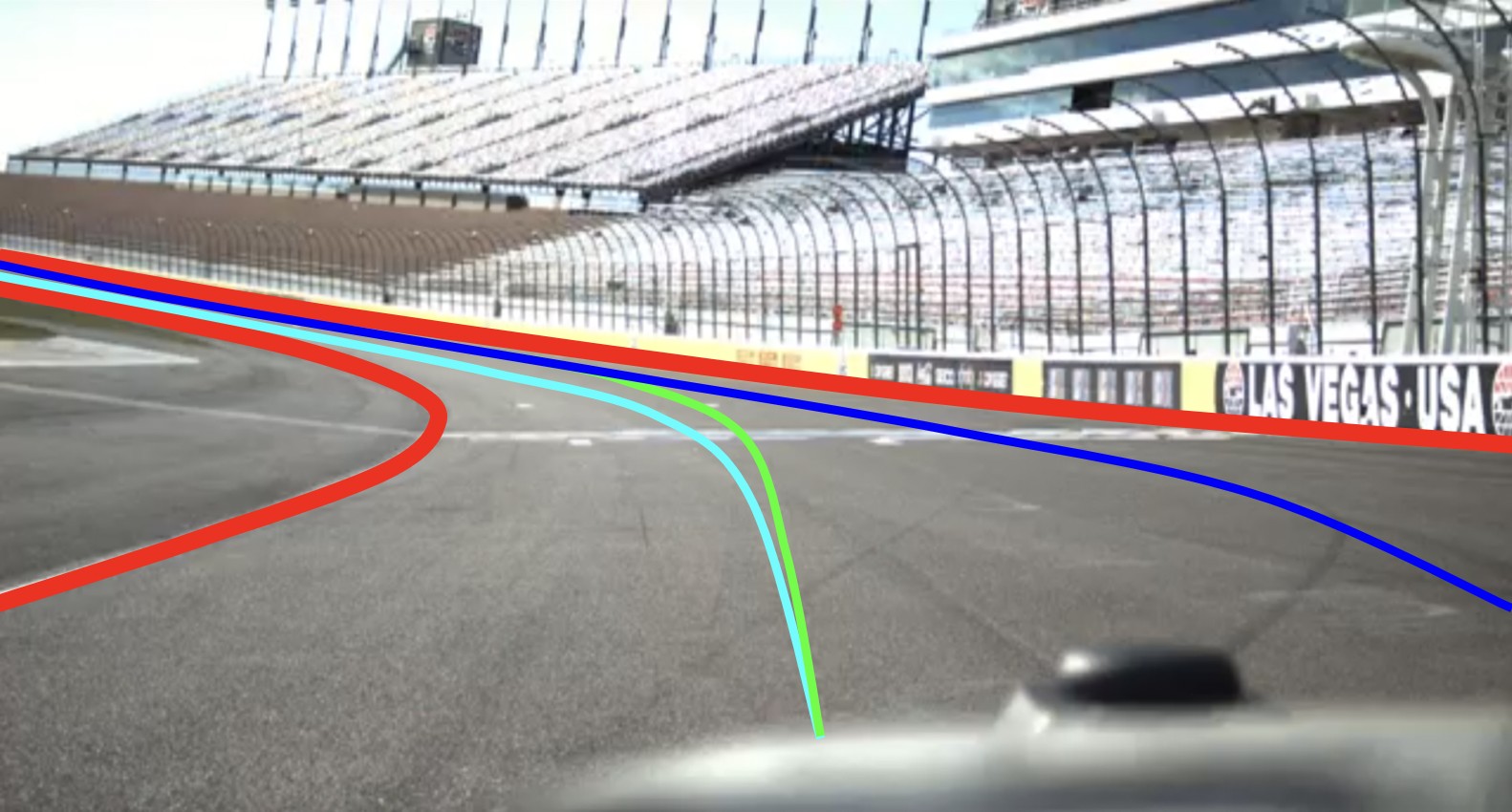}
  \caption{Visualization of planning strategy. Red lines represent the wall boundaries. The light blue line represents the current trajectory on the inner raceline. Dark blue represents the outer raceline the vehicle will merge to for passing the opposing agent. Green line represents the resulting safe-merge trajectory.}
\label{fig:raceline}
\end{figure}

Merging between inner and outer racelines (i.e. ``Safe merging") is accomplished by first calculating the closest pose on the inner raceline $q_k^{in}$ corresponding to every pose on the outer raceline $q_k^{out}$. Next, given a start and a final interpolation pose, $q_k^{in}$ and $q_k^{out}$, as well as the corresponding twist, the spline optimization formulation in Section 3.2 of \cite{Veerapaneni_minjerk} is used to generate a minimum jerk merging trajectory. Time intervals $h_k$ between consecutive time stamps can be decreased as desired to generate a sufficiently smooth raceline between $q_k^{in}$ and $q_k^{out}$. This interpolation can be seen in Figure \ref{fig:raceline}.

\subsubsection{Results and Discussion}
Figure \ref{fig:planning_flowchart} displays a comprehensive flowchart of the mentioned approach and highlights the offline and online computations. Tasks performed by individual components of the planner are agnostic to the actions performed by other components. This includes adding or removing additional raceline primitives as well as action primitives, making this architecture highly modular. Future work on the planner includes automating the process of waypoint selection based on given inside and outside lines, which would consider varying track curvature. Finally, this approach is very fast, efficient, and able to easily execute at $20Hz$ (and capable of a much higher execution rate). Action selection, safe-merge trajectory interpolation, and velocity profiling are all very inexpensive to compute online.


\subsection{Controls}

\subsubsection{Overview of Approach} \label{controls_overview}

\begin{figure}[h!]
\centering
\includegraphics[width=.8\linewidth]{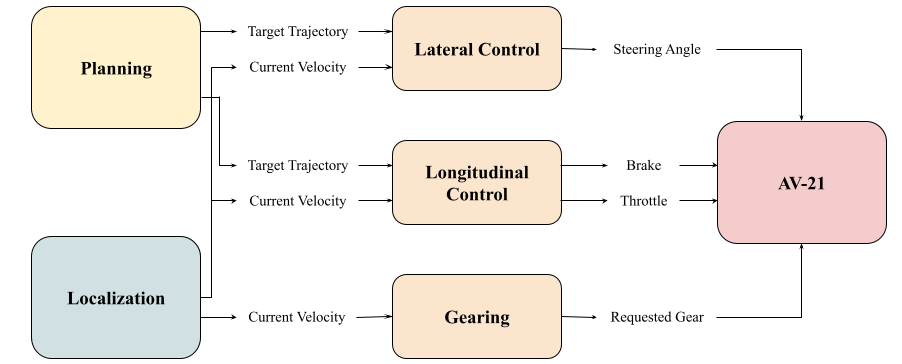}
   \caption{Controller Stack Overview}
\label{fig:stack}
\end{figure}

Our controls architecture was designed for speed and modularity. To achieve this, we decompose our control task of path tracking into three tasks - lateral control, longitudinal control, and gear selection. Lateral tracking generates a steering angle such that the vehicle converges to the path. The longitudinal controller maintains a particular longitudinal velocity. The gear-shifting controller is responsible for selecting the optimal gear for maintaining the current vehicle speed. The majority of the content in this section will focus on the lateral tracking element of the vehicle's controller. The architecture of the stack is described in Figure \ref{fig:stack}. Finally, the work presented here is a continuation of our previous work in \cite{ICRA_WS_CONTROL}.
    
\textbf{Longitudinal Control and Gear Shifting}
The throttle and brake control utilizes a simple P control scheme paired with a feed-forward term to account for drag. This is expressed within Algorithm \ref{alg:long} line \ref{alg:pid_line}, which returns a command value. This command is then interpreted as either a throttle or brake command depending on whether the command is positive or negative, using a scaling factor on the brake signal to account for differences in magnitude between the throttle and brake. Smoothing on the throttle and brake signals prevents instantaneous acceleration or deceleration to help prevent vehicle instability. Algorithm \ref{alg:long} summarizes this algorithm.

The gear-shifting strategy is based on a simple lookup table where the optimal gear is computed given the current velocity of the vehicle. This gearing table was generated based on a model of the engine and track parameters to maximize torque, the computation of which is outside the scope of this paper.

\begin{algorithm}
\caption{Longitudinal Tracking Algorithm}\label{alg:long}
$k_p, k_{\text{feed forward}}, \alpha_{\text{brake}}, \delta_{\text{throttle}}, \delta_{\text{brake}} \gets \text{loadParams}()$\;
$\text{throttle\_previous}, \text{brake\_previous} \gets 0$\;
\While{$true$}{
    $x, y, \dot x, \dot y, \psi, \dot \psi \gets \text{getState}()$\;
    $v_{\text{target}} \gets \text{getTarget}()$\;
    $\text{command} \gets k_p (v_{\text{target}} - \dot x) + k_{\text{feed forward}} * v_{\text{target}}$\; \label{alg:pid_line}
    $\text{throttle}, \text{brake} \gets 0$
    \eIf{$\text{command} \geq 0$}{
        $\text{throttle} \gets \text{command}$\;
    }{
        $\text{brake} \gets - \alpha_{\text{brake}}\text{command}$\;
    }
    $\text{throttle} \gets \text{smooth}(\text{throttle\_previous}, \text{throttle}, \delta_{\text{throttle}})$\;
    $\text{brake} \gets \text{smooth}(\text{brake\_previous}, \text{brake}, \delta_{\text{brake}})$\;
    $\text{comamnd}(\text{throttle}, \text{brake})$\;
    $\text{throttle\_previous} \gets \text{throttle}$\;
    $\text{brake\_previous} \gets \text{brake}$\;
}
\end{algorithm}

\textbf{Lateral LQR Control}
The lateral controller is built around a Linear-Quadratic Regulator (LQR) that generates an optimal feedback policy based on a nominal vehicle dynamics model. LQR was chosen because (1) it is inexpensive and fast to compute the desired controls online and (2) it is optimal given the vehicle dynamics model. However, one downside is that LQR does not reason about actuation constraints and only considers the error concerning the reference at the current time step. However, the vehicle primarily drives on an oval race track where only a few degrees of steering are required even for the most aggressive maneuvers. Hence, LQR can stably control the vehicle and meet the reference tracking requirements to safely avoid the other agent, track barriers, and maintain the desired trajectory.

Given a continuous-time linear system of the form in Equation \ref{continuous_linear} a quadratic cost function such as Equation \ref{cost_function} can be defined which is constrained by Equation \ref{continuous_linear}. Equation \ref{cost_function} also has the constraints that $Q = Q^T \geq 0$ and $R = R^T \geq 0$.

An optimal feedback policy can be derived such that the cost function in Equation \ref{cost_function} is minimized over an infinite time horizon. Solving this problem is trivial and can be done using Matlab's \textit{lqr} function as well as many other available solvers such as ones detailed in \cite{lqr_solver_comparison}. This resulting feedback policy takes the form of Equation \ref{opt_feedback}.
\begin{equation}\label{continuous_linear}
    \dot x(t) = A x(t) + B u(t)
\end{equation}
\begin{equation}\label{cost_function}
    J=\int_{t=0}^{\infty} \left[ x(t)^T Q x(t) + u(t)^T R u(t) \right]dt
\end{equation}
\begin{equation}\label{opt_feedback}
    u^*(t) = K(t)x(t)
\end{equation}

\begin{figure}[h!]
\centering
\includegraphics[width=.85\linewidth]{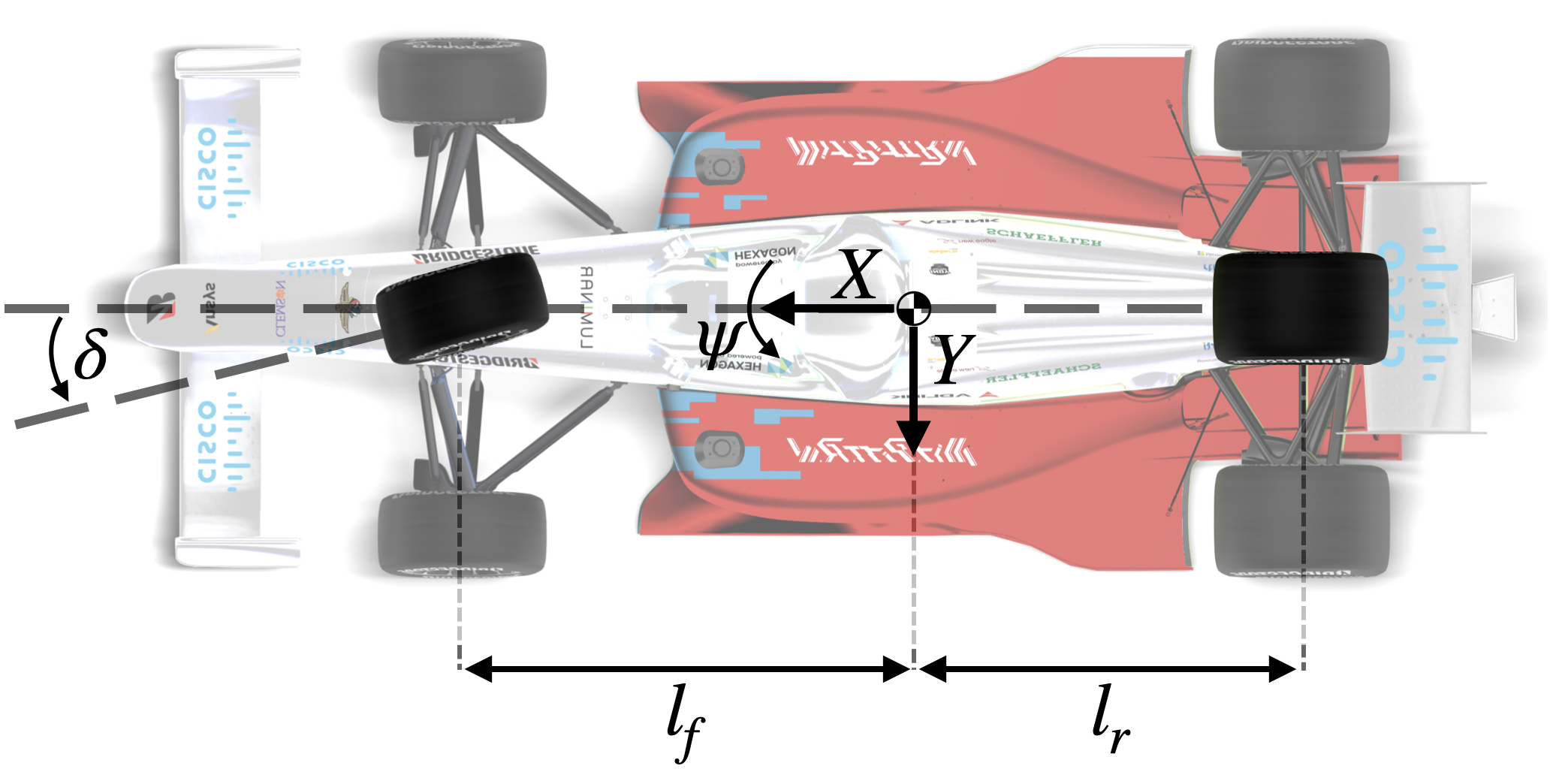}
   \caption{Vehicle Coordinate System}
\label{fig:coordinate_system}
\end{figure}

\textbf{Dynamic Bicycle Model}    
We utilize a four-state dynamic bicycle model from \cite{Rajamani_2005}, shown in Equations \ref{eq:dbm1} and \ref{eq:dbm2}. As noted in Section \ref{controls_overview}, we decompose the problem into separate lateral and longitudinal controllers. The lateral controller reasons about the vehicle's lateral position $y$ and yaw angle $\psi$ in the Local Tangent Plane (LTP) frame provided by localization. The input for this system is the steering angle $\delta$. It also assumes some constant velocity $V_x$, which is the forward component of the speed ignoring lateral slip. The $C_{\alpha f}$ and $C_{\alpha r}$ parameters are the cornering stiffness of the front and rear tires respectively, which reflect the ability of the tires to resist deformation while the vehicle corners. In addition, $l_f$ and $l_r$ represent the length of the vehicle from the center of gravity to the front and rear axles respectively. The parameter $I_z$ is the scalar moment of inertia about the z-axis, and $m$ is the mass of the vehicle. The coordinate system is displayed in Figure \ref{fig:coordinate_system} and vehicle parameters are summarized in Table \ref{tab:model_parameters}.

\begin{table}[h!]
\centering
     \begin{tabular}{|c|c|c|}
        \hline
        \textbf{Symbol} & \textbf{Parameter} & \textbf{Units} \\
        \hline
            $V_x$ & Vehicle Speed & m/s \\
        \hline
            $m$ & Vehicle Mass & m \\
        \hline
            $I_z$ & Moment of Inertia About Z Axis & kg $ m^2$ \\
        \hline
            $C_{\alpha f}$ & Cornering Stiffness of Front Tires & N/rad \\
        \hline
            $C_{\alpha r}$ & Cornering Stiffness of Rear Tires & N/rad \\
        \hline
            $l_f$ & Distance from Center of Gravity to Front Axle  & m\\
        \hline
            $l_r$ & Distance from Center of Gravity to Rear Axle  & m\\
        \hline
    \end{tabular}
    \caption{Dynamic Bicycle Model Parameters}
    \label{tab:model_parameters}
\end{table}

\begin{equation}\label{eq:dbm1}
    \begin{bmatrix}
        \dot y \\ \ddot y \\ \dot \psi \\ \ddot \psi 
    \end{bmatrix} = \begin{bmatrix}
        0 & 1 & 0 & 0 \\
        0 & -\frac{2 C_{\alpha f} + 2 C_{\alpha r}}{m V_x} & 0 & -V_x - \frac{2 C_{\alpha f} l_f - 2 C_{\alpha r} l_r}{m V_x} \\
        0 & 0 & 0 & 1 \\
        0 & -\frac{2 l_f C_{\alpha f} - 2 l_r C_{\alpha r}}{I_z V_x} & 0 & -\frac{2 l_f^2 C_{\alpha f} + 2 l_r^2 C_{\alpha r}}{I_z V_x}
    \end{bmatrix} \begin{bmatrix}
        y \\ \dot y \\ \psi \\ \dot \psi 
    \end{bmatrix} + \begin{bmatrix}
        0 \\
        \frac{2 C_{\alpha f}}{m} \\
        0 \\
        \frac{2 l_f C_{\alpha f}}{I_z}
    \end{bmatrix} \delta
\end{equation}

To use this dynamics model with the LQR formulation above, we re-parameterize the state space as the lateral error ($e_1$) and yaw error ($e_2$) with respect to a target point along the raceline provided by planning $\begin{bmatrix} x^* & y^* & \psi^* & \dot \psi^* \end{bmatrix}$ as well as the vehicle position in the LTP frame provided by localization $\begin{bmatrix} x & y & \dot x & \dot y & \psi & \dot \psi \end{bmatrix}$:
\begin{equation}\label{error_terms}
    \begin{bmatrix}
        e_1 \\
        \dot e_1 \\
        e_2 \\
        \dot e_2
    \end{bmatrix} = \begin{bmatrix}
        (x^* - x) \sin({-\psi^*}) + (y^* - y) \cos({-\psi^*}) \\
        \dot y + \dot x (\psi - \psi^*) \\
        \psi - \psi^* \\
        \dot \psi - \dot \psi^*
    \end{bmatrix}
\end{equation}

With this, we can derive the error dynamics for which we solve the LQR problem to get the optimal stabilizing controller gain $K$ to drive the error to zero.

\begin{equation}\label{eq:dbm2}
     \begin{bmatrix}
        \dot e_1 \\ \ddot e_1 \\ \dot e_2 \\ \ddot e_2
    \end{bmatrix} = \begin{bmatrix}
        0 & 1 & 0 & 0 \\
        0 & -\frac{2 C_{\alpha f} + 2 C_{\alpha r}}{m V_x} & \frac{2 C_{\alpha f} + 2 C_{\alpha r}}{m} & - \frac{2 C_{\alpha f} l_f - 2 C_{\alpha r} l_r}{m V_x} \\
        0 & 0 & 0 & 1 \\
        0 & -\frac{2 l_f C_{\alpha f} - 2 l_r C_{\alpha r}}{I_z V_x} & \frac{2 l_f C_{\alpha f} - 2 l_r C_{\alpha r}}{I_z} & -\frac{2 l_f^2 C_{\alpha f} + 2 l_r^2 C_{\alpha r}}{I_z V_x}
    \end{bmatrix} \begin{bmatrix}
        e_1 \\ \dot e_1 \\ e_2 \\ \dot e_2
    \end{bmatrix} + \begin{bmatrix}
        0 \\
        \frac{2 C_{\alpha f}}{m} \\
        0 \\
        \frac{2 l_f C_{\alpha f}}{I_z}
    \end{bmatrix} \delta
\end{equation}

This model makes several assumptions, including small angle assumptions on the steering angle, which hold well on ellipsoidal tracks where we command a maximum steering angle of approximately $0.1$ [rad]. The model also assumes constant velocity. A reformulation to a linear time-varying model could allow for varying velocity. However, we accept that assumption and mitigate it with a series of feedback controllers derived at different velocities for simplicity. For greater detail on this formulation of the dynamics see \cite{Rajamani_2005}.
    
\textbf{Pure Pursuit, LQR Tracking Algorithm}
Given the model formulation and the LQR formulation, we can make a lateral tracking algorithm by combining a pure-pursuit style look-ahead point and a feedback mechanism similar to Equation \ref{opt_feedback} generated using LQR. The look-ahead point is queried simply by looking for the point on the raceline provided by planning that is ahead of the vehicle a given distance $d$ away. LQR requires an $A$, $B$, $Q$, and $R$ matrix to generate the feedback policy. The $A$ and $B$ matrix can be obtained using the dynamics from Equations \ref{eq:dbm1} and \ref{eq:dbm2}. The $Q$ and $R$ matrices can be obtained empirically where the diagonal $Q$ matrix defines the weights on the error terms in Equation \ref{error_terms}, and the scalar $R$ matrix defines the weight on the steering angle that acts as a dampening factor to prevent over-steering.

\begingroup
\renewcommand{\arraystretch}{1.2} 
\begin{table}[h!]
\centering
     \begin{tabular}{|c|c|c|}
        \hline
        \textbf{Velocity (m/s)} & \textbf{Q} & \textbf{R} \\
        \hline
            [0-10) & $\begin{bmatrix}
                    Q^{10}_{(0,0)} & 0 & 0 & 0\\
                    0 & Q^{10}_{(1,1)} & 0 & 0 \\
                    0 & 0 & Q^{10}_{(2,2)} & 0 \\
                    0 & 0 & 0 & Q^{10}_{(3,3)}
                    \end{bmatrix}$  & $\begin{bmatrix}
                                        R^{10}_{(0,0)} \\
                                        \end{bmatrix}$ \\        
        \hline
            [10-20) & $\begin{bmatrix}
                    Q^{20}_{(0,0)} & 0 & 0 & 0\\
                    0 & Q^{20}_{(1,1)} & 0 & 0 \\
                    0 & 0 & Q^{20}_{(2,2)} & 0 \\
                    0 & 0 & 0 & Q^{20}_{(3,3)}
                    \end{bmatrix}$  & $\begin{bmatrix}
                                        R^{20}_{(0,0)} \\
                                        \end{bmatrix}$ \\ 
        \hline
            [20-25) & $\begin{bmatrix}
                    Q^{25}_{(0,0)} & 0 & 0 & 0\\
                    0 & Q^{25}_{(1,1)} & 0 & 0 \\
                    0 & 0 & Q^{25}_{(2,2)} & 0 \\
                    0 & 0 & 0 & Q^{25}_{(3,3)}
                    \end{bmatrix}$  & $\begin{bmatrix}
                                        R^{25}_{(0,0)} \\
                                        \end{bmatrix}$ \\ 
        \hline
            . & . & . \\ 
            . & . & . \\
            . & . & . \\
        \hline
            [55-60) & $\begin{bmatrix}
                    Q^{60}_{(0,0)} & 0 & 0 & 0\\
                    0 & Q^{60}_{(1,1)} & 0 & 0 \\
                    0 & 0 & Q^{60}_{(2,2)} & 0 \\
                    0 & 0 & 0 & Q^{60}_{(3,3)}
                    \end{bmatrix}$  & $\begin{bmatrix}
                                        R^{60}_{(0,0)} \\
                                        \end{bmatrix}$ \\ 
        \hline
    \end{tabular}
    \caption{LQR controller speed brackets. By tying controller parameters to the vehicle's current velocity, better performance and stability can be achieved. Varying the $Q$ matrix adjusts the relative weight to the yaw \& lateral errors and their derivatives. The $R$ gain provides a penalty on control, thereby adjusting how reactive the controller is. Empirically, these "knobs" have proven effective for tuning the vehicle to be stable while also tracking effectively enough to be within our desired safety margins.}
    \label{tab:controller_tuning_parameters}
\end{table}
\endgroup
    
A different parameter set $p = (Q^{v}_{(0,0)}, Q^{v}_{(1,1)}, Q^{v}_{(2,2)}, Q^{v}_{(3,3)}, R^{v}_{(0,0)})$ is utilized depending on the current velocity $v$ of the vehicle as shown in Table \ref{tab:controller_tuning_parameters}. Offline, the velocity bracket ranges $[\dot x_{\text{low}}, \dot x_{\text{high}})$ are defined to cover the range of velocities the vehicle is expected to cover, or generally the range $[0, \infty)$ with no overlap between them.

The speed brackets serve two purposes: first, it provides linearization points across a range of velocities to account for the constant velocity assumption in the dynamics model. Secondly, as the vehicle reaches higher speeds, the expectation is that the dynamics model mismatch is greater. Additionally, it is preferred that the controller is less reactive at higher speeds. By varying the $R$ gain, we can tune the controls to be less aggressive as we reach higher speeds. By varying the $Q$ matrix, we can provide different emphases on the various components of our error, i.e. lateral \& yaw deviations and their derivatives. Empirically, we have found that, at higher speeds, the vehicle is more stable when the yaw \& yaw rate error consists of a higher overall emphasis in the resulting control output than the lateral deviations. Finally, the pure-pursuit look-ahead distance is also varied as a function of speed where $d = d_{\text{base}} + k_{v,d} \dot x$. These parameters provide numerous knobs for a controls engineer to rapidly tune the controller to achieve the desired performance.

The LQR tracking algorithm runs as follows. First, it initializes by loading in all velocity brackets and the associated parameters. It then generates the feedback policies $K$ for each bracket using the average velocity of the bracket. Note that if one of the bounds is $\infty$, the lower velocity is utilized. Next, at each time step, given the current state of the vehicle and a raceline, it performs a look-ahead query on the raceline to get the goal point. Using this goal point and the current state, an error state can be generated. Finally, with the current vehicle velocity, it queries the velocity brackets for the relevant feedback policy which is applied to the error state to get the optimal steering angle. This algorithm is summarized in Algorithm \ref{alg:tracking}. 
    
\SetKwComment{Comment}{/* }{ */}
\begin{algorithm}
\SetAlgoLined
\caption{Lateral Tracking Algorithm}\label{alg:tracking}
$P \gets \{(v_{1, \text{low}}, v_{2, \text{low}}, K_1), \hdots, (v_{n, \text{low}}, \infty, K_n)\}$ 

\Comment*[r]{We assume parameters have been loaded and the feedback policies generated.}
$d_{\text{base}}, k_{v,d} \gets \text{loadParams}()$\;
\While{$true$}{
    $x, y, \dot x, \dot y, \psi, \dot \psi \gets \text{getState}()$\;
    $\tau \gets \text{getTrajectory}()$\;
    $d \gets d_{\text{base}} + k_{v,d} * \dot x$\;
    $K \gets \text{getFeedbackMatrix}(\dot x, P)$\;
    $x^*, y^*, \psi^*, \dot \psi^* \gets \text{lookahead}(x, y, d, \tau)$\;
    $e \gets \text{errorMatrix}$\ $(x, y, \dot x, \dot y, \psi, \dot \psi, x^*, y^*, \psi^*, \dot \psi^*)$\;
    $u \gets -K e$\;
    $\text{commandSteering}(u)$\;
}
\end{algorithm}

\subsubsection{Results and Discussion}

\begin{table}[]
    \centering
         \begin{tabular}{|c|c|c|c|c|c|c|}
            \hline
            \textbf{Date} &  \textbf{$>10m/s$} & \textbf{$>60m/s$} & \textbf{$55-60m/s$} & \textbf{$50-55m/s$} & \textbf{$45-50m/s$} & \textbf{$40-45m/s$}  \\
            \hline
                12-30-2021            & 0.323 & 0.679 & 0.529 & 0.479 & 0.412 & 0.394  \\
                01-03-2022            & 0.244 & 0.540 & 0.457 & 0.573 & 0.456 & 0.403  \\
                01-07-2022            & 0.555 & 0.885 & 1.135 & 1.025 & 0.883 & 0.745  \\
                \textbf{Average}      & \textbf{0.372} & \textbf{0.689} & \textbf{0.832} & \textbf{0.653} & \textbf{0.519} & \textbf{0.476} \\
            \hline
            \textbf{Date} & & \textbf{$35-40m/s$} & \textbf{$30-35m/s$} & \textbf{$25-30m/s$} & \textbf{$20-25m/s$} & \textbf{$10-20m/s$} \\
            \hline
                12-30-2021            & & 0.413 & 0.260 & 0.281 & 0.060 & 0.048 \\
                01-03-2022            & & 0.275 & 0.279 & 0.248 & 0.057 & 0.074 \\
                01-07-2022            & & 0.733 & 0.587 & 0.512 & 0.210 & 0.100 \\
                \textbf{Average}      & & \textbf{0.420} & \textbf{0.382} & \textbf{0.287} & \textbf{0.065} & \textbf{0.079} \\
            \hline
        \end{tabular}
        \caption{Cross-Track Error [m] over several speed brackets for various high-speed runs. The first column provides an average error calculated over the entire run. The bottom row provides an average of the error across all three runs for that bracket.}
        \label{tab:cte_runs}
\end{table}

\begin{figure}[h!]
\centering
\includegraphics[width=.9\linewidth]{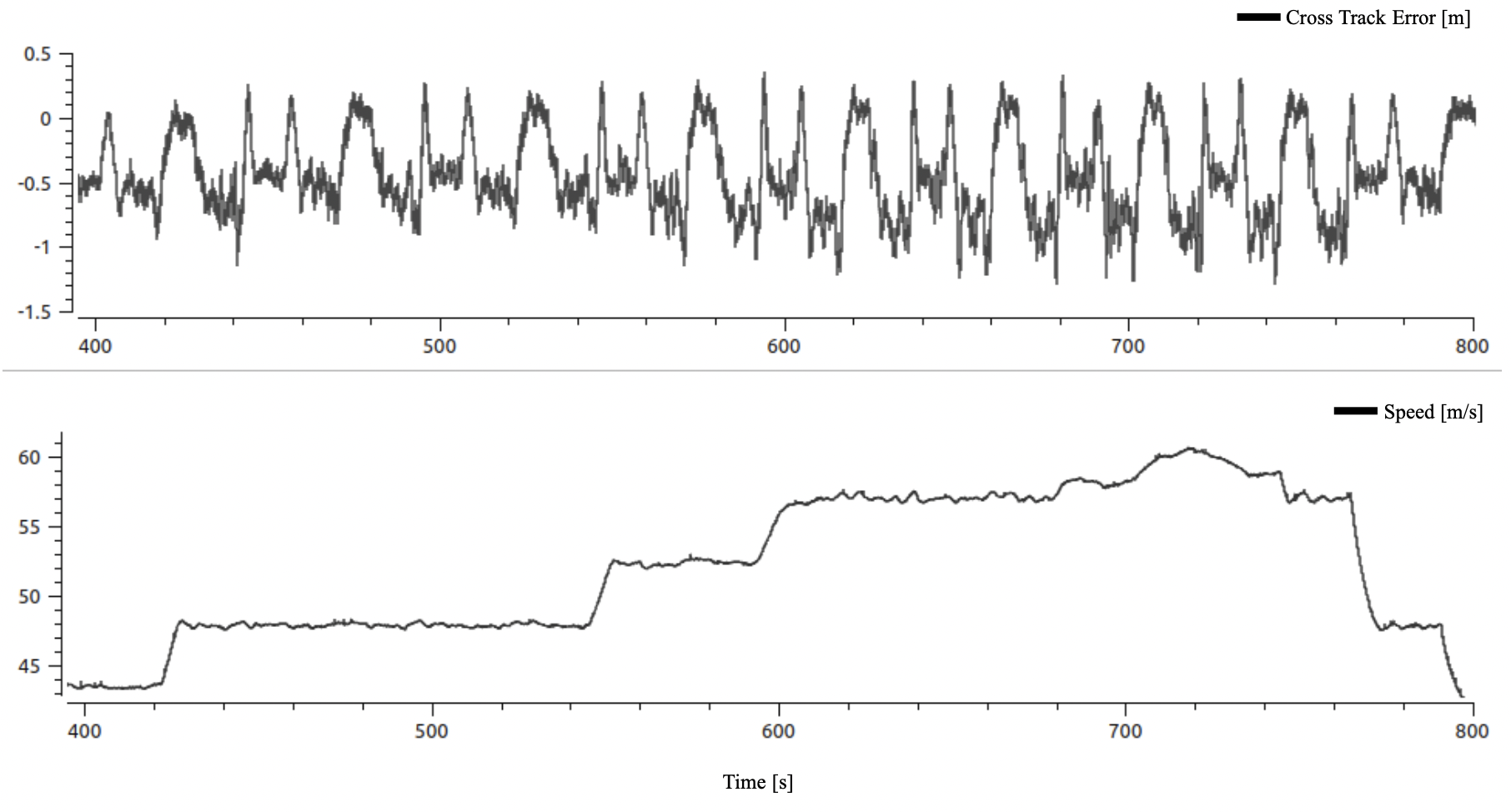}
  \caption{Controller performance from Season One. Top plot: cross-track error (CTE). Lower plot: Vehicle Velocity. The controller averaged between $0.5-1.5m$ CTE while traveling between $45-60m/s$. The vehicle was only allowed to command a max of $60\%$ throttle, which is why the vehicle speed is not flat at approximately 700 seconds into the run. }
\label{fig:cte_plots}
\end{figure}

The controller stack shown in Figure \ref{fig:stack} was evaluated over several performance laps at the LVMS with velocities ranging between $25m/s$ and $60.5m/s$. We experienced the worst performance at the highest speed targeting $60m/s$. We plot velocity and cross-track error (CTE) for the portion of a run in which speeds of $138mph$ were achieved in Figure \ref{fig:cte_plots}. The maximum CTE experienced was $1.3m$, which occurred around bends, while the mean absolute CTE was $0.323m$. Additionally, the average CTE for several high-speed runs ($138mph$, $141mph$, and $137mph$ max speeds) at various speed brackets are presented in Table \ref{tab:cte_runs}. Notably, for the run on January  $7^{th}$, the controller was tuned to be less aggressive, which explains the degraded tracking.

\begin{figure}[h!]
\centering
\includegraphics[width=.9\linewidth]{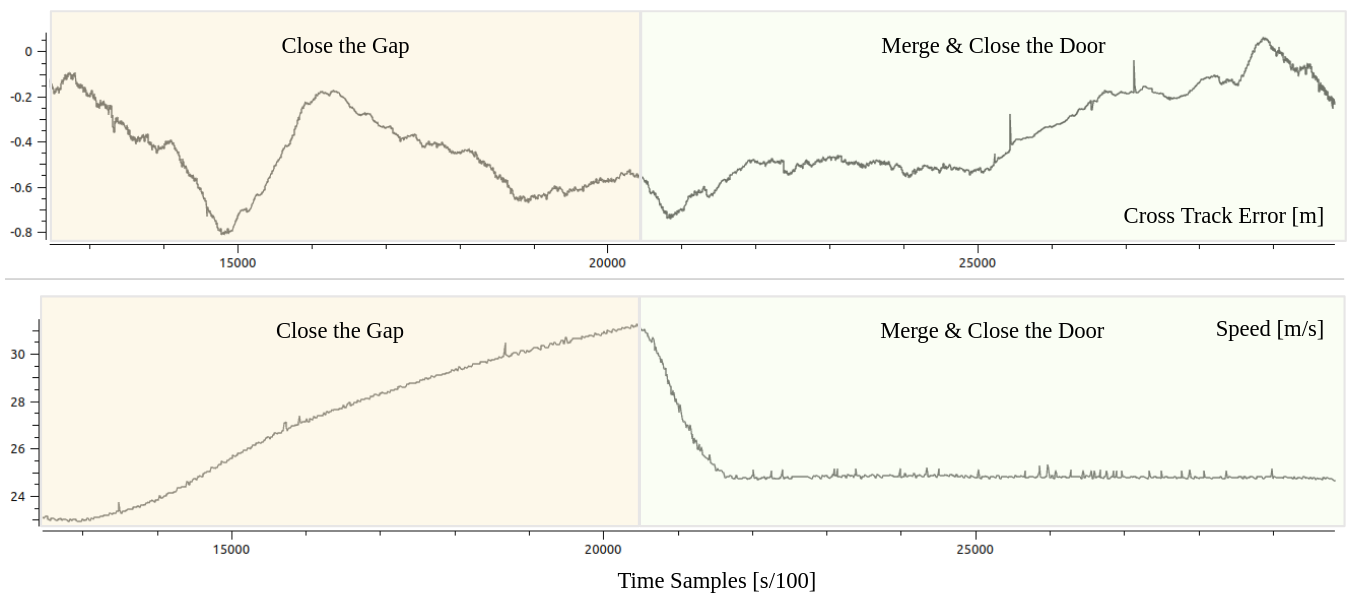}
  \caption{Controller performance results from season one. Top plot: Cross-Track Error (CTE). Lower plot: Vehicle Velocity. In this scenario, the ego vehicle is closing the gap and catching up to the other agent, a ``ghost" vehicle. Once caught up, the planner generates a smooth merging trajectory that returns to the inside line.}
\label{fig:merge_close_door_plots}
\end{figure}

\begin{figure}[h!]
\centering
\includegraphics[width=\linewidth]{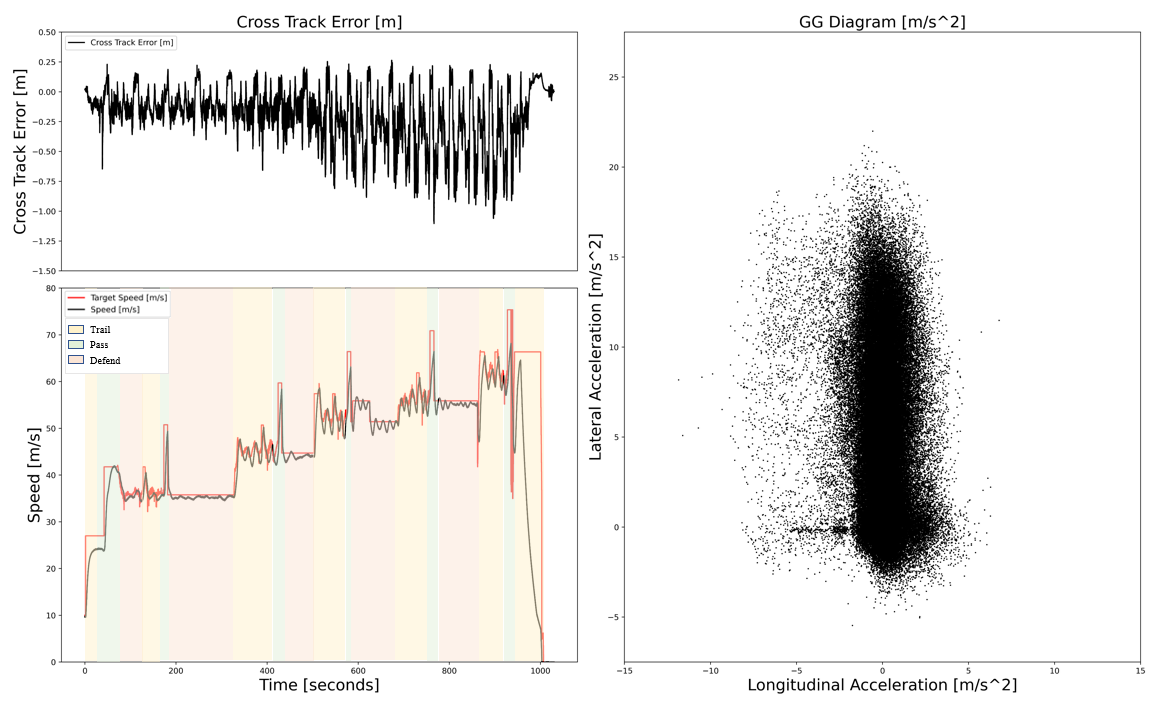}
  \caption{Controller performance results from season two. Top left plot: Cross-Track Error (CTE). Lower left plot: Vehicle Velocity. Right plot: G-G diagram. This scenario depicts the performance of the controller during the semi-final round of the passing competition in Las Vegas in Season Two, 2023.}
\label{fig:season2_controller_plots}
\end{figure}

We also evaluate the controller on lane-change tasks on the same track. Results from Season One are shown in Figures \ref{fig:merge_out} \ref{fig:merge_close_door_plots}. Season Two results are show in Figure \ref{fig:season2_controller_plots}. The Season One results are from running the planner and controller in the loop on the vehicle, but the opposing agent was simulated, referred to as a ``ghost" agent, as the vehicle will react to something that is not physically present. Results from the semi-finals match in Season Two, a real AV-21 multi-agent passing competition event, are also presented. Overall, the controller works very well in both seasons, with the CTE during the lane change maneuvers, staying between approximately $0.5-1.5m$. In the Season Two match, the vehicle reached lateral accelerations over $20m/s^2$.


\subsection{Localization and State Estimation}
\subsubsection{Challenges}

\begin{table}[h!]
\centering
     \begin{tabular}{|c|c|c|}
        \hline
        \textbf{Measurement Source} & \textbf{State} & \textbf{Update Rate} \\
        \hline
            BESTPOS & $<x, y, z>$ & $20Hz$ \\
        \hline
            BESTVEL & $<\dot{x},\dot{y},\dot{z},\psi>$ & $20Hz$ \\
        \hline
            HEADING2 & $<\theta>$ & $1Hz$ \\ 
        \hline
            IMU \& Gyro & $<\ddot{x}, \ddot{y}, \ddot{z}, \dot{\theta}>$ & $125Hz$ \\
        \hline
            Wheel Speed (x4) & $<\dot{x},\dot{y}>$ & $100Hz$\\
        \hline
    \end{tabular}
    \caption{Localization Measurement Sources}
    \label{tab:local_sources}
\end{table}

To localize itself on the track, the AV-21 is equipped with two dual-antenna NovAtel PwrPak7D-E1 GNSS units with RTK, as well as individual wheel speed sensors (Table \ref{tab:local_sources}). However, the high speed and high acceleration nature of the competition necessitated a custom solution that is robust to partial and total failures. In particular, a solution was needed that could meet the following requirements, all while traveling at high speeds:

\begin{itemize}
    \item Detect partial and full failures of either unit finding an adequate GNSS solution
    \item Handled degraded position and/or heading estimates from one or both units
    \item Accurately estimate the vehicle pose, yaw, and velocity at $100Hz$, despite only receiving GNSS measurements at $20Hz$
\end{itemize}

\subsubsection{Overview of Approach}

\begin{figure}[h!]
     \centering
         \centering
         \includegraphics[width=.7\linewidth]{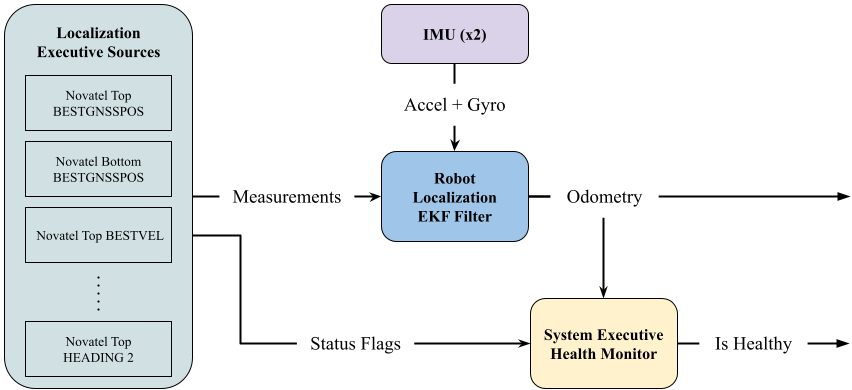}
         \caption{Overview of localization pipeline. All raw solutions are first pre-processed by the Localization Executive. The transformed and filtered measurements are then fused by an open-source Extended-Kalman Filter (EKF) package. Finally, health status flags are communicated to the System Executive, which triggers a safety response if needed.}
         \label{fig:local_overview}
\end{figure}


The Localization module is split into two main components: the Localization Executive and the Robot Localization \cite{Rl_EKF} EKF Filter node. Figure \ref{fig:local_overview} shows an overview of the localization stack. On the AV-21 are dual NovAtel PowerPak Dual-Antenna RTK GNSS systems with built-in IMU and Gyro. The NovAtels are configured to provide the best position ($20Hz$, velocity, and heading solutions). These measurements, from the varying sources, undergo the following:
\begin{itemize}
    \item Transformed into a Local Tangent Plane (LTP) coordinate frame, if applicable
    \item Solution status flags, variance measurements, and other health indicators are tested against a set of heuristics to verify that the measurement is good and worth fusing
    \item Finally, if the measurement meets quality checks, the measurement is converted to a standard ROS message type and sent to the filter for fusion
\end{itemize}

\begin{figure}[t!]
\centering
\includegraphics[width=.95\linewidth]{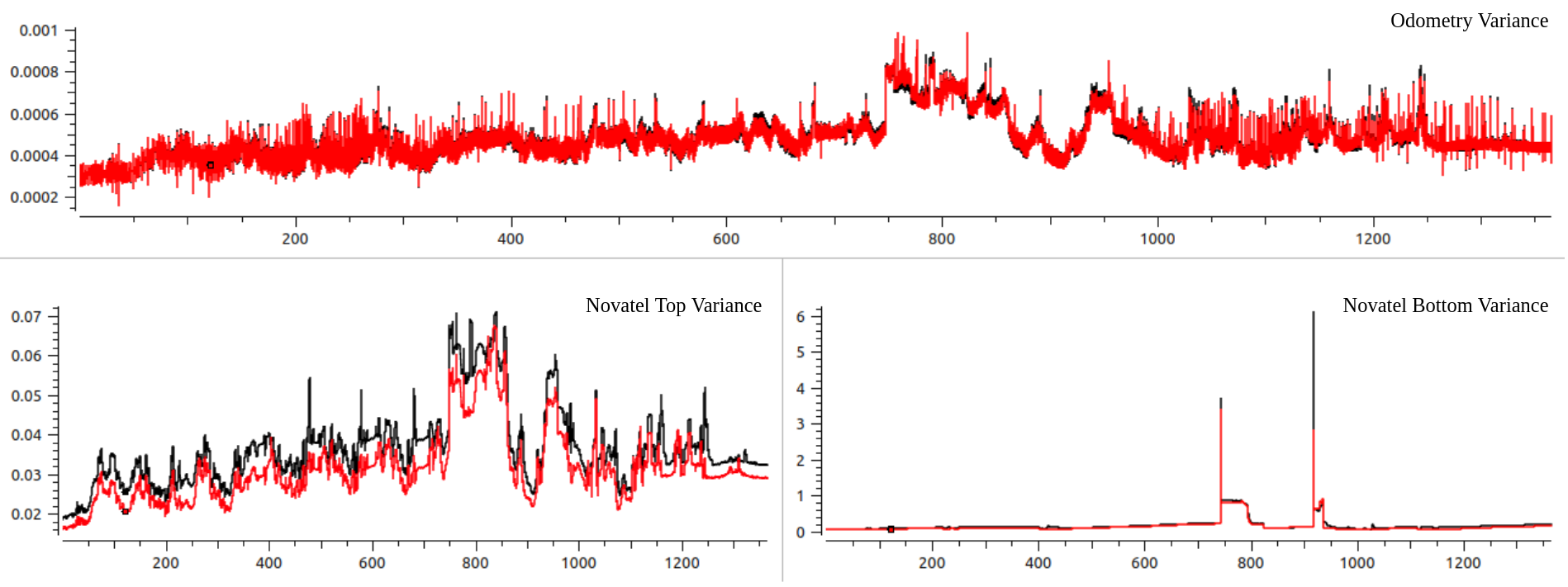}
   \caption{Localization handling degraded GPS measurements from one unit. The bottom graphs show variance (Red: Longitude, Black: Latitude) in pose measurements from each GNSS unit. Top denotes variance in fused pose (Red: y, Black: x). Note that at approximately 750 seconds into the run, the bottom GNSS unit starts reporting a much higher variance, measured in meters. However, despite the degraded sensor, the localization stack ignores this source and continues fusing the other unit. Once the unit has recovered, localization can fuse both again and has full redundancy.}
\label{fig:gps_bad}
\end{figure}

The node responsible for all of the above is called the Localization Executive. When building the Localization Executive, it was important to have a flexible, generic, and modular framework for defining sources, safety thresholds, and defining safety checks and heuristics. As testing progressed and more track time was put onto the vehicle, we quickly learned and adapted our fusion strategies. For example, the quality of the NovAtel with its dual antennas on the top and front nose of the vehicle often produces an overall better heading and position result than the other unit, likely due to a wider baseline. The Localization Executive's modularity allows us to prioritize high quality and filter out bad measurements on a per-source (i.e. pose, heading, velocity) basis. An example of a degraded source can be seen in Figure \ref{fig:gps_bad}.

The resulting transformed and filtered measurements are then processed by Robot Localization \cite{Rl_EKF}, an open-source EKF package that provides a flexible and modular interface for fusing an arbitrary number of odometry sources. Robot Localization was chosen because 1) it is very well tested and maintained in the robotics and unmanned vehicles community, 2) its modular design provides extreme flexibility during development and testing, and 3) it works well and exposes several tuning parameters and control, while able to update its state at $100Hz$ and fuse all incoming measurements asynchronously, including with the ability to handle out of order measurements.

Finally, health status flags for the various sources and final odometry results are communicated to the System Executive, the high-level arbiter of the stack. If any of the following scenarios occur, the vehicle is brought to an immediate controlled stop:

\begin{itemize}
    \item Incoming measurements, from both units, do not meet a strict update rate requirement or cut out completely
    \item Total connection loss from either GPS unit
    \item Loss of satellites or highest solution status, from both units
    \item Loss of RTK for a long enough time that the variance in the measurements exceeds safety thresholds, for both units
    \item Fused odometry covariance exceeds the safety threshold
\end{itemize}

If one unit is healthy, or if a combination of pose, heading, and velocity can be constructed from measurements from both units, then the stack will continue operating normally. Additionally, all health status flags are also communicated to the base station and presented in the interface, which is presented in Section \ref{section:basestation}. If partial, but not disastrous, failures occur, it is the responsibility of the base station operator decide whether to terminate the run or not.



\subsection{System Monitoring (Base Station)}\label{section:basestation}

\begin{figure}[h!]
\centering
\includegraphics[width=.8\linewidth]{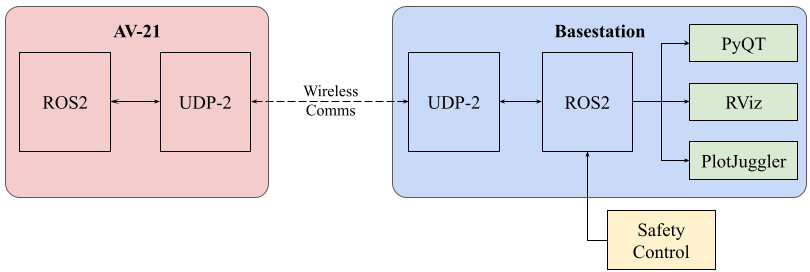}
  \caption{Base station architecture. A UDP server and client on the AV-21 and the basestation communicate via a lossy wireless connection, so network load must be managed to avoid overwhelming the network and only transmitting the essential bits of information.}
\label{fig:basestation}
\end{figure}

Developing an ARV stack requires monitoring software to ensure safe operations during high-speed testing and live operation during races. In this section, we describe the {\em base station}, our multi-level user interface for displaying the safety statuses of the vehicle and cumulatively all parts of the stack. The role of the interface is to be able to see the entire overall health of the vehicle at a glance. If any sensor or service goes offline, a quick look at the base station will clearly show what and where the problem is. There are primary and secondary user interfaces, each serving a specific role and level of surveillance over the stack and physical vehicle. Each interface is modular, meaning the operator can set what telemetry, track, and sensor data they want to see. There are no competition requirements for either interface, but it is a useful tool for the operators in the pit lane to be able to monitor the full health of the system at a glance.

Like all parts of the stack, the base station architecture, shown in Figure \ref{fig:basestation}, was developed with performance and reliability in mind. Existing solutions to rebroadcast ROS 2 DDS topics over the track's mesh network did not work reliably over the lossy wireless connection. High spikes in network load resulted in a large number of dropped packets, impacting the operator's ability to communicate with the vehicle. Our solution was built on top of a UDP Server-Client solution shared by the TUM Autonomous Motorsports. A raw UDP stream provided precise control over how much bandwidth is utilized for telemetry and management of quality of service. The telemetry nodes on the AV-21 and the basestation aggregate the desired information (i.e. vehicle odometry, joystick override commands, etc.) and transmit the resulting telemetry packet using UDP, thereby reducing network traffic. Once a telemetry packet is received by the basestation, it is parsed and then published back to ROS 2 for convienient integration with numerous open source visualizations tools, including RViz and PlotJuggler, and a custom PyQT-based user interface.

\subsubsection{Primary Interface}

\begin{figure}[h!]
\centering
\includegraphics[width=.9\linewidth]{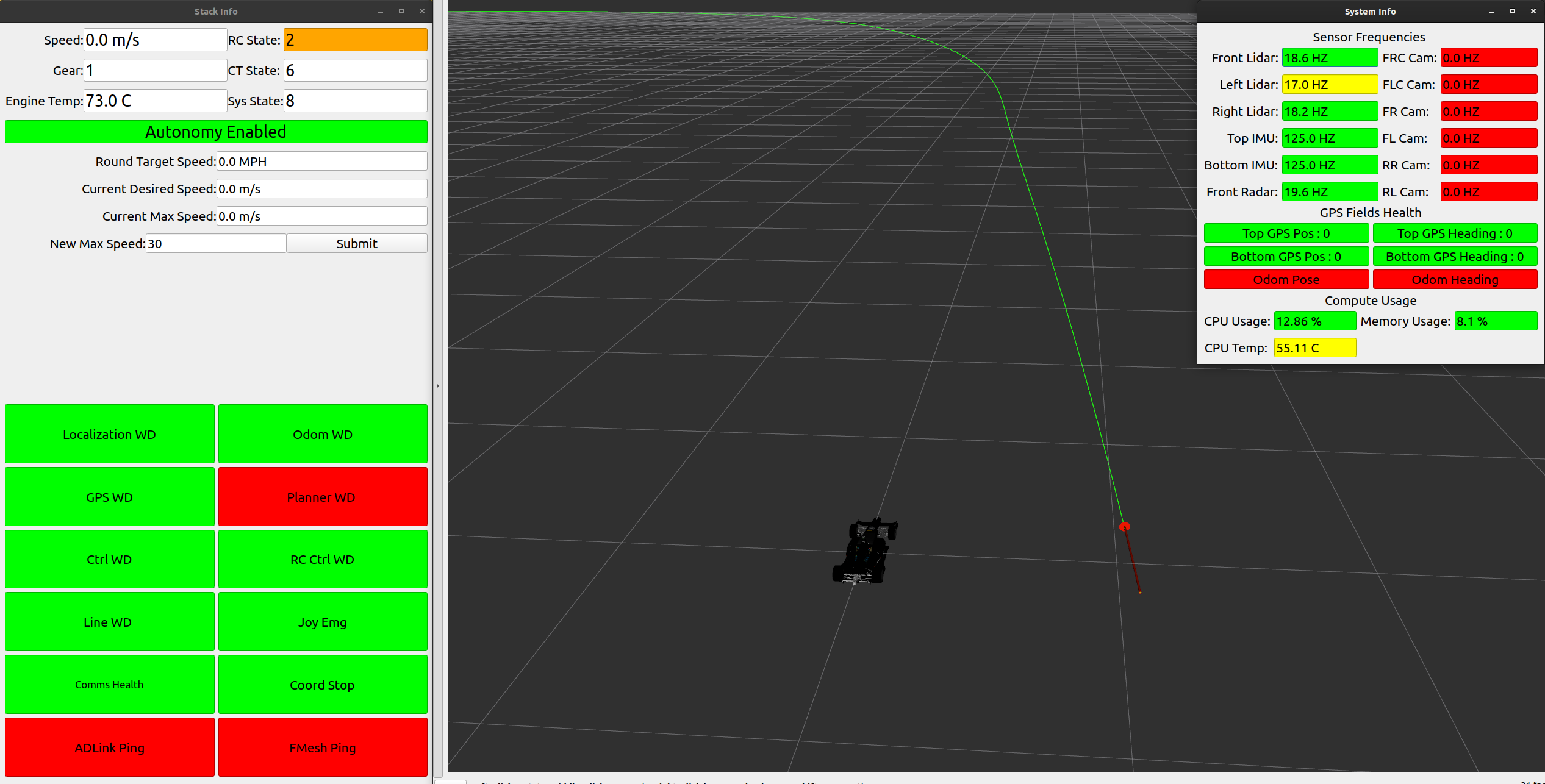}
\caption{Primary interface for the Base Station}
\label{fig:primary_ui}
\end{figure}

The primary interface displays track states, vehicle states and speeds, watchdog states, sensor frequencies, GPS health, and the computer's compute utilization. Figure \ref{fig:primary_ui} shows the display for the primary interface, made up of a GUI using PyQT and an odometry display using RViz. Using PyQT, we can easily add or modify panel modules on the GUI to fit our area of concentration for a test run or competition run. With RViz, we can visualize the path the vehicle is following, the current vehicle odometry, and any agents we have detected and are tracking. RViz is easily configurable for what topics and data we want to see. This interface is meant primarily to be an information display and not to give base station operators control over vehicle functions.

Available telemetry information includes the state of the power train, vehicle position and speed, and target speed. This information informs the operator if the vehicle is in a healthy state overall and what is it trying to do. The CPU and memory usage is also displayed to ensure the computer onboard is running smoothly. The watchdogs serve as surveillance agents over the stack, indicating if any component of the stack fails or crashes. When a watchdog fails, the stack will intelligently bring the car to a safe stop. All sensor frequencies are displayed, indicating if there are any current or imminent problems within the perception or localization stacks. Colors vary based on different safety thresholds (i.e. ``good", ``warning", ``bad"). GPS health information is also displayed.

\subsubsection{Secondary Interface}

\begin{figure}[h!]
\centering
\includegraphics[width=.9\linewidth]{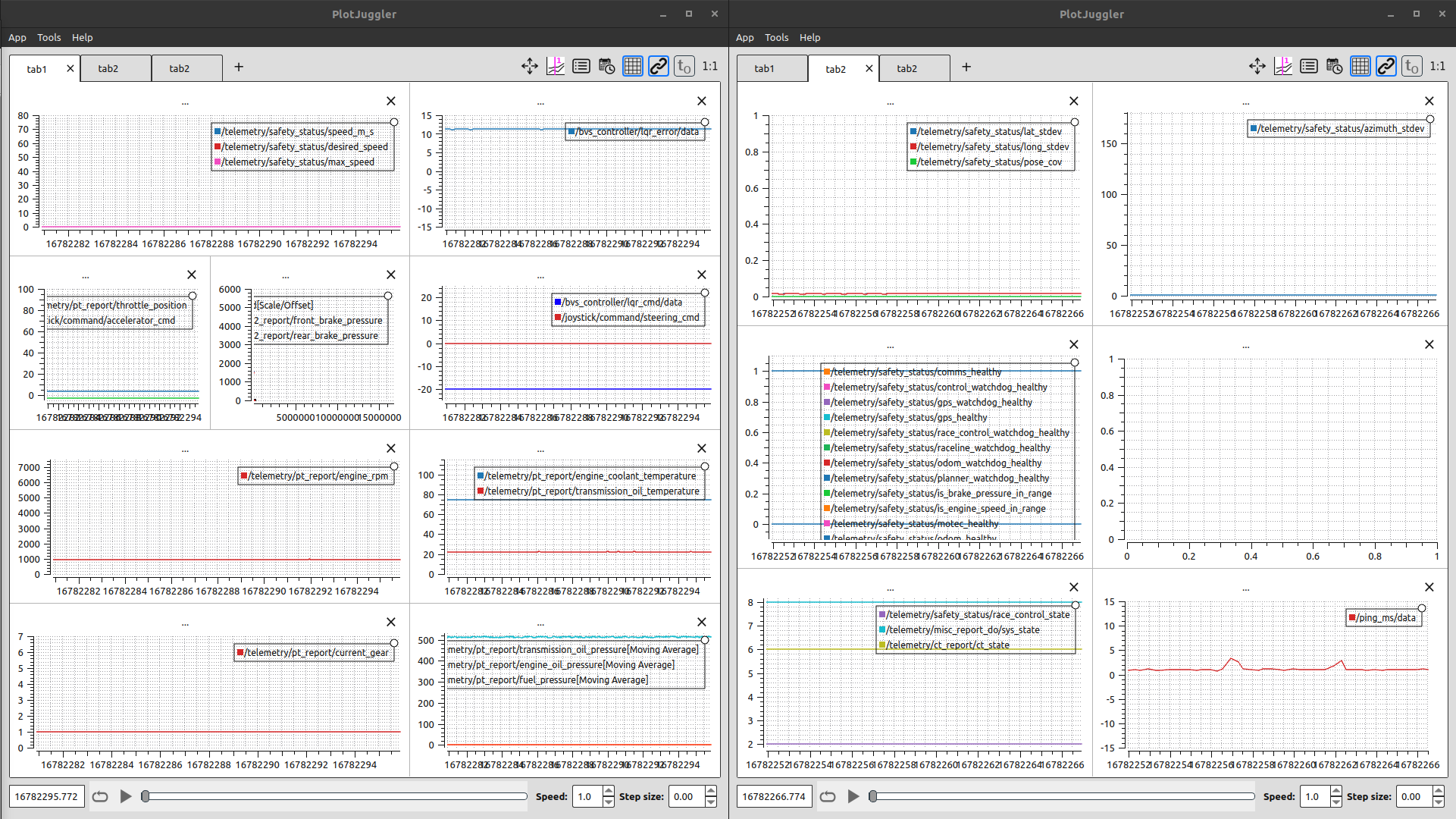}
\caption{Secondary interface for the Base Station}
\label{fig:secondary_ui}
\end{figure}

The secondary interface displays all incoming telemetry data and more readings, such as fluid pressures and controller errors and commands. Figure \ref{fig:secondary_ui} shows the display for the secondary interface. Like the primary interface, the secondary is modular and easily configurable to fit the area of concentration. It is constructed using PlotJuggler to plot all of the incoming telemetries. The interface serves to provide a detailed live feed of the health and performance of the vehicle and the stack, rather than serving as a quick-glance check. The benefit of the secondary interface is that operators can monitor the trends of the vehicle, and determine possible impending issues. 

\section{Evaluation and Results}\label{section:evaluation}

The aforementioned stack has been developed for and fielded on four different oval raceways across two competition seasons and three years. Over that time, the AV-21 ARV has seen hundreds of autonomous miles, in both single and multi-agent scenarios. In this section, we present evaluations of several real-world runs that demonstrate the stack\'s effectiveness and shortcomings. 

Additionally, between Seasons One and Two, while the overall approach and design of the stack had not changed materially, the execution and implementation details have been improved based on lessons learned from the first season. These insights are valuable for designing and fielding complex autonomous systems. 

First, we present results and analysis from runs in Season One. These runs are both single and multi-agent. We also evaluate a critical multi-agent run, where the vehicle was unable to demonstrate all requirements to qualify for the final competition event for the IAC@CES Event on January $7^{th}$, 2022. We lay out what failed and the key takeaways. Next, we present a similar analysis for runs from Season Two, highlighting the key changes in the actualization of the stack that enabled single-agent driving and head-to-head passing in speeds above $150mph$.

\subsection{Season One: Single-Agent High-Speed Run \& Spinout}

In this evaluation, we will examine a single-agent run where the vehicle ultimately exceeded $140mph$ in top speed before spinning out of control. The hope is to present a thorough analysis of the LQR Pure-Pursuit controller at higher speeds and demonstrate its strengths and weaknesses. Finally, we will discuss some insights into controller design and deployment that are driving our current development.

\subsubsection{Run Objectives}
The intended goal of the run was to perform a thorough verification of the controller's performance to maintain a fixed speed and its ability to quickly and safely slow down in the event of a safety trigger, such as a communication or race control timeout. Because the track is severely banked (up to over $20^{\circ}$ in the steepest portions), too sudden of decelerations can result in a spin-out, as the rear tires lose grip with the ground during a deceleration event, due to a forward shift in the mass distribution of the vehicle. When the tires lose this grip, the vehicle is more prone to spinning out and losing control. So, the controller must maintain safe, smooth decelerations at all times.

After verifying the speed controller's performance, the run was supposed to progress onto multi-agent testing; however, the second team was not ready and our sing-agent run continued. At this point, we set a goal to reach higher speeds (greater than $145mph$) and see the controller's performance above our previous record of $138mph$. While we did not achieve our original goal, the insights gleaned from what ensued are likely valuable for future research.

\subsubsection{Pre-Instability Performance}

\begin{table}[h!]
    \centering
         \begin{tabular}{|>{\centering}p{2cm}|>{\centering}p{2cm}|>{\centering}p{2cm}|>{\centering}p{2cm}|>{\centering}p{2cm}|>{\centering}p{2cm}|}
            \hline
                \textbf{$>45m/s$} & \textbf{$>10m/s$} & \textbf{$>60m/s$} & \textbf{$55-60m/s$} & \textbf{$50-55m/s$} & \textbf{$45-50m/s$} \tabularnewline
            \hline
                \textbf{0.501} & 0.244 & 0.540 & 0.457 & 0.573 & 0.456 \tabularnewline
            \hline
        \end{tabular}
        \label{spinout_table:1}
\end{table}

\begin{table}[h!]
    \centering
         \begin{tabular}{|>{\centering}p{2cm}|>{\centering}p{2cm}|>{\centering}p{2cm}|>{\centering}p{2cm}|>{\centering}p{2cm}|>{\centering}p{2cm}|}
            \hline
                \textbf{$40-45m/s$} & \textbf{$35-40m/s$} & \textbf{$30-35m/s$} & \textbf{$25-30m/s$} & \textbf{$20-25m/s$} & \textbf{$10-20m/s$} \tabularnewline
            \hline
                0.403 & 0.275 & 0.279 & 0.248 & 0.057 & 0.074 \tabularnewline
            \hline
        \end{tabular}
        \caption{Cross-Track Error [m] over several speed brackets. The value in bold represents the average cross-track error when the vehicle was traveling at more than $45m/s$, or over $100mph$. Error over the whole run, when the vehicle is moving above $10m/s$, is also provided.}
        \label{spinout_table:2}
\end{table}


\begin{figure}[h!]
\centering
\includegraphics[width=\linewidth]{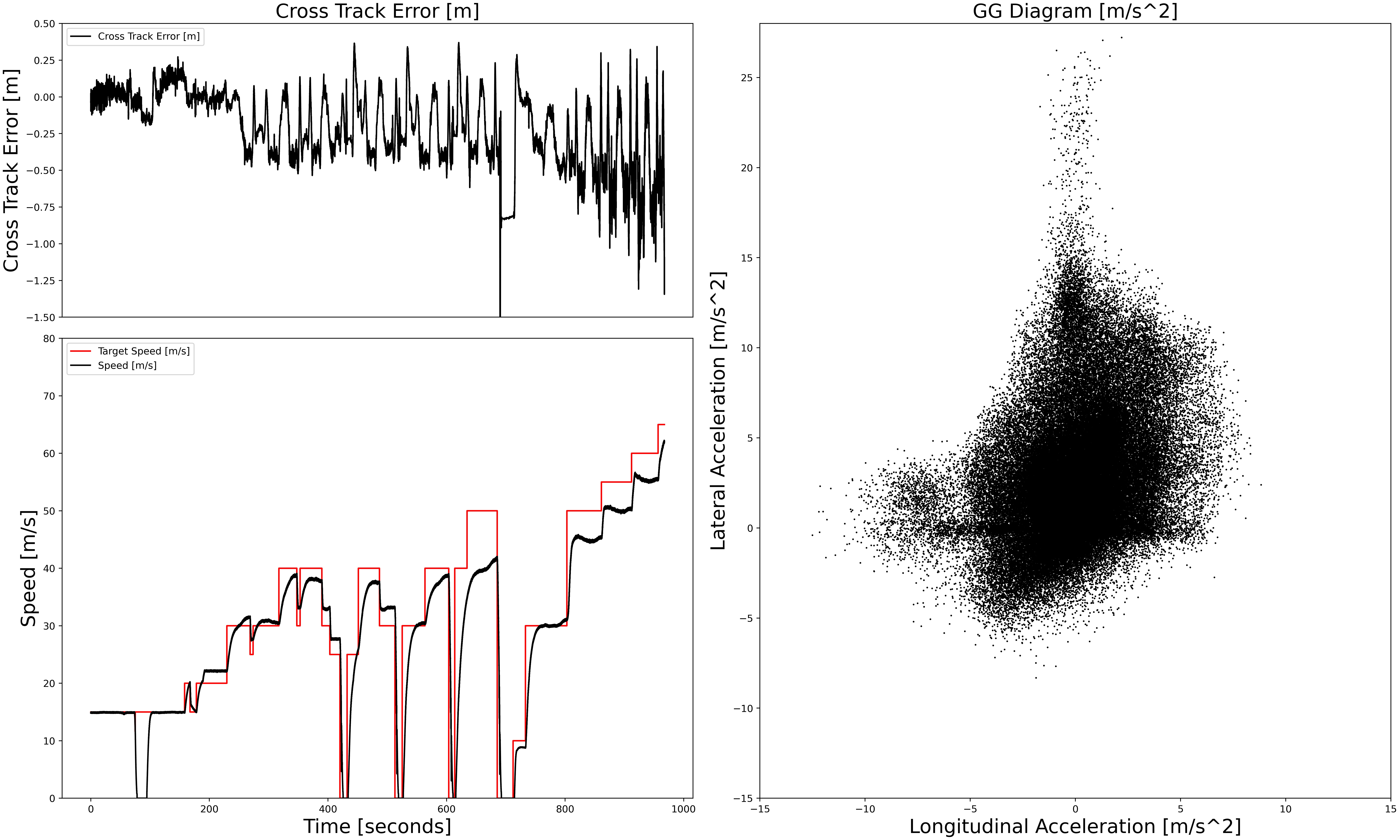}
\caption{ \textbf{(Top Left)} Cross-track error. \textbf{(Bottom Left)} Vehicle speed. \textbf{(Right)} G-G diagram showing the vehicle acceleration, in $m/s^2$. During this run, braking tests were conducted to identify the ability of the controller to respond to instantaneous  requests to stop the vehicle. The controller tracked the reference trajectory within $1.25m$ cross-track error, up until the spin-out. Finally, the G-G diagram shows significant levels of longitudinal acceleration while cornering.}
\label{fig:141_cte}
\end{figure}

During the repeated braking events, the controller maintained stability. The speed controller still needed more tuning, and the purpose of this run was to collect data to further tune the controller. Notably, there was a point at about 700 seconds into the run where the braking resulted in some traction issues, which corresponds to the jump in cross-track error as the vehicle slows to a stop. Information like this helps provide an estimate of where the braking limits of the tires are. Overall, as noted in Table \ref{spinout_table:2}, the average cross-track error (CTE) over the entire run was $0.244m$. When traveling over $100mph$, the CTE was $0.501m$. At its worst, the CTE does not exceed roughly $1.25m$ when traveling around the corners (see Figure \ref{fig:141_cte}).

The G-G diagram plots the longitudinal and lateral accelerations. As tires have a maximum deformation, thereby maximum total force, it is important to evaluate what region the vehicle and tires are operating in. For example, if the vehicle were braking or accelerating into a corner, a high lateral and longitudinal acceleration is expected. However, if the vehicle is maintaining the same speed while traveling around the track, the expectation is that the vehicle would not see as high longitudinal accelerations, but still have significant lateral accelerations as it makes the turns.

In this run, due to the repeated speeding up and braking, the vehicle experienced a wide spread of accelerations. As shown later, the vehicle was accelerating significantly even while going into the corners, which is a precarious scenario because there is a risk of saturating the tires, which is ultimately what happened. Preventing future events like this will require explicitly modeling the tire limits and constraining the vehicle's controls to stay within the dynamically feasible and safe region. However, determining these limits is still an open area of research.

\subsubsection{Spin-Out}
The night before January 3rd at LVMS saw the weather settle down to freezing temperatures. As it progressively turned colder, dipping to around $30^{\circ}$F by morning, conditions on the track became too cold to operate safely on. These conditions forced teams to shift testing operations to begin in the afternoon when temperatures reached about $50^{\circ}$F. The tires and power train of the AV-21 are designed to operate in warmer conditions. Freezing temperatures, such as those seen on January 3rd, can negatively impact performance.

The most obvious consequence of lower-than-optimal tire temperatures on the AV-21 is reduced tire traction and grip. In traditional motorsports, even when conditions are very warm, human drivers must ensure their tires are brought up to an optimum temperature. A race car's tires must be operating at a temperature of at least $175^{\circ}$F to generate traction and grip \cite{smith_tune_1978}. At the time of the spin-out, the tires were only at about $29^{\circ}$C, or about $85^{\circ}$F.

\begin{figure}[h!]
\centering
\includegraphics[width=\linewidth]{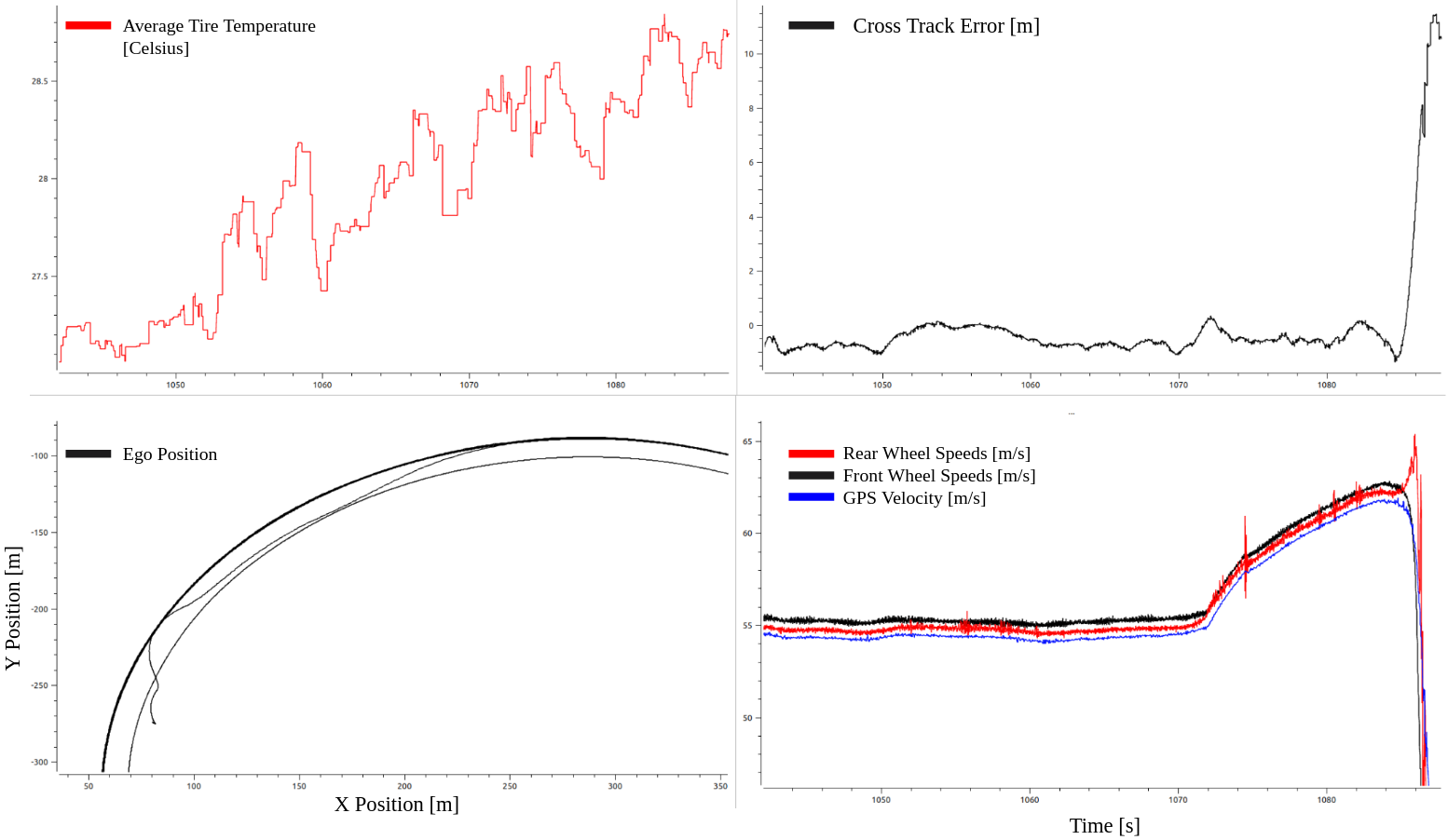}
\caption{ \textbf{(Top Left)} Average temperature of all four tires. \textbf{(Bottom Left)} Vehicle position. Note the deviation from the racing line when the vehicle spun out of control. \textbf{(Top Right)} Controller Cross-Track Error. \textbf{(Bottom Right)} Rear and Front wheel speeds and speed as measured by GPS. Notably, the rear wheels start to move much faster than the front wheels and deviate from what GPS is measuring. This indicates an over-steering event, where the rear tires lost traction and began to spin out.}
\label{fig:spinout}
\end{figure}

In Figure \ref{fig:spinout}, the vehicle is stable up until the point of failure. Notably, the tires were very cold (about $84^{\circ}$F); meanwhile, the vehicle was commanded to speed up very aggressively (command increase of about $20mph$) while going into the corner. These commands were issued by the base station operator. Additionally, given how the rear wheel speeds spike while the front wheels do not, traction was lost on the rear tires, indicating an over-steering event. Finally, due to the decoupled longitudinal and lateral controls, the steering controller was not able to dictate a more reasonable velocity profile given the trajectory ahead. Instead, the base station operator was setting a speed, and the speed controller was ramping up to meet it, regardless of where the vehicle was on track and what trajectory it was taking.

\subsubsection{Lessons Learned}

Several insights and changes were made resulting from this incident, some software, others operational. First, the base station operator needs to verify that the tire temperatures are warm enough before trying to command the vehicle to a high speed. At the time of the incident, the base station did not display tire temperature. Moving forward, this feature has been added. Additionally, when doing constant and high-speed tests, unless there is confidence in the vehicle's ability to navigate while maintaining that fixed speed, speed changes are avoided on turns.

Secondly, the development of improved tire modeling, with an online estimation component, is underway. It is not enough to set tire model parameters pulled from a data sheet and expect the vehicle to perform as expected all the time; rather, some online estimation of the tire parameters is needed. This is not a novel idea; rather, it has been explored before \cite{tum-friction}.

Finally, to take advantage of a better tire model, it is important to have a controller that can better reason about the vehicle dynamics and can jointly optimize the vehicle speed, steering, and accelerations. Development is underway on a model-predictive controller (MPC) that can predict and optimize the vehicle's performance over a finite time horizon. With a better vehicle model, controller, and better operating procedures, spinouts should be less common.

In terms of total cost and damage, the spinout was relatively minor, with a single bent rear suspension piece. However, the data collected and insights are invaluable for future development and testing. One of the operating limits of the tires has been found, a limit that would not have been found otherwise. Because of the insights learned and data collected, this run was considered a success by the team, despite the result.

\subsection{Season One: Failed Qualification Run Evaluation}
In this evaluation, we discuss the compounding issues that prevented the vehicle from detecting and tracking an opponent vehicle, which ultimately led to the team being disqualified from participating in the multi-agent passing competition. Ultimately, the issues were narrowed down to perception and tracking failures, which stemmed from a multitude of causes.

\subsubsection{Perception \& Tracking Failures}
In the run-up to competition day, several of last-minute decisions and changes were made to the software, vehicle and sensor configuration, and computer hardware. All of these changes compounded and resulted in a total failure of our perception and tracking stack on the last day to qualify for the head-to-head competition event. While disappointing, the lessons learned reached far beyond software design and development.

\begin{figure}[h!]
\centering
\includegraphics[width=\linewidth]{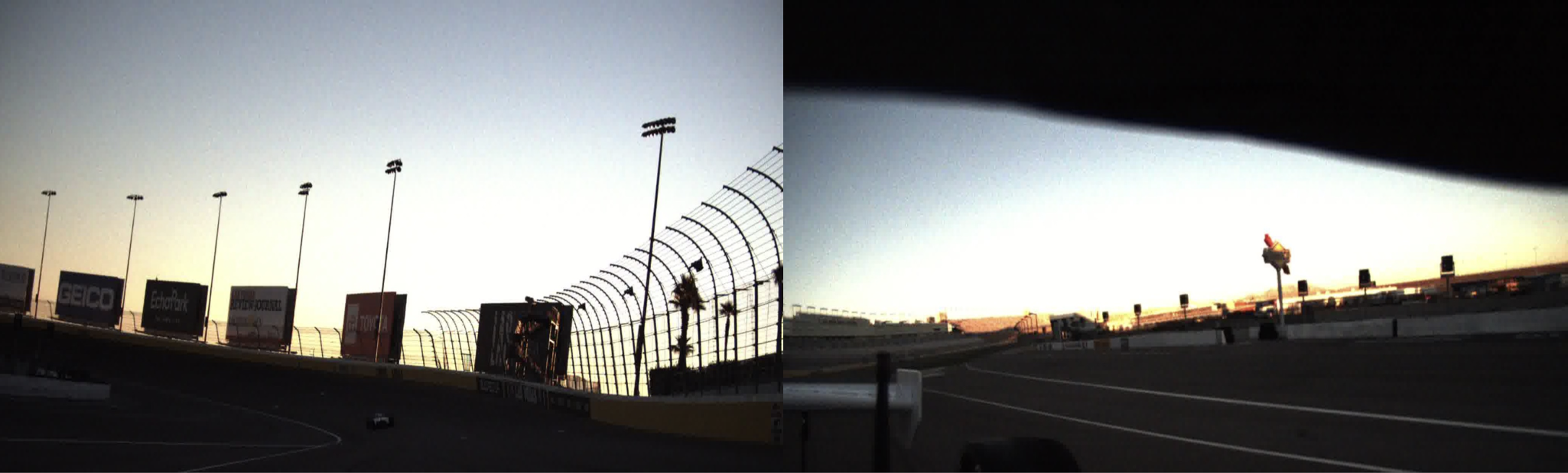}
\caption{Examples of camera images from the qualification run. Additionally, due to incorrect network bandwidth settings on the cameras, the frame rates were severely limited, operating under effectively $1Hz$ for some cameras.}
\label{fig:cam_bad}
\end{figure}

\subsubsection{Camera Hardware \& Driver Changes}
Due to issues with the network connections on the vehicle, camera drivers had to be reconfigured to reduce their network bandwidth, resulting in the streaming of wrongly-cropped or poor quality images. These images, examples shown in Figure \ref{fig:cam_bad}, on top of a lack of training data from the LVMS track, resulted in a higher-than-expected false positive rate.

\subsubsection{Failed Motion Compensation Changes}


\begin{figure}
    \centering
        \subgraphics{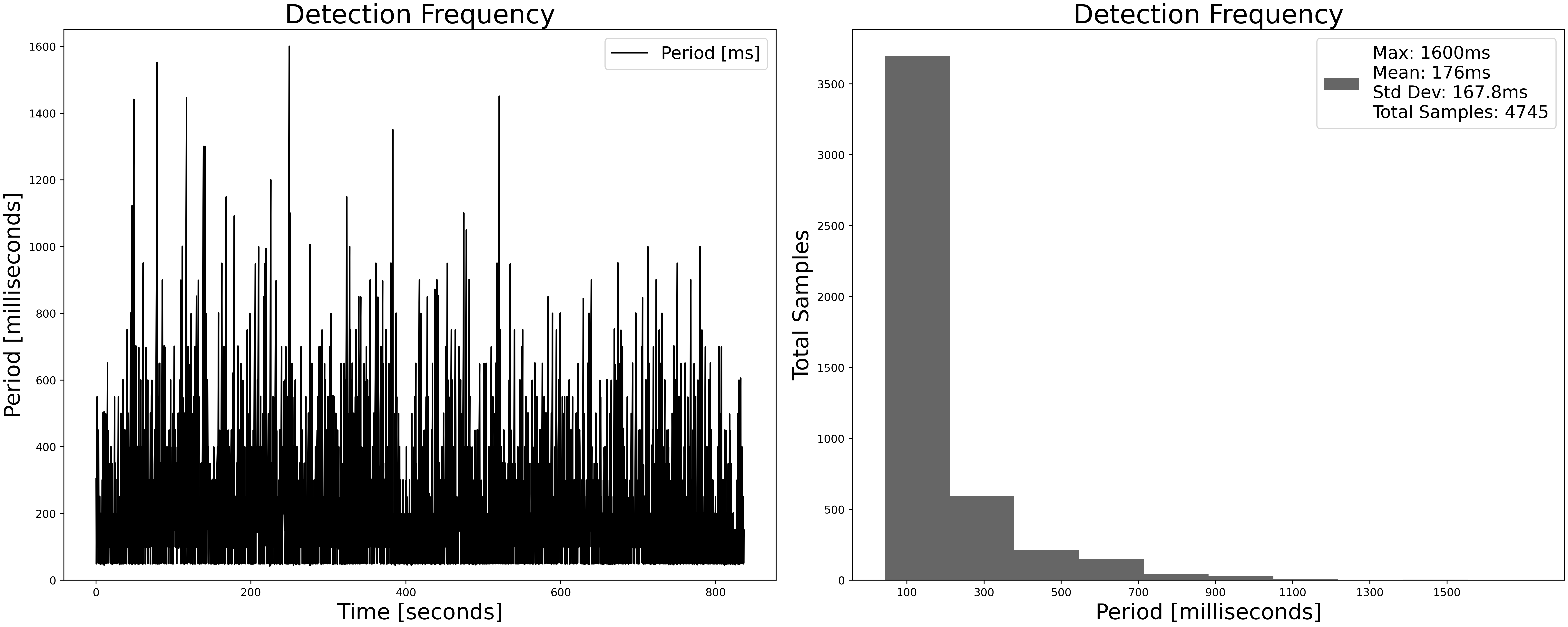}{6cm}
        \subgraphics{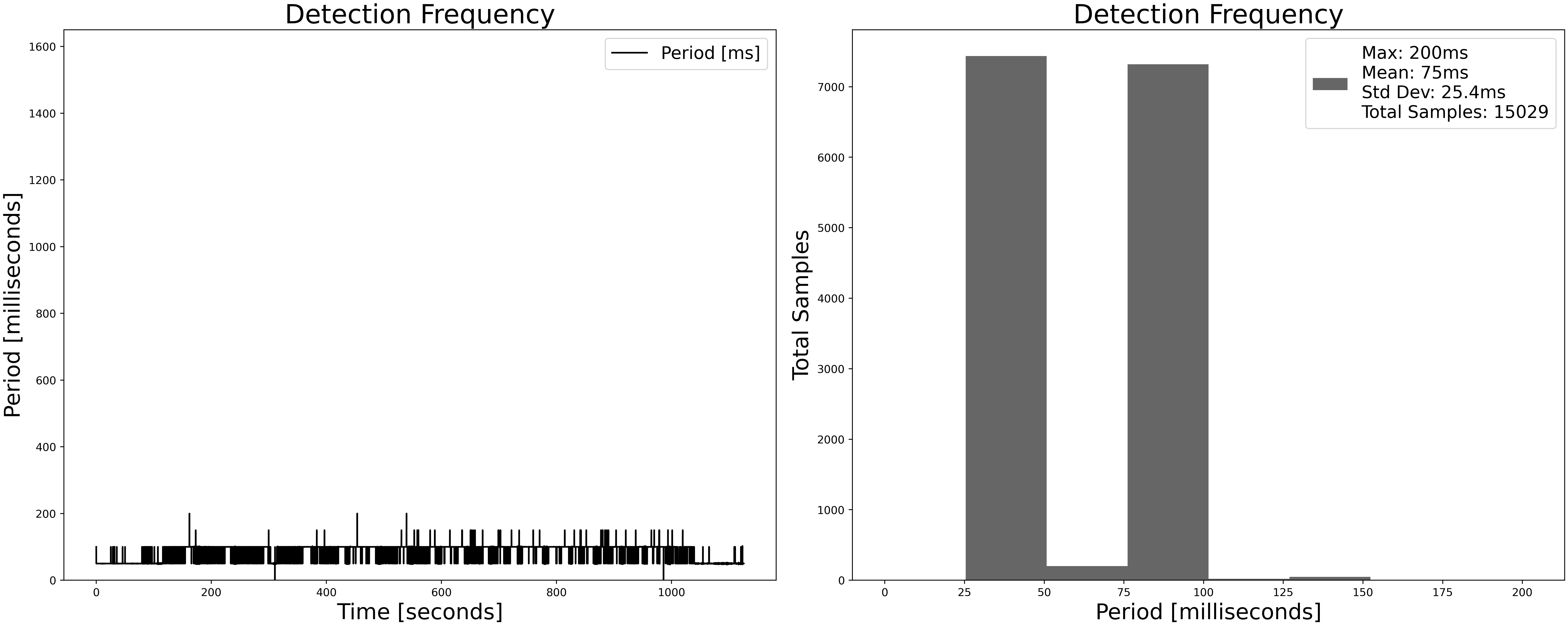}{6cm}
    \caption{\textbf{(Top Row)} Frequency of LiDAR detections during the failed run. \textbf{(Bottom Row)} Frequency of LiDAR detections during the semi-finals match in Season Two. The time period between detections was calculated by taking the time delta between sequential LiDAR detections. Since the LiDAR operates at $20Hz$, the expected frame-to-frame period is $50ms$, assuming that detections process every frame. In the results from Season One, the time period between detections was far longer (an average of $200ms$, or an effective frequency of $5Hz$), due to dropped data packets arising from middleware library issues. As a result, the tracker had difficulty tracking the agent over time. Note that the above detections may still have low processing latency ($50ms$ between the time the sensor data was taken and the detection was computed, as shown in Fig.~\ref{fig:lidar_latency_good}), illustrating the difference between latency and frequency. Finally, the results from Season Two show significant improvement, with LiDAR detections processed half the time running at sensor frame rate and the other half with at most one detection missed. Occasionally, additional missed frames were witnessed, but much less frequently than in Season One.}
    \label{fig:latency_bad}
\end{figure}   

Since the LiDAR is not scanning every point instantaneously it is important to account for the ego vehicle's motion during the LiDAR scan for the highest-accuracy result. However, leading up to the competition, the solution developed was not thoroughly validated on the vehicle while it was under full system load. On the AV-21, there are three LiDARs, each scanning at $20Hz$. The motion distortion node needed to combine all three clouds, project each point into the world to correct the distortion, and, optionally, project the cloud back into a local frame of reference. ROS 2 provides a message synchronization library that gives a convenient interface to subscribe to multiple topics and have all three topics delivered at the same time, in the same callback.

When under full system load, the DDS middleware and synchronization library was not delivering every LiDAR measurement from all three LiDARs. Messages were frequently being dropped. However, because the issue only showed up when the system was under load, it was not discovered until deployed and run on the vehicle during the qualification run.

Due to the dropped messages, the effective LiDAR frame rate received by PointPillars dropped to below $5Hz$. This corresponds to an average period of roughly $200ms$. Assuming sensor processing is keeping up with the sensor frame rates, the expected period is $50ms$, as the LiDAR is scanning at $20Hz$. In actuality, the period between LiDAR measurements was very inconsistent, often spiking over half a second (see Figure \ref{fig:latency_bad}). This inconsistent and slow update rate compounded with other issues with the tracker that resulted in very poor perception performance overall. 

\subsubsection{Poorly Tuned \& Flawed Outlier-Rejection}
With the camera pipeline producing false positives, albeit at a low rate due to network bandwidth issues; and LiDAR running at a much lower frequency than what was expected, the fusion relied primarily on the Radar for tracks. However, last-minute code changes and tuning resulted in poor outlier rejection, which is especially important due to the high noise in radar tracks. While some good tracks surfaced and were followed, tracking failed to prune too many false positives. Additionally, the radar can only see in front of the vehicle, which is insufficient on its own for passing.

\subsubsection{Lessons Learned}

All three issues quickly compounded and it was clear to the base station operator that there were too many false positives from the perception and tracking stack. As a result, a strategic decision was made to terminate the run and have the vehicle return safely home. While a disappointing result, the run provided invaluable data and insights into testing and deploying a full, complicated system onto real hardware. Most importantly, this failure highlighted the importance of proper integration testing to identify unexpected issues sooner. Additionally, the next section will address the solutions that were developed and tested over Season Two and the resulting performance on track in competition.

\subsection{Season Two: IAC@CES 2023 Competition Performance}

\begin{table}[h!]
    \centering
    \begin{tabular}{|c|c|c|c|c|}
        \hline
            Run & Format       & Peak Speed [mph] & Competing Team & Result\\
        \hline
            1   & Single-Agent & 145 mph          & -----          & Time Trial, Run One \\
            2   & Single-Agent & 150 mph          & -----          & Time Trial, Run Two \\
            3   & Multi-Agent  & 115 mph          & KAIST          & Quarter-Final, MPRW Win \\
            4   & Multi-Agent  & 146 mph          & KAIST          & Quarter-Final Re-Run, MPRW Win \\
            5   & Multi-Agent  & -----            & Polimove       & Radio Died, MPRW Disqualified \\
            6   & Multi-Agent  & 153 mph          & AI Racing Tech & Fuel Depleted, MPRW Loss \\
        \hline
    \end{tabular}
    \caption{Competition Day Runs. Formats consisted of single-agent time trials and the Passing Competition multi-agent events. The peak speed reached by the vehicle during the run is presented. Despite a challenging five days of testing, that included three separate hardware failures and a disastrous crash (described in detail in Section \ref{cam_discussion_crash}) less than 72 hours before competition, the team finished $4^{th}$ overall.}
    \label{tab:final_competition_results}
\end{table}

Six out of nine total teams qualified to participate in the passing competition at CES in Las Vegas. In total, MIT-Pitt-RW (MPRW) had two single-agent runs, three successful multi-agent events, and one multi-agent run where the primary communication radio on the vehicle failed, rendering the car unable to compete. A full breakdown of every run, peak speed, and the result is presented in Table \ref{tab:final_competition_results}. The following evaluations will focus on Runs 2 and 6 from the table above.

\subsection{Season Two: Time Trial, Single-Agent Performance}

\begin{table}[h!]
    \centering
    \begin{tabular}{|c|c|c|}
        \hline
            Ranking & Team Name & Average Speed [mph] \\
        \hline
            1 & PoliMove & 168.2 \\
            2 & TUM & 164.9 \\
            3 & TII Euroracing & 144.4 \\
            4 & MIT-Pitt-RW & 143.8 \\
            5 & KAIST & 138.2 \\
            6 & AI Racing Tech & 65.9 \\
        \hline
    \end{tabular}
    \caption{Final Time Trial rankings and the average of the fastest laps from two runs. To determine the starting brackets and run order, a time trial was held on the morning of the event. Each team had two single-agent runs, each consisting of up to ten laps. The fastest lap time from each run was averaged to determine the team's overall score and ranking.}
    \label{tab:time_trial_rank}
\end{table}

To determine the starting brackets and run order, a time trial was held on the morning of the event. Each team received two, single-agent runs, each consisting of up to ten laps. Then, the speed from fastest lap time from each run was averaged to determine the team's overall score and ranking. MPRW finished fourth, coming within $0.6mph$ of the third-ranking team, TII Euroracing. The full rankings are presented in Table \ref{tab:time_trial_rank}.

\begin{figure}[h!]
\centering
    \includegraphics[width=\linewidth]{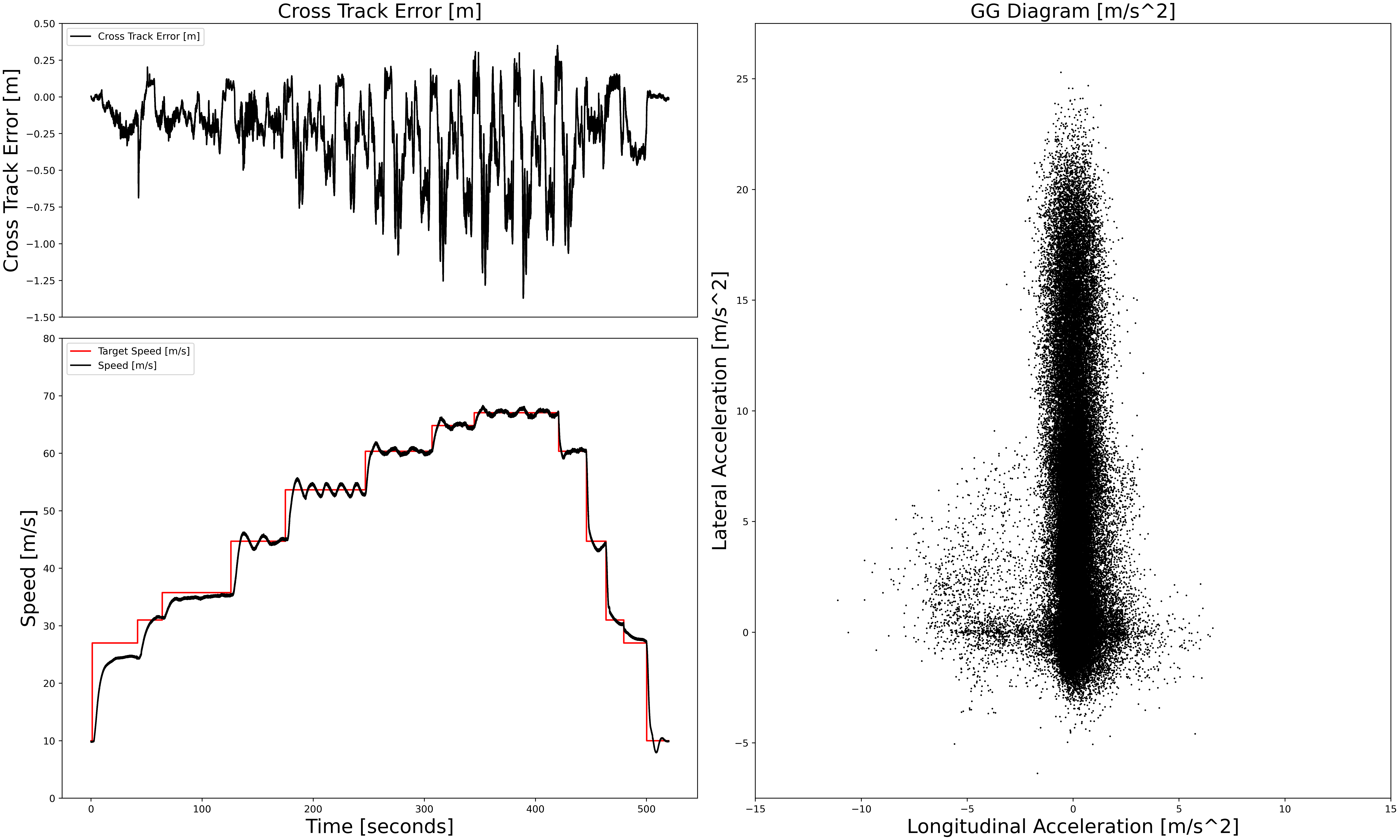}
    \caption{\textbf{(Top Left)} Cross-track error. \textbf{(Bottom Left)} Commanded and actual vehicle speed. \textbf{(Right)} G-G diagram showing the vehicle acceleration, in $m/s^2$. The vehicle reached a peak speed of over $67 m/s$, which equates to $150 mph$. The cross-track error peaked at approximately $1.4 meters$. Lateral acceleration reached over $24 m/s^2$, or almost $2.5 g's$ of acceleration.}
\label{fig:time_trial_controls}
\end{figure}

Overall, the controller was stable and able to navigate at high speeds. As shown in Figure \ref{fig:time_trial_controls}, the LQR controller balanced performance (maintaining $\leq 1.5 meter$ cross-track error (CTE)) while navigating turns with over $24 m/s^2$ of acceleration. The speed controller exhibited low-frequency oscillations in tracking the desired speed, centered around the desired speed. Part of this failure is believed to be a result of changed power train dynamics after the engine was repaired and the vehicle was reassembled. Future work includes better power train modeling and improving the tuning of the throttle controller. The data from this run in particular is instrumental for that future work.

\subsubsection{Discussion and Lessons Learned}
The first time trial run (at $145mph$) broke the team's speed record from the previous year ($141mph$ before spinning out). The second run quickly broke the team's record again, finally pushing through the $150mph$ barrier after more than three years of development. These milestones boosted team morale, in preparation for the passing competition. It also set the team up for the quarterfinal match-up against the KAIST team. 

This run also validated that simple, feedback-based controllers could navigate an ARV at high speeds on an oval race track. It is still unknown what issues will happen on more complicated circuits, i.e. ones with sharp left and right turns. In particular, LQR on its own does not reason about control limits, outside of a naive clamp. Additionally, because it is only ever trying to drive the current state error to zero, LQR can be myopic. On an oval track, this behavior does not cause issues, as the vehicle rarely needs full control bandwidth. However, on more complicated circuits, it is not uncommon to hit the steering limits. Future work is two-pronged: adding a feed forward component to the control produced by LQR and developing a model predictive controller that explicitly considers the maximum control constraints.

\subsection{Season Two: Passing Competition, Multi-Agent Performance}
After the radio failure in Run 5\ref{tab:final_competition_results}, MPRW was disqualified and was not able to compete in the finals. AI Racing Tech also experienced hardware issues earlier in the day, which prevented them from a second time trial run, thus impacting their overall score seen in Table \ref{tab:time_trial_rank}. As a consolation for the hardware failures, the competition organizers had the two teams compete in a match for $3^{rd}$ place.

\begin{figure}[h!]
\centering
    \includegraphics[width=\linewidth]{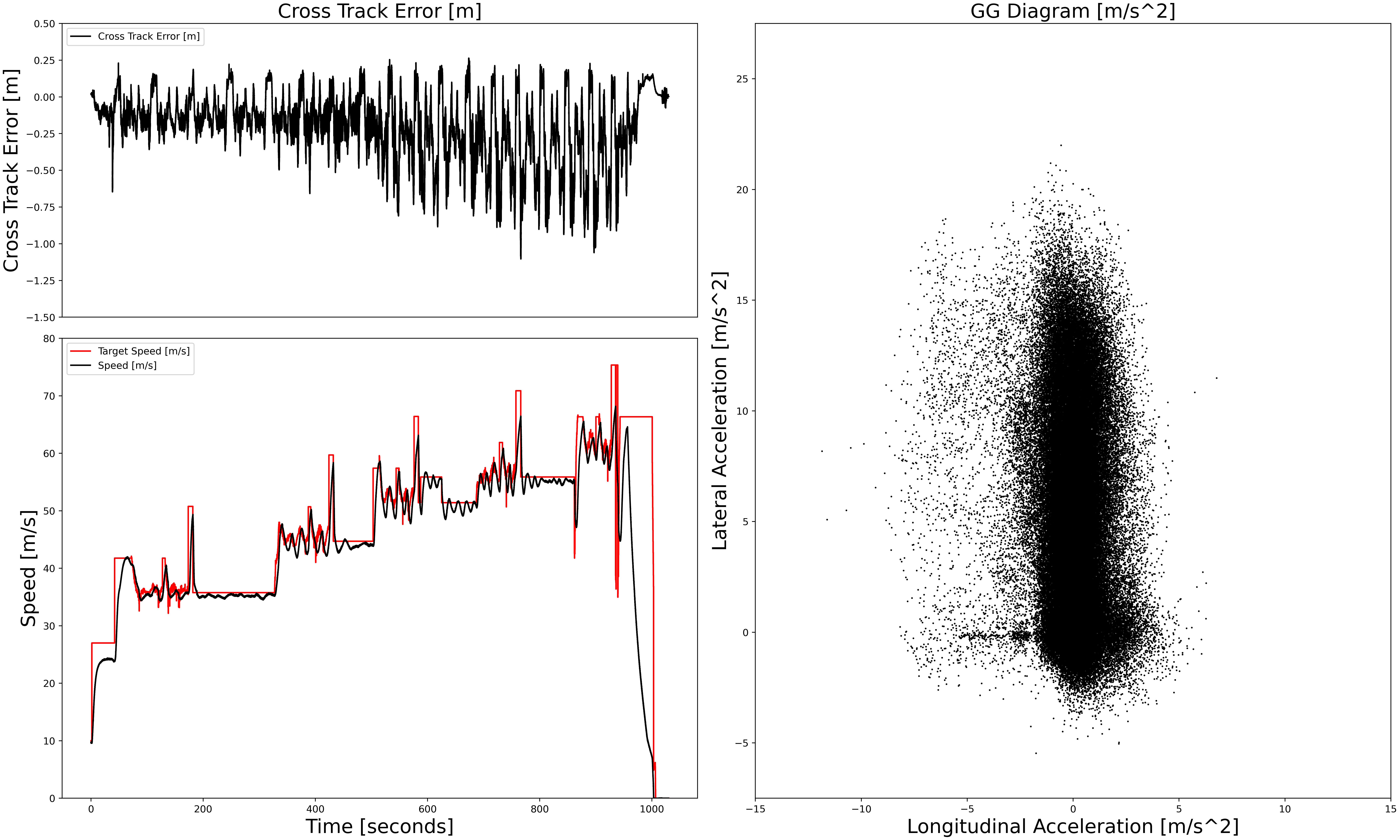}
    \caption{\textbf{(Top Left)} Cross-track error. \textbf{(Bottom Left)} Commanded and actual vehicle speed. \textbf{(Right)} G-G diagram showing the vehicle acceleration, in $m/s^2$. The vehicle reached a peak speed of over $68 m/s$, which equates to $153 mph$. The cross-track error peaked at approximately $1.4 meters$. Lateral acceleration reached over $20 m/s^2$, or over $2 g's$ of acceleration.}
\label{fig:ac_23_final_controls}
\end{figure}

In the match, both teams passed back and forth at the 80, 100, and 115 $mph$ speed brackets. The majority of the event was flawless, with only two minor issues, neither of which were the fault of either team's autonomy software. First, early in the run, communication was delayed with the vehicle, so a warning was sent to Race Control as the behavior was similar to what was experienced when the radio had died in the previous match. Race Control preemptively slowed down the AI Racing Tech vehicle, but that was not needed as communication restored itself without any further issues. The second issue was after the $115 mph$ bracket was successfully passed. Instead of maintaining the round speed at $115 mph$ for AI Racing Tech, Race Control set the speed to $125 mph$ prematurely. This was quickly communicated over the radio and Race Control remedied the issue by setting the correct speed and allowing AI Racing Tech an extra lap before considering the round started.

\begin{figure}[t!]
    \centering
        \subgraphics{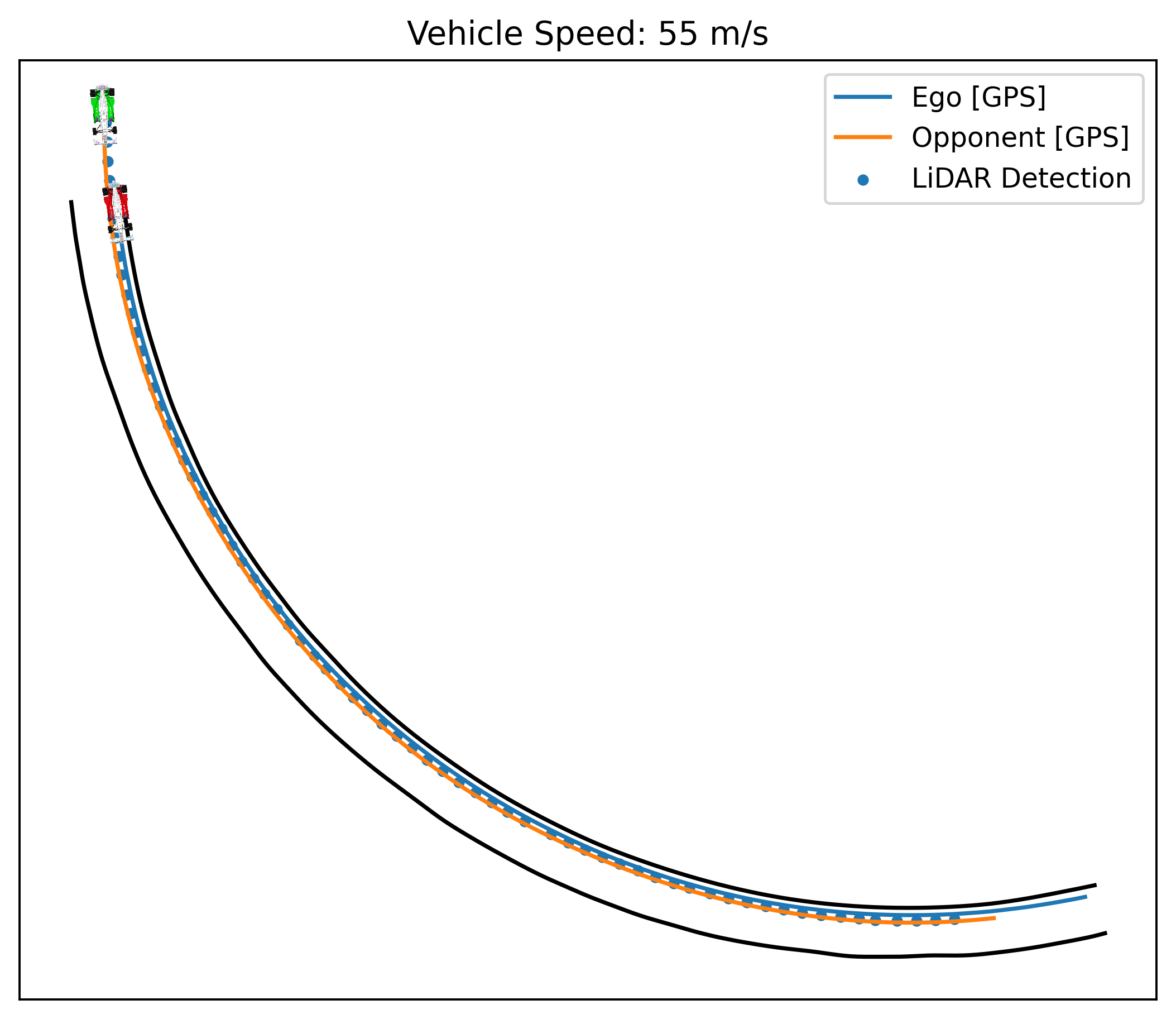}{4.5cm}
        \subgraphics{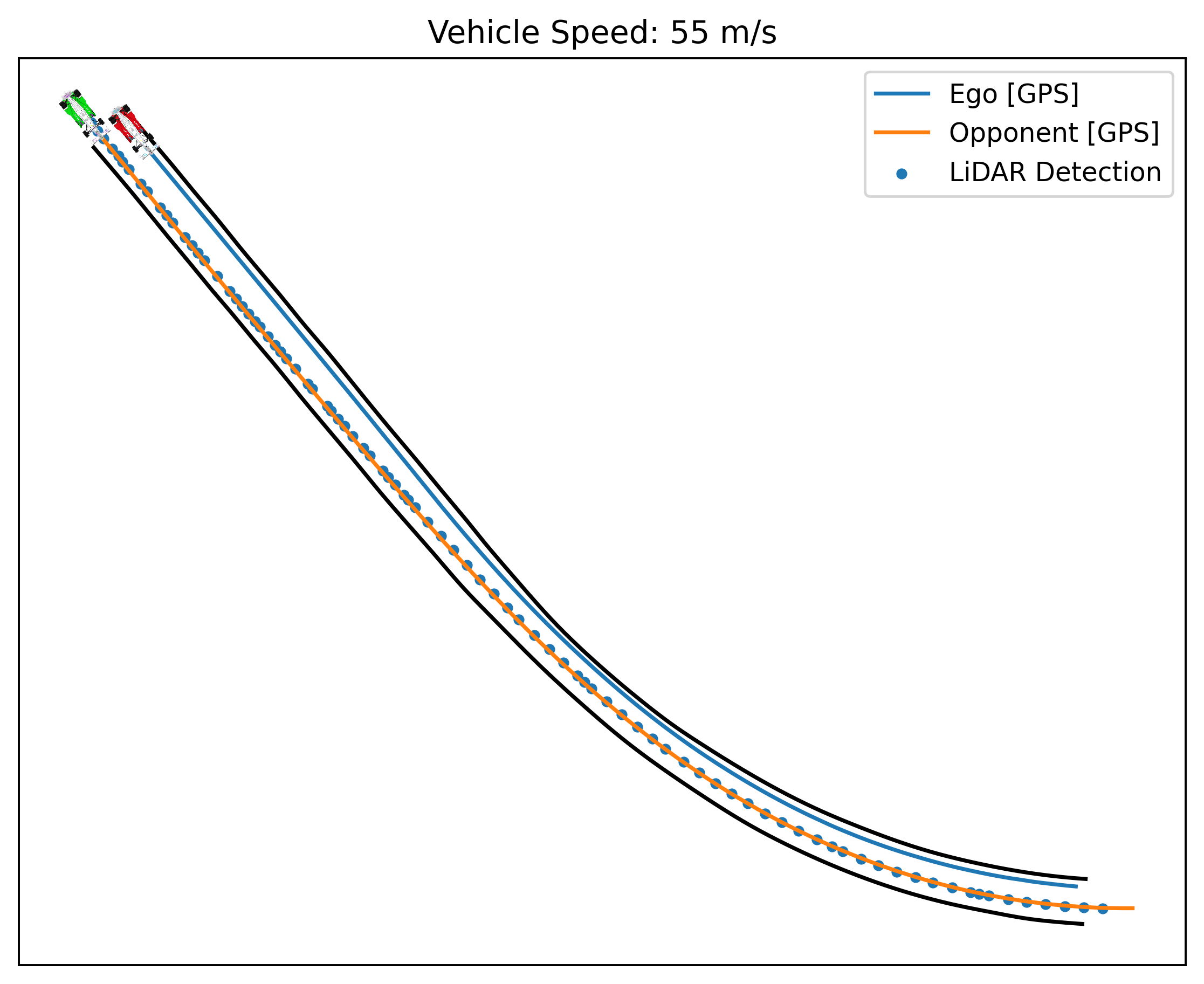}{4.5cm}
        \subgraphics{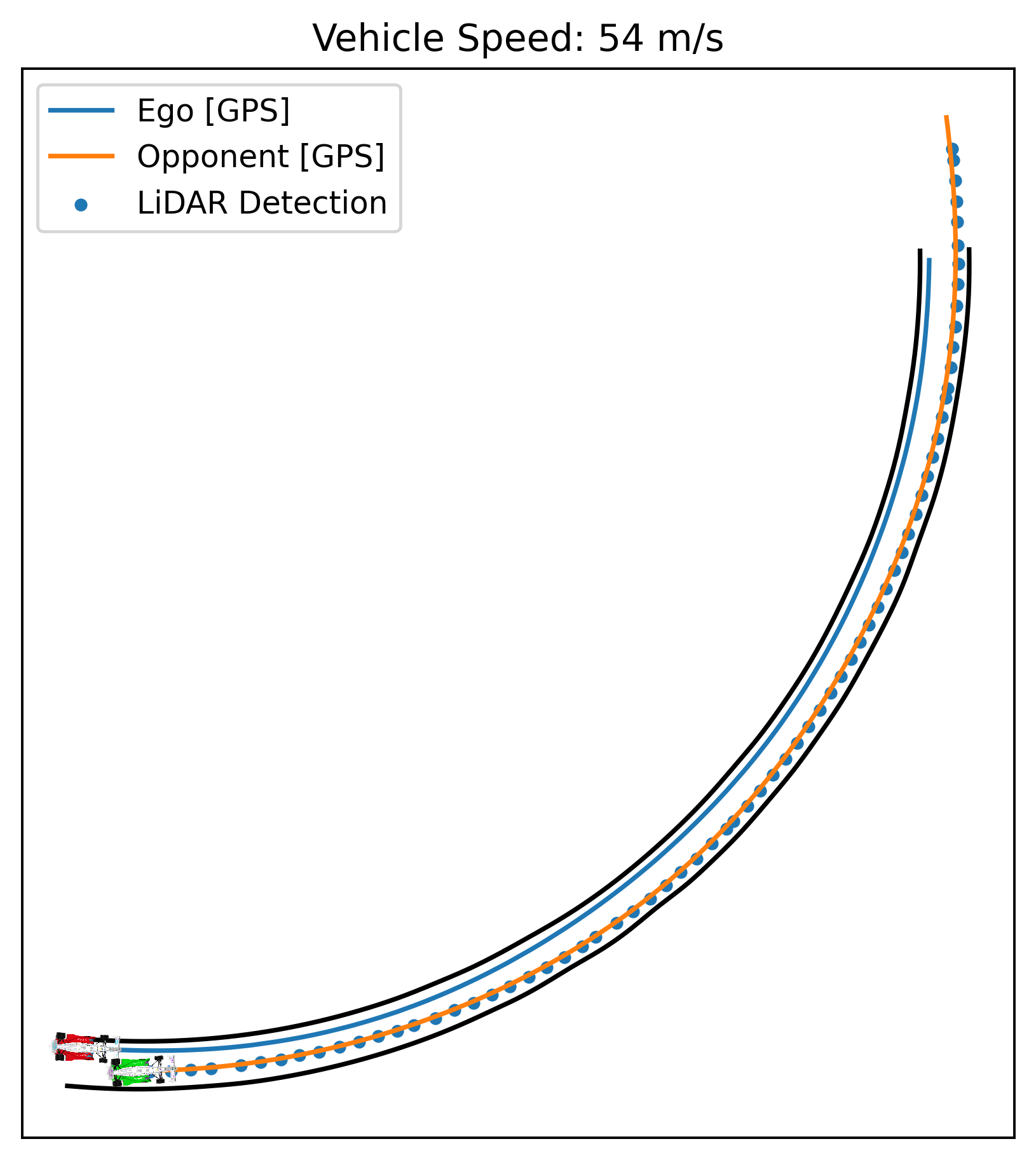}{4.5cm}
    \caption{Sequences from ``Defending" at the $125 mph$ speed. Each sequence is 7 seconds long. From left to right, the opponent AV-21 starts from behind the vehicle, moves to the outside lane, and accelerates and maintains a clear inside lane for our vehicle while completing the pass. Our AV-21 is able to detect and track the opponent during the entire sequence, even when they accelerate to pass while our vehicle is maintaining a speed of $125 mph$. Finally, the planner followed all right of way rules dictated by the competition.}
    \label{fig:finl_ac_23_125mph_defender}
\end{figure}


In Figure \ref{fig:ac_23_final_controls}, it is possible to see when the vehicle is ``trailing" the opponent versus Defending. Periods of a flat, constant commanded speed are when the vehicle is Defending. Periods of varying commanded speed, followed by large spikes, are when the vehicle is trailing and then passing the other vehicle. In total, four passes were completed, and a fifth was initiated but not completed. When attempting the final pass, a bug with the planner caused a drop in the requested speed. The planner would oscillate between identifying the agent as being in front or behind the ego vehicle, which resulted in an oscillating speed command. At the same time, the vehicle was also running out of fuel. Immediately after falling back behind AI Racing Tech, the planner commanded a higher speed to catch up; however, the vehicle ran out of fuel and could not maintain speed. It is unknown exactly when the engine started experiencing a drop in fuel, as it could have been during the initiation of the pass or after. Additionally, there is a possibility that if more fuel had been in the vehicle, the vehicle would have been slower (due to increased mass), thereby changing the entire dynamics up to this moment. Finally, this occurred in lap one of two, so the vehicle would have had a second chance at completing the pass, but did not due to running out of fuel. 

\begin{figure}[t!]
    \centering
        \subgraphics{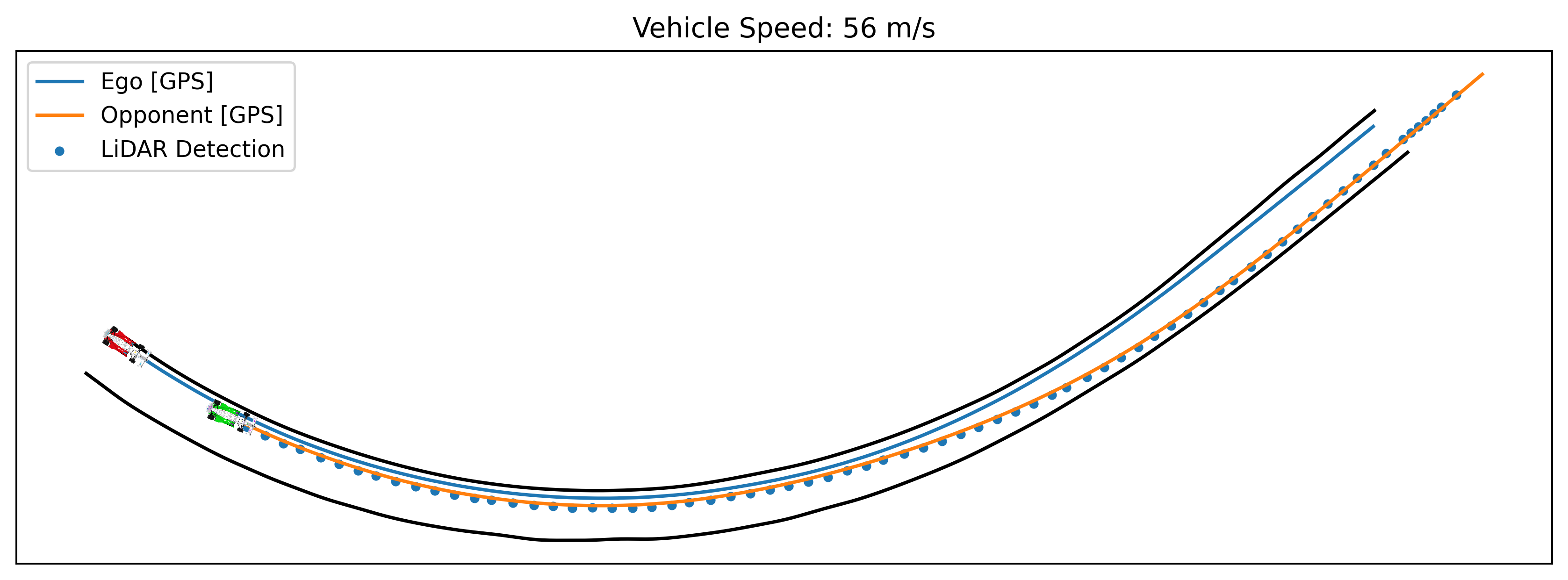}{4cm}
        \subgraphics{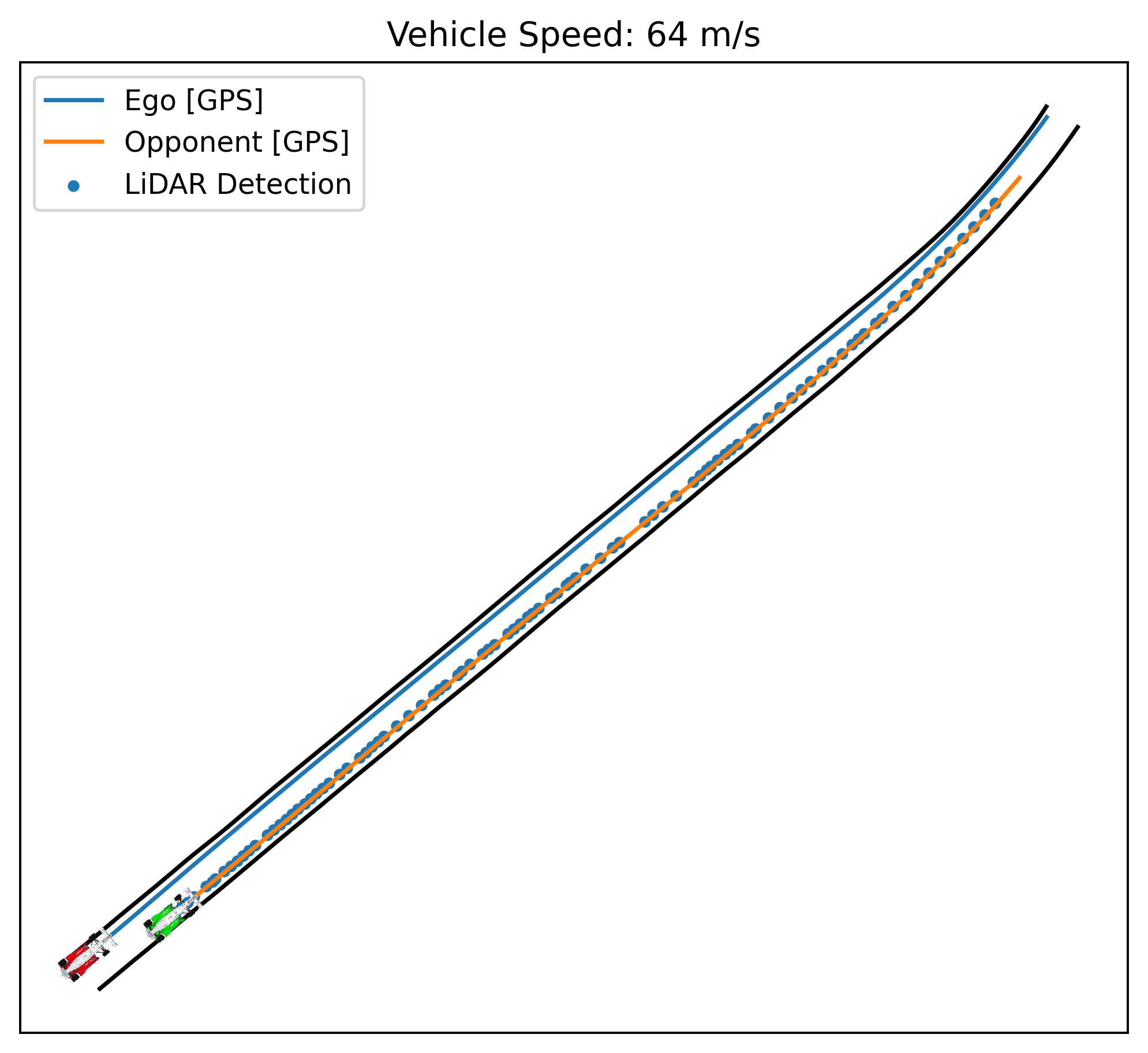}{4cm}
    \caption{Sequences from ``Attacking" at the $125 mph$ speed. Each sequence is 7 seconds long. During the pass, the vehicle reached a peak speed of almost $150 mph$. From left to right, the vehicle is preparing for the pass by choosing to maintain the inside lane. By the time the vehicles enter the turn, our vehicle has completed the pass and begins to drop back to the round speed. During the entire sequence, the stack is able to detect and track the other agent, providing the planner a reliable belief of where the other agent is, allowing for a safe pass, even at such high speeds.}
    \label{fig:finl_ac_23_125mph_attacker}
\end{figure}

\begin{figure}[t!]
    \centering
        \subgraphics{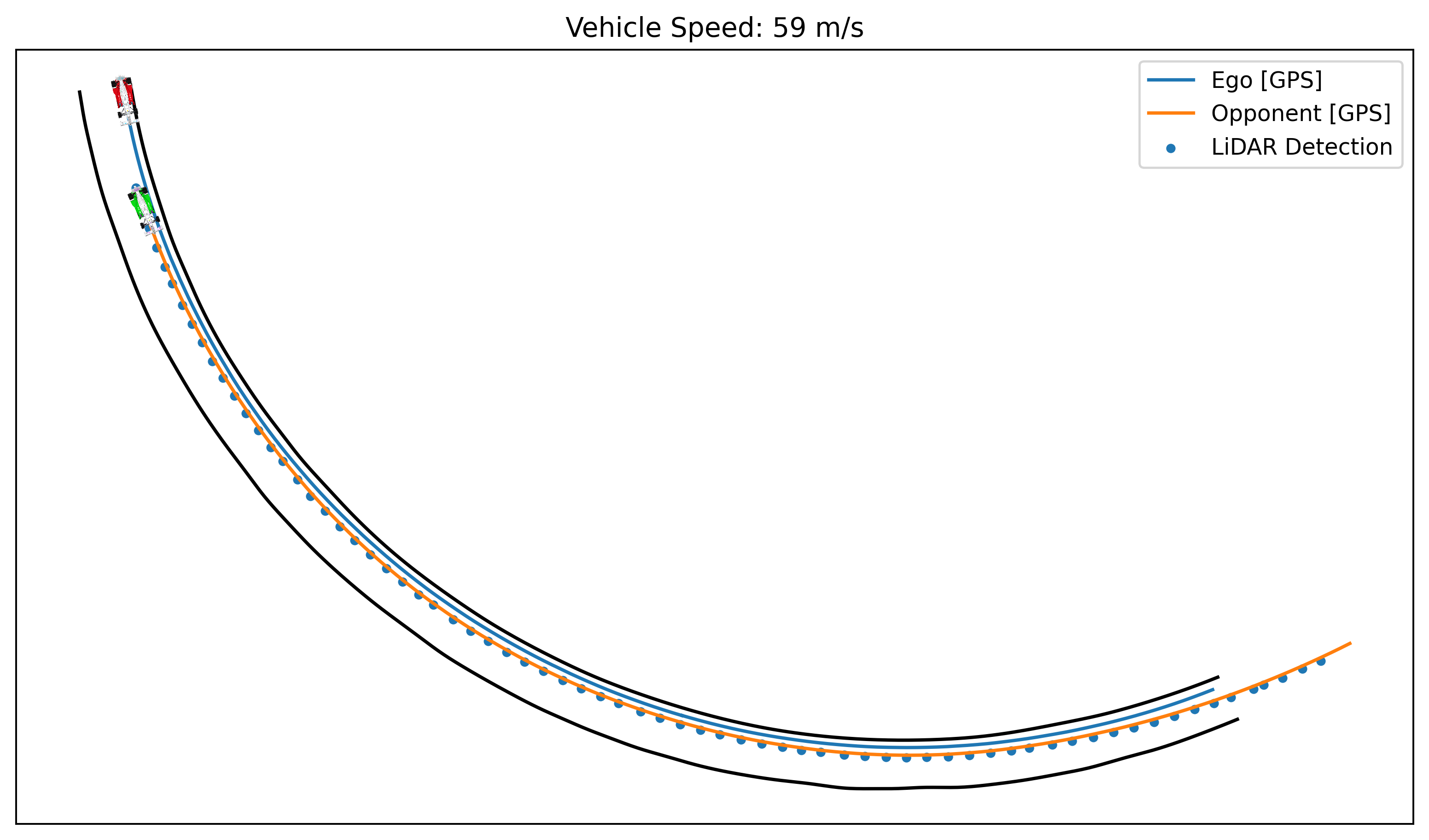}{5cm}
        \subgraphics{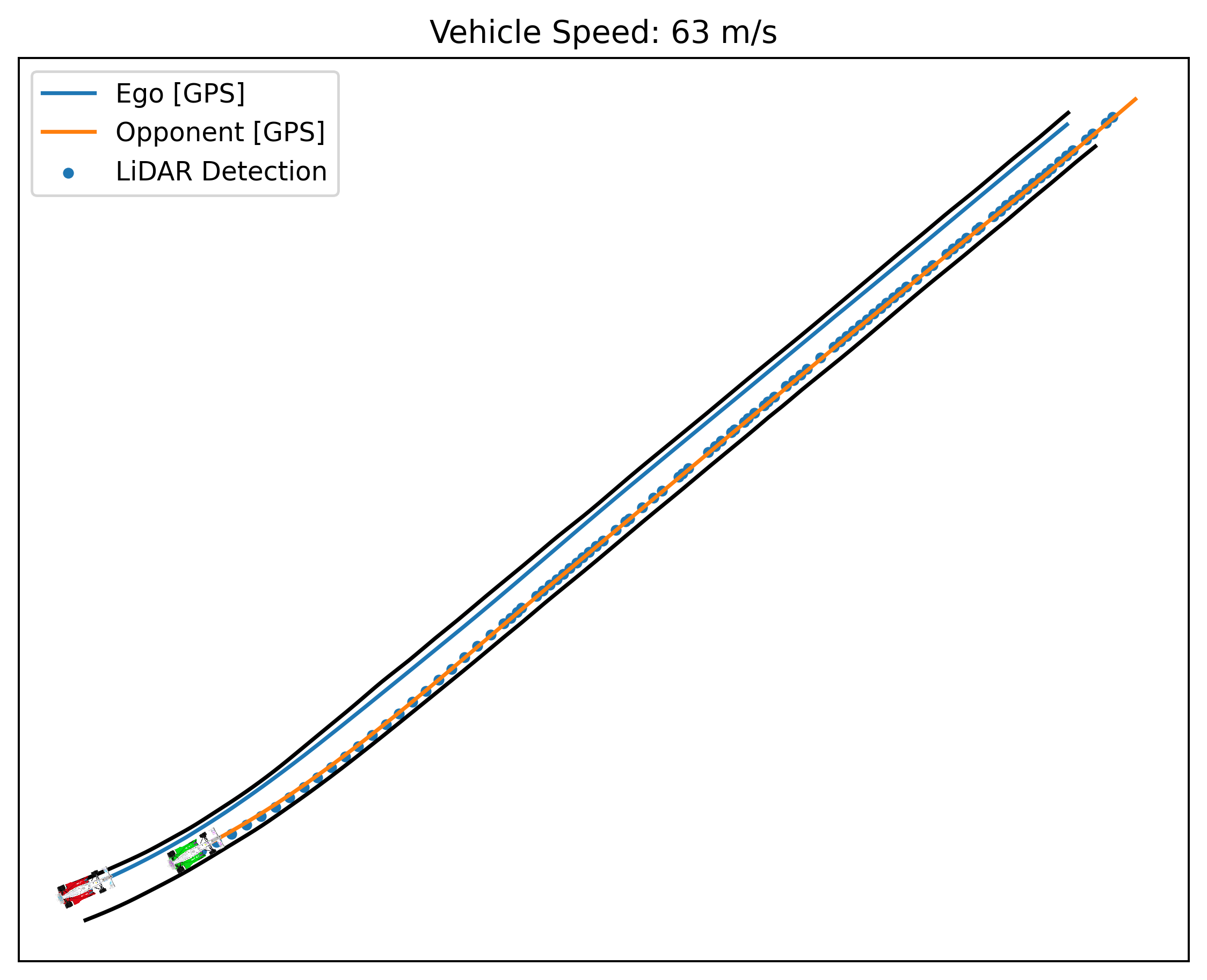}{5cm}
    \caption{Sequences from ``Attacking" at the $135 mph$ speed. Each sequence is 7 seconds long. Even at over $60 m/s$, the vehicle continued to track the other agent very well. Not long after these sequences, the final pass failed and the vehicle ran out of fuel, rendering the match over. In the second sequence, by the end of the straight, the vehicles are nearly side by side, but the vehicle is not able to complete the pass and win the round.}
    \label{fig:finl_ac_23_135mph}
\end{figure}

Figures \ref{fig:finl_ac_23_125mph_defender}, \ref{fig:finl_ac_23_125mph_attacker}, and \ref{fig:finl_ac_23_135mph} show 7 second long sequences from each of the speed rounds during the match. The RTK GPS measurements from both ego and opponent vehicles are plotted, with the LiDAR detections overlaid. The starts of the sequences are denoted by the vehicle graphics. The ego vehicle is red and the opponent is green. 

Figure \ref{fig:finl_ac_23_125mph_defender} shows the performance of the vehicle while maintaining the Defender role. As Defender, the vehicle must maintain a set speed and follow right of way, which depends on roles, relative distances between cars, and where the cars are on the track. While in this role, our planner meets these requirements. Additionally, the perception stack is able to reliably detect and track the other agent, even during a pass. Figure \ref{fig:finl_ac_23_125mph_attacker} shows the vehicle passing another AV-21, reaching a top speed of over $150mph$. During the sequence, the stack is again able to detect and track the opponent and complete the pass safely. Finally, Figure \ref{fig:finl_ac_23_135mph} shows another pass sequence, but, in this attempt, a combination of a planner bug and the vehicle running out of fuel, the pass is not completed. During all of the sequences, the full stack is working well, able to complete all requirements of the competition and pass another vehicle while traveling at very high speeds. 

\subsubsection{Discussion and Lessons Learned}

In this run, the vehicle achieved a new personal record for its highest peak speed, and it was also the only time the vehicle passed another car going over $125 mph$. In the testing leading up to race day, the fastest defender ever passed was maintaining $80 mph$. Additionally, before race day, only five passes were ever achieved, due in part to lost testing time from the crash a few days earlier. On race day, we achieved the following:
\begin{enumerate}
    \item Three passes at $80mph$ bracket
    \item Two passes at $100mph$ bracket
    \item Two passes at $115mph$ bracket
    \item One pass at $125mph$ bracket
\end{enumerate}

In the end, an operational mishap was the deciding factor in finishing in fourth place. While it is likely that the planner's indecisiveness may have surfaced again, we will never know how the vehicle would have performed if it had more fuel. However, in the end, the software stack demonstrated strong performance in executing the IAC Passing Competition.

In the future, fuel consumption will be more carefully monitored and accounted for. Additionally, further testing and focus will be on path planning, to help determine the root cause of the behavior seen in the final pass attempt. As the IAC evolves and tackles more complicated operational design domains (OODs), it is also important for the software stack to evolve as well. Future work will dismantle the assumptions and simplifications to unlock more general, robust performance.

\section{Lessons Learned}

\textbf{\subsubsubsection{Keep it Simple}}
At many critical points in the project, a decision had to be made on what direction to pursue for specific portions of the stack. What kind of controller do we implement? How should we do camera detection? What methods should we pursue for localization? The tendency was to pursue a complicated solution; however, we recognized that identifying a functional solution was more critical than identifying the best possible solution from the outset. In doing so, we chose to focus on simple solutions and build out complexity when necessary. 

One example of this struggle was LiDAR object detection. This proved to be challenging, as we recognized that clustering could be applied as a temporary solution, but was not intended to be a long-term solution. We also recognized that by implementing clustering while simultaneously building a parallel solution, we had a solution to fall back to. Eventually, with enough maturity, a more complicated solution (PointPillars) was produced that met our requirements and was orders of magnitude better than clustering. Additionally, this more complicated solution will scale better as the competition travels to new circuits and incorporates more agents.

\textbf{\subsubsubsection{The Smallest Issues have the Largest Consequences}}
Issues with time synchronization, balancing Ethernet network load across multiple links, cable and connector integrity, and more can severely diminish system performance. For example, the first week of testing with our recently assembled vehicle was progressing as planned, with quick integration and with the absence of major issues. However, the very next week, when testing on the Indianapolis Motor Speedway for the first time, the RTK GPS did not function properly. An entire test day, less than three weeks from the first IAC event, was lost. Eventually, the root cause was traced to our RTK login being used by a competitor's unit, causing a conflict with the RTK service. This issue was not caught the week prior due to our teams' testing schedules being on alternating days.

In Field Robotics, deploying a system requires understanding how each component interacts and what failures may occur. When many complicated pieces are combined into one package, it is easy to overlook the tiniest details. How does the RTK service handle two units trying to communicate to the server with the same license at the same time? How does DDS handle message delivery if a ROS 2 node cannot keep up with the message rate? How does that library being called decide how many parallel threads to use for processing and what are the downstream effects on other software components running on the same system? All of these issues can destroy the performance of a whole system, which is why testing and validation are so critical.

\textbf{\subsubsubsection{Know What to Test and Actually Do It}}
Due to the complexity of the systems being built, it is important to have a rigorous and principled testing regime. The size, speed, and operating costs associated with testing full-sized ARVs, such as the AV-21, make testing arduous, expensive, and rare. Offline testing, such as in simulation or off of collected datasets, is critical to catching issues. While the time pressures of a high-paced competition and the need to develop an entire ARV stack can make thorough validation difficult, our experience has shown that it is imperative to successful deployments. Balancing development and testing is non-trivial when working with limited resources.

Between Seasons One and Two, the team focused on identifying the software failures, and also determining why our development practices failed. In particular, the testing of the perception stack was completely revamped. Datasets were created by merging our log collections with logs from other teams so that the perception detections could be compared against the RTK GPS of both vehicles at any instant. Any perception code change was validated on this dataset and run on hardware that matched the performance of the ADLINK from Season One (see Figure \ref{fig:s1_s2_compute}). A simulator was developed to simulate the object detection pipelines, with customizable levels of noise, false positive rate, and output ``detection" rate, to aid the development and testing of the tracker pipeline. These practices, and more, allowed the team to more thoroughly evaluate performance and prepare for the Season Two events.

\textbf{\subsubsubsection{Autonomous Racing Demands Strong Algorithms and Systems}}
With this work, we are beginning to address some of the largest challenges in Autonomous Racing. We recognize that there are many failure modes with our approach and assumptions that it makes that only apply under our Operational Design Domain (ODD). What has been achieved is a full, baseline stack that is capable of participating in the Indy Autonomous Challenge head-to-head Passing Competition. 

However, in designing and building this software stack, it is clear that no one algorithm can meet every requirement. For example, Model Predictive Control (MPC) is a state-of-the-art control approach; but, a complex numerical optimization approach carries the risk of ill-conditioning or high computation time. A potential solution is to use LQR as a fallback for MPC. For both LQR and MPC, there still exists the potential for model mismatch, such as when driving at very high speeds, or at the traction limits of the tires. Solving these problems requires exploring multiple solutions, including designing better algorithms (i.e. robust MPC) and building better systems (i.e. safety monitoring and response). 

Designing the software stack requires a holistic approach. Individual components depend on a set of assumptions about their inputs, the problem, and what their output should be. A misalignment in assumptions between two successive components can lead to degraded performance. For example, a planner may assume an upper bound on the quality of the incoming agent beliefs and a certain level of performance from the trajectory tracking controller. If the planner is too optimistic, it may guide the vehicle too close to other agents. A better algorithm at one level (i.e. using PointPillars over clustering) allows dependent tasks to be more optimistic. Finding a good balance of performance and understanding how to set assumptions is a nontrivial task in systems-level engineering.

\section{Conclusion and Future Work}
We have presented a modular and fast software stack for an Autonomous Racing Vehicle (ARV) capable of navigating at high speeds with minimal lateral deviations, reliably detecting vehicles and tracking an opponent ARV at over $100m$ away, even at high speeds, and safely trailing and passing opponent ARVs. Modularity, speed, and efficiency permeate the entire stack through our choices of algorithms and the systems built around them.

With our approach, we have competed in the Indy Autonomous Challenge events in Indianapolis, Las Vegas, and Texas, which will serve as the base for our entry into future competitions. As MIT-Pitt-RW's approach has evolved, the team's performance has become increasingly competitive. The results for the competition are as follows: Did Not Finish (DNF), Did Not Qualify (DNQ), Quarter-Finalist, and Semi-Finalist. The testing, lessons learned, and data gleaned over this series of events, especially in Las Vegas, are informing future developments of the stack.

Moving forward, we intend to continue to validate our tracking and fusion stack and improve its performance. Additionally, we will build new models with data collected from tooling for auto-labeling; and evaluation metrics built from opponent GPS data will increase our stack's ability to detect competitor ARVs. With a more robust data set, we can explore alternative approaches to improve performance, particularly for long-range detection and velocity estimation. Finally, with a more intelligent controller and the introduction of online vehicle model estimation, we can improve our ability to navigate highly dynamic scenarios at even higher speeds. 

Our current software stack addresses many of the challenges laid out previously but notably does not address adversarial agents. Several research directions stem from interactions between ARVs; including motion prediction and forecasting in highly dynamic scenarios, planning under uncertainty in racing scenarios, planning to maximize the reward to the agent while minimizing the risk of collision or instability at high speeds, and more. We hope to explore several of these directions in the future to meet the challenges needed to solve full head-to-head autonomous racing.

\section{Acknowledgments}
As the only student-led team participating in the Indy Autonomous Challenge, the MIT-Pitt-RW team's successes would not have been possible without many supporters. We would first like to thank our current and past team members: Andrew Tresansky, Adam Barber, Alan Yu, Alexander Hadik, Alexander Pletta, Alisha Fong, August Soderberg, Anika Cheerla, Colin Mulloy, Eric Schneider, Eric Sherman, Erick Valencia Torres, Erin Kust, Gabriel Leuenberger, Haotian Tang, Ishan Baliyan, Jatin Mehta, Jeana Choi, Jiaming Huang, Jack Bowers, JP Ramassini, Jun Yin, Karmen Lu, Kazuki Shin, Kendrick Cancio, Kevin Shao, Kyle Anderson, Lucas Blascovich, Matthew Beveridge, Matthew Millendorf, Mia Lei, Michael Okolo, Miriam Dukaye, Morgan Vinesky, Nestor Diaz-Ortaz, Nick Wight, Ritesh Misra, Shanti Mickens, Sibo Zhu, Sydney Way, Thomas Detlefsen, Van Pham, Will Schwarting, Yilan Gao, Yueyamg Ying, and Zhijan Liu. Our team's success and growth would not have been possible without your contributions, thank you!

To our industry partners, Oracle for Research, and Mobilitas Insurance, your financial support, technical guidance, and assistance are an integral component of our team, we thank you.
Finally, Andrew Saba and Deva Ramanan are supported by the Carnegie Mellon University Argo AI Center for Autonomous Vehicle Research.


\printbibliography

\end{document}